\documentclass[sigconf]{acmart}

\copyrightyear{2024}
\acmYear{2024}
\setcopyright{rightsretained}
\acmConference[MM '24]{Proceedings of the 32nd ACM International Conference on Multimedia}{October 28-November 1, 2024}{Melbourne, VIC, Australia}
\acmBooktitle{Proceedings of the 32nd ACM International Conference on Multimedia (MM '24), October 28-November 1, 2024, Melbourne, VIC, Australia}
\acmDOI{10.1145/3664647.3681094}
\acmISBN{979-8-4007-0686-8/24/10}

\usepackage{newfloat}
\usepackage{listings}
\usepackage{threeparttable}
\usepackage{booktabs}
\usepackage{graphicx} % DO NOT CHANGE THIS
\usepackage{algorithm}
\usepackage{algorithmic}
\usepackage{caption,subcaption}
\usepackage{comment}
\usepackage{balance} 
\newtheorem{lemma}{Lemma}
\newtheorem{theorem}{Theorem}
\newtheorem{Assumption}{Assumption}

\begin{document}

%%
%% The "title" command has an optional parameter,
%% allowing the author to define a "short title" to be used in page headers.

%%
%% The "author" command and its associated commands are used to define
%% the authors and their affiliations.
%% Of note is the shared affiliation of the first two authors, and the
%% "authornote" and "authornotemark" commands
%% used to denote shared contribution to the research.
\title{FedBCGD: Communication-Efficient Accelerated Block Coordinate Gradient Descent for Federated Learning}

\author{Junkang Liu}
%\email{trovato@corporation.com}
%\orcid{1234-5678-9012}
%\author{G.K.M. Tobin}
%\authornotemark[1]
%\email{webmaster@marysville-ohio.com}
\affiliation{%
	\institution{School of Artificial Intelligence}
	\city{Xi'an}
	%\state{Ohio}
	\country{Xidian University, China}
}

\author{Fanhua Shang}
%\email{trovato@corporation.com}
%\orcid{1234-5678-9012}
%\author{G.K.M. Tobin}
%\authornotemark[1]
%\email{webmaster@marysville-ohio.com}
\authornote{Corresponding author}
\affiliation{%
	\institution{College of Intelligence and Computing}
	\city{Tianjin}
	%\state{Ohio}
	\country{Tianjin University, China}
}

\author{Yuanyuan Liu}
%\email{trovato@corporation.com}
%\orcid{1234-5678-9012}
%\author{G.K.M. Tobin}
%\authornotemark[1]
%\email{webmaster@marysville-ohio.com}
\authornotemark[1]
\affiliation{%
	\institution{School of Artificial Intelligence}
	\city{Xi'an}
	%\state{Ohio}
	\country{Xidian University, China}
}

\author{Hongying Liu}
\authornotemark[1]
\affiliation{%
	\institution{Medical College, Tianjin University\\Peng Cheng Laboratory}
	\city{Tianjin}
	%\country{Peng Cheng Laboratory, China}
	\country{ China}
}

%\email{xx@xx.xx}

\author{Yuangang Li}
\affiliation{%
	\institution{University of Southern California}
	% \streetaddress{1 Th{\o}rv{\"a}ld Circle}
	\city{Los Angeles}
	\country{US}}
%\email{yuangang@usc.edu}

\author{YunXiang Gong}
\affiliation{%
	\institution{School of Artificial Intelligence}
	\city{Xi'an}
	\country{Xidian University, China}}
%%
%% By default, the full list of authors will be used in the page
%% headers. Often, this list is too long, and will overlap
%% other information printed in the page headers. This command allows
%% the author to define a more concise list
%% of authors' names for this purpose.
\renewcommand{\shortauthors}{Junkang Liu et al.}
\begin{abstract}
	Although Federated Learning has been widely studied in recent years, there are still high overhead expenses in each communication round for large-scale models such as Vision Transformer. To lower the communication complexity, we propose a novel Federated Block Coordinate Gradient Descent (FedBCGD) method for communication efficiency. The proposed method splits model parameters into several blocks, including a shared block and enables uploading a specific parameter block by each client, which can significantly reduce communication overhead. Moreover, we also develop an accelerated FedBCGD algorithm (called FedBCGD+) with client drift control and stochastic variance reduction. To the best of our knowledge, this paper is the first work on parameter block communication for training large-scale deep models. We also provide the convergence analysis for the proposed algorithms. Our theoretical results show that the communication complexities of our algorithms are a factor $1/N$ lower than those of existing methods, where $N$ is the number of parameter blocks, and they enjoy much faster convergence  than their counterparts. Empirical results indicate the superiority of the proposed algorithms compared to state-of-the-art algorithms.
    The code is available at \url{https://github.com/junkangLiu0/FedBCGD}.
\end{abstract}

%%
%% The code below is generated by the tool at http://dl.acm.org/ccs.cfm.
%% Please copy and paste the code instead of the example below.
%%

%\received{13 Apr 2024}
%\received[revised]{21 Jul 2024}
%\received[accepted]{7 Aug 2024}

\maketitle

\section{Introduction}
\label{sec:intro}

\begin{figure}[ht]
	\centering
\includegraphics[width=0.4\textwidth]{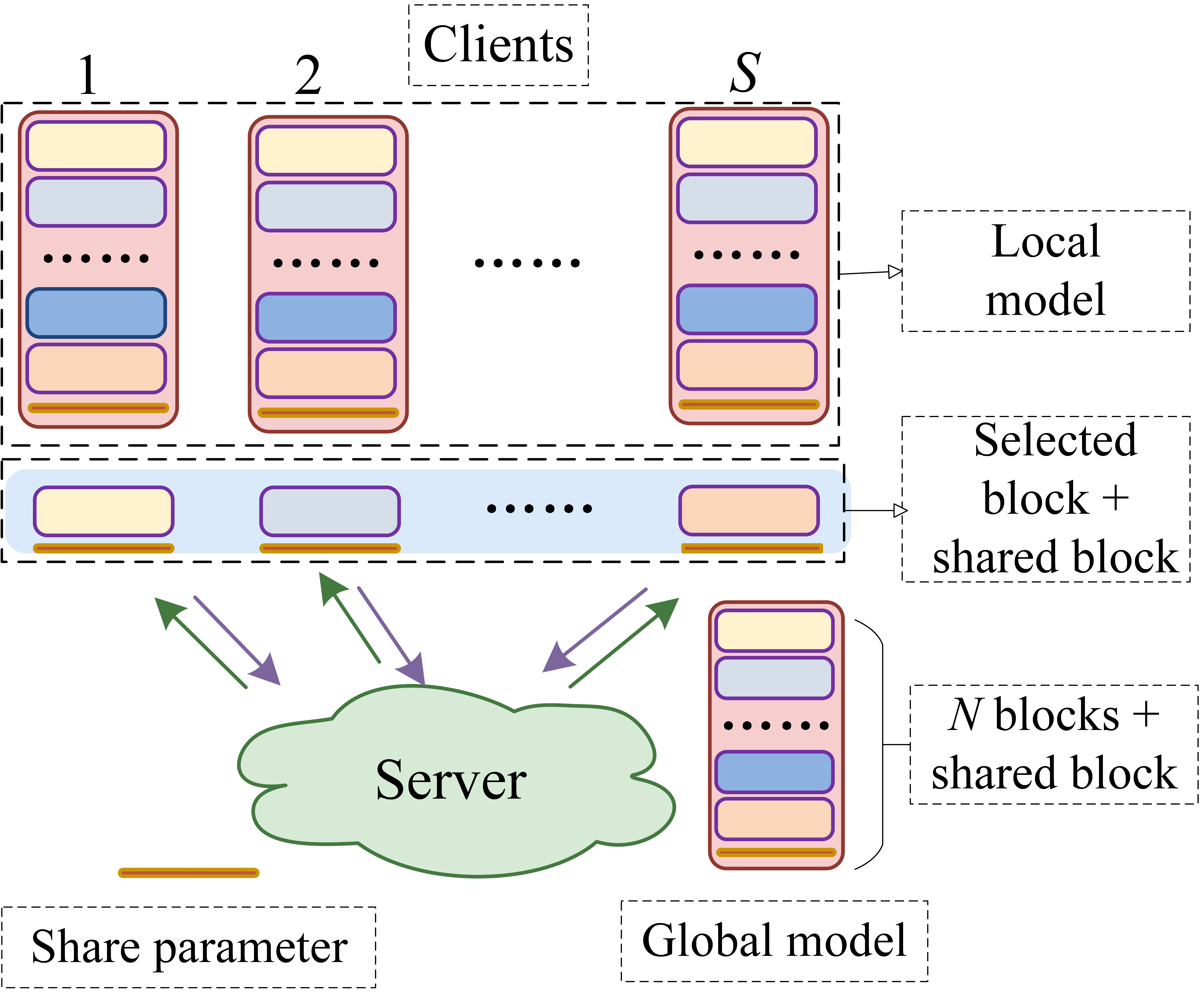}
	\vspace{-3mm}
    \caption{The diagram of the proposed FedBCGD framework, where $S\geq N$, $S$ and $N$ are the numbers of clients and parameter blocks, respectively. }
	\label{fig:1}
\end{figure}

Federated Learning (FL) is an emerging machine learning paradigm, which aims at achieving collaborative model training among multiple parties to preserve data privacy. FL achieves model training by training models locally on client devices and then uploading them to a central server for model aggregation \cite{mcmahan2017communication}. Compared to centralized learning in a data center \cite{goyal2017accurate}, the parallel computing clients of FL have private data stored in them and communicate remotely with a central server. The clients are responsible for local training, while the central server in charge of aggregating the models uploaded by each client. Currently, FL has been widely applied in different fields such as mobile intelligence devices, medical health, and financial risk control  \cite{rieke2020future,antunes2022federated,byrd2020differentially}.

Our theoretical results show that the communication complexities of our algorithms are a factor $1/N$ lower than those of existing methods, where $N$ is the number of parameter blocks, and they enjoy much faster convergence  than their counterparts.

In mainstream frameworks for federated learning, the communication between clients and their server is slow, costly, and unreliable \cite{konevcny2016federated}. 
%Therefore, the communication cost dominates federated learning and is the main bottleneck that needs to be addressed various practical applications application of FL \cite{konevcny2016federated}. 
In recent years, large models such as BERT and ChatGPT \cite{devlin2018bert,brown2020language} have emerged, leading to an exponential increase in the model size and data volume on FL clients. The upload of these large models further escalates the cost of communication in FL. To significantly lower the communication complexity, this paper proposes a novel method for FL, called Federated Block Coordinate Gradient Descent (FedBCGD) based on block coordinate descent (BCD) methods \cite{wright2015coordinate}. 

%

%Next, we explain the details of FedBCGD.

\newcommand{\tabincell}[2]{\begin{tabular}{@{}#1@{}}#2\end{tabular}}
\begin{table*}[!ht]
	\caption{Comparison of the communication complexities and communication overheads of different algorithms in the $\mu$-strongly convex setting, where $\sigma$ is the variance of stochastic gradients, $G$ is heterogeneity due to client data distribution, $S$ is the number of participating clients, $K\!=\!S / N$, $K$ is the number of clients involved in each parameter block, $N$ is the number of parameter blocks, and $T$ is the number of local training iterations. The number of floats sent per round by FedAvg is $d$, and $\mathcal{O}$ describes the worst-case complexity of different algorithms, where $\alpha=\frac{1}{1-\lambda}$. In non-convex settings, $\tau$ is the second-order heterogeneity (see \cite{karimireddy2020mime}), $G$ is the first-order heterogeneity,	$F:=f\left(\boldsymbol{x}^0\right)-f^{\star}$, and $f^{\star}$ is a minimum value of Problem (1) below.}\label{tab:tablenotes}
	\vspace{-2mm}
	\setlength{\tabcolsep}{8pt}
	\centering
	\begin{threeparttable}          %ÕâÐÐÒªÌí¼Ó
		\begin{tabular}{*6{c}}\toprule         %ÕâÐÐÒªÌí¼Ó
			%\rowcolor{orange!10}
			\text { Algorithm } & \tabincell{c}{Strongly convex\\ Communication complexity} & \tabincell{c}{Non-convex\\ Communication complexity} & \tabincell{c}{ Client\\ sample} &  \tabincell{c}{ Stochastic \\Gradient}& \tabincell{c}{Floats sent\\ per round} \\ \midrule
			%\rowcolor{green!40}
			
			\text {FedAvg \cite{mcmahan2017communication}}  &    
			$\mathcal{O}\Big(\frac{\sigma^2+G^2}{\mu S T \epsilon}+\frac{\sigma+G}{\mu \sqrt{\epsilon}}+\frac{\beta}{\mu} \log \frac{1}{\epsilon}\Big)$ & $ \mathcal{O}\Big(\frac{\beta \sigma^2}{T S \epsilon^2}+\frac{\sqrt{\beta} G+\sqrt{\frac{\beta}{T}} \sigma}{\epsilon^{\frac{3}{2}}}+\frac{F\beta}{\epsilon}\Big)$&\text {Yes } & \text {Yes  } & $d$ \\
			
			%\hline
			%\text { \textbf{FedBCGD} (ours) } &$ \mathcal{O}\Big(\frac{\sigma^2+G^2}{\mu K T  \epsilon}+\frac{\sigma+G}{\mu \sqrt{\epsilon}}+\frac{\beta}{\mu}\log \frac{1}{\epsilon}\Big) $&$ \mathcal{O}\Big(\frac{\beta \sigma^2}{TK \epsilon^2}+\frac{\sqrt{\beta} G+\sqrt{\frac{\beta}{T}} \sigma}{\epsilon^{\frac{3}{2}}}+\frac{F\beta}{\epsilon}\Big)$&\text {Yes } & \text {Yes  } & $d/N $   \\
			%\rowcolor{rgb}{1, 0.9, 0.7}
			\text { \textbf{FedBCGD} (ours) } &$\mathcal{O}\Big(\frac{\sigma^2+G^2}{\mu S T  \epsilon}+\frac{\sigma+G}{\alpha \mu N \sqrt{\epsilon}}+\frac{\beta}{ \mu N}\log \frac{1}{\epsilon}\Big)$&$\mathcal{O}\Big(\frac{\beta \sigma^2}{TS \epsilon^2}+\frac{\sqrt{\beta} G+\sqrt{\frac{\beta}{T}} \sigma}{N\epsilon^{\frac{3}{2}}}+\frac{F\beta}{N\epsilon}\Big)$ &\text {Yes } & \text {Yes  } &$d/N $ \\
			\hline
			
			\text {SCAFFOLD  \cite{karimireddy2020scaffold}}  & $\mathcal{O}\Big(\frac{\sigma^2}{\mu S T \epsilon}\!+\!\frac{\sigma}{\mu \sqrt{\epsilon}}\!+\!\Big(\frac{M}{S}\!+\!\frac{\beta}{\mu}\Big) \log \frac{1}{\epsilon}\Big)$&$ \mathcal{O}\Big(\frac{\beta \sigma^2}{T S \epsilon^2}+\frac{\sqrt{\frac{\beta}{T}} \sigma}{\epsilon^{\frac{3}{2}}}+\frac{\beta F}{\epsilon}\Big(\frac{M}{S}\Big)^{\frac{2}{3}}\Big)$& \text {Yes } & \text {Yes  } & $2d$ \\
			
			\text {FedLin \cite{mitra2021linear} } &$\mathcal{O}\Big({\frac{\beta}{\mu}} \log \frac{1}{\epsilon}\Big)$&—— &\text {No } & \text {No} & $2d$  \\	
			\text { S-Local-GD \cite{gorbunov2021local} } &$\mathcal{O}\Big({\frac{\beta}{\mu}} \log \frac{1}{\epsilon}\Big)$&——  &\text {No } & \text {No} & $2d$\\
			%	\text {MIME  \cite{karimireddy2020mime} }&——   &$\mathcal{O}\Big(\frac{G^2+\sigma^2}{\epsilon}+\frac{(G+\sigma) F \tau}{\epsilon^{\frac{3}{2}}}\Big)$ &\text {No } & \text {Yes } & $2d$  \\	
			\text { CE-LSGD \cite{patel2022towards} } &——  &$\mathcal{O}\Big(\frac{G F \tau}{M\epsilon^{\frac{3}{2}}}\Big)$ &\text {Yes } & \text {Yes } & $3d$  \\	
			\text {BVR-L-SGD \cite{murata2021bias} } &——  &$\mathcal{O}\Big(\frac{F \tau}{\epsilon}\!+\!\frac{F \beta}{\sqrt{T} \epsilon}\!+\!\frac{\sigma^2}{M T \epsilon}\!+\!\left(\frac{\sigma F \beta}{M T \epsilon}\right)^{\frac{3}{2}}\Big)$ &\text {Yes } & \text {Yes } & $3d$  \\	
			\text { \textbf{FedBCGD+} (ours) } &$\mathcal{O}\left(\left(\frac{M}{S}+\sqrt{\frac{\beta}{\mu}}\right)\log \frac{1}{\epsilon}\right)$&$ \mathcal{O}\big(\frac{\beta F}{\epsilon}\big(\frac{M}{S}\big)^{\frac{2}{3}}\frac{1}{N}^{\frac{1}{3}}\big)$&\text {Yes } & \text {Yes  } &$2d/N $\\
			
			\bottomrule			
		\end{tabular}
		%\begin{tablenotes}    %ÕâÐÐÒªÌí¼Ó£¬ ´ÓÕâ¿ªÊ¼
		%\footnotesize               %ÕâÐÐÒªÌí¼Ó
		%\item[(1)]This is a special case of S-Local-SVRG, which is a more general method presented in \cite{gorbunov2021local}. S-Local-GD arises as a special case when full gradient is
		%computed on each client.
		%\end{tablenotes}            %ÕâÐÐÒªÌí¼Ó
	\end{threeparttable} 
	\label{table:1}
\end{table*}

In FL, the upload speed of the client model is more than a hundred times slower than the download speed, so this paper mainly resolves the issue of upload communication cost. As shown in Figure \hyperref[fig:1]{1}, we divide the model parameter $\boldsymbol{x}$ into $N$ blocks and $\boldsymbol{x}_{s}$, i.e., $\boldsymbol{x}=\big[\boldsymbol{x}_{(1)}^{\top}, \ldots, \boldsymbol{x}_{(N)}^{\top},\boldsymbol{x}_{s}^{\top}\big]^{\top}$, where $\boldsymbol{x}_{s}$ denotes the shared parameters in each client (usually the parameters of the last layer classifier, and their number is small but important, \cite{luo2021no} suggests that the deeper the model, the greater the variance of the parameters. In FL, it is often the parameters in the last layer of the classifier that are most important and have a very small number of covariates (0.01\% of the overall number in ResNet-18)). Each client is responsible for optimizing 
one selected  parameter block $\boldsymbol{x}_{(j)}$ and shared parameter block $\boldsymbol{x}_{s}$. After local training for all model parameters, the updated parameter block $\boldsymbol{x}_{(j)}$ and shared parameter block $\boldsymbol{x}_{s}$ are sent to the server, which takes average aggregation of parameters for different parameter blocks to get the complete model. 
% $\boldsymbol{x}=\big[\boldsymbol{x}_{(1)}^{\top}, \ldots, \boldsymbol{x}_{(N)}^{\top},\boldsymbol{x}_{s}^{\top}\big]^{\top}$

The initial idea is to require each client to perform local updates only on the specified parameter block $\boldsymbol{x}_{(j)}$ and $\boldsymbol{x}_{s}$ while freezing the remaining parameter blocks (called FedBCGD\_freezing). After local training, the specified parameter blocks would be uploaded for model aggregation. However, due to a large drift between parameter blocks, such scheme often results in bad convergence in our experiments (see Figure \hyperref[fig:fig1]{5} for details). More specifically, only updating certain parameter blocks locally results in a large gap between the updated parameter blocks and other freezing parameter blocks, and it is not possible to establish good connections between parameter blocks during the server-side aggregation process.

Therefore, we propose a novel FedBCGD method to address these issues. In the proposed algorithm, we employ stochastic gradient descent to update all parameters instead of  parameter freezing  during local training, but only transmit two specified parameter blocks ($\boldsymbol{x}_{(j)}$ and $\boldsymbol{x}_{s}$) during the upload process. In addition, to compensate for some missing parameters in block parameter transmission, we add parameter block momentum on the server side. This algorithm design maintains the advantages of low communication costs and has demonstrated a significantly improved convergence speed in our experiments (see Figure \hyperref[fig:fig1]{5} for details). Moreover, adding one shared parameter block in each client can significantly improve accuracy performance. However, due to the impact of data heterogeneity, it still leads to inconsistent update directions between parameter blocks, called parameter block drift, resulting in poor performance of the aggregated model. Thus, we also propose an accelerated version (called FedBCGD+) to address data heterogeneity.  
%BCD is an iterative optimization algorithm that divides the parameters to be optimized into several blocks and performs gradient descent updates on each block sequentially. 
The main difference between FedBCGD and BCD is that FedBCGD incorporates shared one small parameter block and updates all model parameters in each client (i.e., no parameter freezing), while BCD only updates one parameter block in each iteration.
%\vspace{-4mm}
%\subsection{Motivations and contributions}

\textbf{Our motivations and contributions:}  To address these issues such as communication effectiveness, acceleration, theoretical guarantees and parameter block drift, we design a novel federated block coordinate descent  framework FedBCGD and its acceleration variant FedBCGD+ for training large-scale deep models such as  Transformer. 
%Influenced by parameter block drift, there is reduced correlation between different parameter blocks during the aggregation process, potentially resulting in a relatively slow pace of global model updates. 
%To solve the issue of parameter block drift, we propose an effective solution, namely the FedBCGD+ algorithm with parameter block update correction.
The main contributions of this work are listed as follows:

\textbf{$\bullet$ Novel FL Paradigm:} We propose the first block coordinate descent  algorithm FedBCGD for horizontal FL. FedBCGD demonstrates remarkable communication efficiency in distributed learning scenarios. That is, this paper presents the first block coordinate descent algorithm for horizontal FL. Moreover, we also  introduce an accelerated version, FedBCGD+, which exhibits an even faster convergence rate while maintaining high communication efficiency.

\textbf{$\bullet$ Convergence Analysis:} We provide a thorough analysis of the convergence properties of the proposed FedBCGD algorithm and its accelerated version, FedBCGD+. By investigating the impact of partitioned parameter blocks, the number of clients, and the local training rounds, we provide valuable insights into their convergence behavior. From a practical perspective, FedBCGD+ achieves faster convergence than FedBCGD, and it is proved  faster from a theoretical perspective. Moreover, FedBCGD+ has a much lower communication complexity than existing algorithms in strong convexity settings (e.g., $\mathcal{O}\big(\big(\frac{M}{S}+\sqrt{\frac{\beta}{\mu}}\big)\log \frac{1}{\epsilon}\big)$ for FedBCGD+ vs. $\mathcal{O}\big(\frac{\sigma^2}{\mu S T \epsilon}\!+\!\frac{\sigma}{\mu \sqrt{\epsilon}}\!+\!\big(\frac{M}{S}\!+\!\frac{\beta}{\mu}\big) \log \frac{1}{\epsilon}\big)$ for SCAFFOLD \cite{karimireddy2020scaffold}. Furthermore, we can achieve a significant lower communication complexity of $ \mathcal{O}\big(\frac{\beta F}{\epsilon}\big(\frac{M}{S}\big)^{{2}/{3}}\frac{1}{N}^{{1}/{3}}\big)$ in the non-convex setting, compared to that of SCAFFOLD, $\mathcal{O}\big(\frac{\beta \sigma^2}{T S \epsilon^2}+\frac{\sqrt{\frac{\beta}{T}} \sigma}{\epsilon^{{3}/{2}}}+\frac{\beta F}{\epsilon}\big(\frac{M}{S}\big)^{{2}/{3}}\big)$. In other words, the communication complexities of our algorithms are a factor $1/N$ lower than those of existing methods, where $N$ is the number of parameter blocks.

%Also in both theory and practice, we show that increasing the number of blocks $N$ can reduce the communication complexities.

%\textbf{$\bullet$ Shared Parameter Strategy:} In the proposed algorithms, we employ one shared parameter updating strategy, where the number of the shared parameter is very small. That is, each client uploads the shared parameter block in addition to their specific parameter block. This strategy aims to expedite the convergence of FedBCGD and its accelerated variant.
\textbf{$\bullet$ Overcoming Data Heterogeneity :}  The convergence of FL algorithms is hindered by two sources of high variance: (i) heterogeneous clients, and (ii) the noise from local stochastic gradients. We propose two sets of control variance variables to reduce client heterogeneity and the noise variance of the local gradients in FedBCGD+. And we demonstrate the validity of the two sets of control variables through theory and experiment.

%\end{itemize}
\section{Related Work}
%In this section, we mainly review existing FL and block coordinate descent methods.
We  review existing FL and block coordinate descent methods.\\
\textbf{$\bullet$ Local Training:} Local Training (LT) is a communication-acceleration technique for FL \cite{mcmahan2017communication}.
One key challenge in LT is client drift, where the local model of each client gradually approaches the minimum of its own local cost function $f_i$ after multiple local GD steps.
To address this issue, SCAFFOLD \cite{karimireddy2020scaffold} is proposed, which is to incorporate control variates to correct for client drift and ensure linear convergence to the exact solution. Subsequent algorithms such as S-Local-GD \cite{gorbunov2021local} and FedLin \cite{mitra2021linear} also aimed to provide similar convergence properties. 
The analysis of algorithms for non-convex FL can be classified into several approaches. SCAFFOLD \cite{karimireddy2020scaffold} is the first federated algorithm capable of eliminating client data heterogeneity. However, its convergence speed is still affected by stochastic gradients, achieving only a convergence rate of $\mathcal{O}(1/\epsilon^2)$. MIME \cite{karimireddy2020mime} is essentially a combination of local SGD and variance reduction techniques as in SVRG \cite{johnson2013accelerating}, with a derived communication complexity of $\mathcal{O}(1/\epsilon^{3/2})$. BVR-L-SGD \cite{murata2021bias} assumed second-order data heterogeneity and achieved a communication complexity of $\mathcal{O}(1/\epsilon)$ with full client participation. The two-sided momentum (STEM) algorithm \cite{khanduri2021stem} can also attain a communication complexity of $\mathcal{O}(1/\epsilon)$ with full client participation. Inspired by the Storm algorithm \cite{CutkoskyO19}, CE-LSGD \cite{patel2022towards} can achieve a communication complexity of $\mathcal{O}(1/\epsilon^{3/2})$ with partial client participation and $\mathcal{O}(1/\epsilon)$ when all clients participate. 
FedBCGD \cite{liu2024fedbcgd} proposes an accelerated block coordinate gradient descent framework for FL.
FedSWA \cite{liuimproving} improves generalization under highly heterogeneous data by stochastic weight averaging.
FedAdamW \cite{liu2025fedadamw} introduces a communication-efficient AdamW-style optimizer tailored for federated large models.
FedNSAM \cite{liu2025consistency} studies the consistency relationship between local and global flatness in FL.
FedMuon \cite{liu2025fedmuon} accelerates federated optimization via matrix orthogonalization.
DP-FedPGN \cite{liu2025dp} develops a penalizes gradient norms to encourage globally flatter minima in DP-FL.
FedPAC \cite{liu2026tamingpreconditionerdriftunlocking} mitigates preconditioner drift to unlock the potential of second-order optimizers.

\textbf{$\bullet$ Block Coordinate Descent Methods:}
The block coordinate descent method is one of the most successful algorithms in the field of big data optimization. BCD is based on the strategy of updating a single coordinate or a single block of coordinate of a vector of variables at each iteration, which usually significantly reduces the memory requirements as well as the arithmetic complexity of a single iteration. The effectiveness of the BCD method for training deep neural networks (DNNs) has been demonstrated in recent years \cite{zeng2019global}. However, due to the highly non-convex nature of deep neural networks, its convergence is difficult to maintain. In addition, BCD can be easily implemented in a distributed and parallel manner \cite{mahajan2017distributed,richtarik2016distributed}.  \citet{liu2022fedbcd} proposed a vertical FL \cite{liu2022vertical} framework (FedBCD) for distributed features, in which parties share only the internal product of model parameters and raw data for each sample during each communication. Unlike the above works, this paper proposes the first block coordinate descent algorithm for horizontal FL. Horizontal FL is applied to scenarios where the client's datasets have the same feature space and different sample spaces \cite{yang2019federated}. %(what is it? what is its advantage?）

\textbf{$\bullet$ Communication-efficient FL: }Communication efficient FL algorithms can be divided into two categories, quantization and sparsification compression methods. The classical FL quantization method is proposed by Reisizadeh \cite{reisizadeh2020fedpaq}, which is a cycle averaging and quantization processing method named FedPAQ, and the quantization compression generally belongs to the unbiased compressions. While sparsification methods include $top$-$k$ and $rand$-$k$ methods \cite{sattler2019robust}, $top$-$k$ method is a biased compression method that uploads the gradient at the first $k$ large positions in the gradient to the server, while $rand$-$k$  method is an unbiased compression method that uploads the gradient at random $k$ positions to the server.  FedBCGD is different from all of the above methods and utilizes the idea of block gradient descent to address federated efficient communication, in addition to the above mentioned compression method that allows for secondary compression of our transferred block gradient to achieve more efficient communication, which is demonstrated in the following experiment.

\vspace{-2mm}

\section{Communication-Efficient Block Coordinate Gradient Descent  FL}

In this section, we propose a new communication-efficient block coordinate gradient descent  FL algorithm FedBCGD, and its pseudocode is given in Algorithm \hyperref[algorithm:1]{1}. 

We formalize the FL problem as the minimization of a sum of stochastic functions:
\vspace{-1mm}
%We formalize the federated learning problem as the minimization of a sum of stochastic functions and only access random samples:
%\vspace{-2mm}
%\begin{equation}
%	\min _{\boldsymbol{x} \in \mathbb{R}^d}\Big\{f(\boldsymbol{x}):=\frac{1}{M} \sum_{i=1}^M\Big(f_i(\boldsymbol{x}):=\mathbb{E}_{\zeta_i}\big[f_i\big(\boldsymbol{x} ; \zeta_i\big)\big]\Big)\Big\},
%\end{equation}
%where the function $f_i$ denotes the loss function on client $i$, $M$ is the number of clients. In this paper, we assume that $f$ is a $\beta$-smooth function. $\zeta_i$ is a random variable due to the client's use of stochastic gradient updates.
\begin{equation}
	\min _{\boldsymbol{x} \in \mathbb{R}^d}\Big\{f(\boldsymbol{x}):=\frac{1}{M} \sum_{i=1}^M\Big(f_i(\boldsymbol{x}):=\frac{1}{n_i} \sum_{\nu=1}^{n_i} f_i\left(\boldsymbol{x} ; \zeta_{i, \nu}\right)\Big)\Big\},
\end{equation}
where the function $f_i$ denotes the loss function on client $i$, $M$ is the number of clients, $n_i$ is the number of data points in client $i$,  and $\left\{\zeta_{i, 1}, \ldots, \zeta_{i, n_i}\right\}$ denotes  the local data of the $i$-th client. In this paper, we assume that each $f_i$ is a $\beta$-smooth function. 

\subsection{The proposed FedBCGD Algorithm}

\begin{algorithm}[h]
	\caption{FedBCGD}
	\begin{algorithmic}[1] %[1] enables line numbers
		\STATE $\textbf{Initialize } \boldsymbol{x}_{i}^{0,0}=\boldsymbol{x}^{i n i t}$, $\forall i \in [M]$.
		\STATE \textbf{Divide} the model parameters $\boldsymbol{x}$ into $N\!+\!1$ blocks.
		\FOR{$r=0,...,R$}
		\STATE{\textbf{Client:}}
		\STATE{$\textbf{Sample} \text{ clients } \mathcal{ S} \subseteq\{1, \ldots, M\}$,  $|\mathcal{S}|=N\cdot K$;}
		\STATE{ \textbf{Divide} \text{the sampled clients into} $N$ \text{client blocks};}
		\STATE $\textbf{Communicate }(\boldsymbol{x}^r) \text { to all clients } i \in \mathcal{S}$;
		\FOR{$j=1, \ldots, N$ client blocks in parallel}	
		\FOR{$k=1, \ldots, K$ clients in parallel}	
		\FOR{$t=1, \ldots, T$ local update}	
		\STATE \!Compute batch gradient $\nabla\! f_{k, j}\big(\boldsymbol{x}_{k, j}^{r,t} ; \zeta\big)$,
		\STATE$\boldsymbol{x}_{k, j}^{r, t+1}=\boldsymbol{x}_{k, j}^{r, t}-\eta \nabla f_{k, j}\big(\boldsymbol{x}_{k, j}^{r, t} ; \zeta\big)$;				
		\ENDFOR
		\STATE Send $\boldsymbol{x}_{k, j,(j)}^{r, T}$, $\boldsymbol{x}_{k, j,s}^{r, T}$ to server;
		\ENDFOR
		\ENDFOR
		\STATE \textbf{Server:}
		\FOR{$j=1, \ldots, N$ Blocks in parallel}	
		\STATE Block $j$ computes,
		\STATE$\boldsymbol{x}_{(j)}^r=\frac{1}{K} \sum_{k=1}^K \boldsymbol{x}_{k, j,(j)}^{r, T}$;
		$v_{(j)}^r=\lambda v_{(j)}^{r-1}+\boldsymbol{x}_{(j)}^{r}-\boldsymbol{x}_{(j)}^{r-1}$;
		\STATE$\boldsymbol{x}_{(j)}^r=\boldsymbol{x}_{(j)}^{r}+v_{(j)}^{r},$
		\ENDFOR
		\STATE$\boldsymbol{x}_{s}^{r}=\frac{1}{NK}\sum_{j=1}^N\sum_{k=1}^K\boldsymbol{x}_{k, j,s}^{r, T}$;	$v_{s}^r=\lambda v_{s}^{r}+\boldsymbol{x}_{s}^{r-1}-\boldsymbol{x}_{s}^{r-1}$;
		\STATE$\boldsymbol{x}_{s}^r=\boldsymbol{x}_{s}^{r}+v_{s}^{r} ;\boldsymbol{x}^r=\big[\boldsymbol{x}_{(1)}^{r \top}, \ldots, \boldsymbol{x}_{(N)}^{r \top},\boldsymbol{x}_{s}^{r \top}\big]^{\top}$;
		\STATE $\boldsymbol{v}^r=\big[\boldsymbol{v}_{(1)}^{r \top}, \ldots, \boldsymbol{v}_{(N)}^{r \top},\boldsymbol{v}_{s}^{r \top}\big]^{\top}$;
		\ENDFOR
	\end{algorithmic}
	\label{algorithm:1}
\end{algorithm}

\begin{figure}[t]
	\centering
	\includegraphics[width=0.3\textwidth]{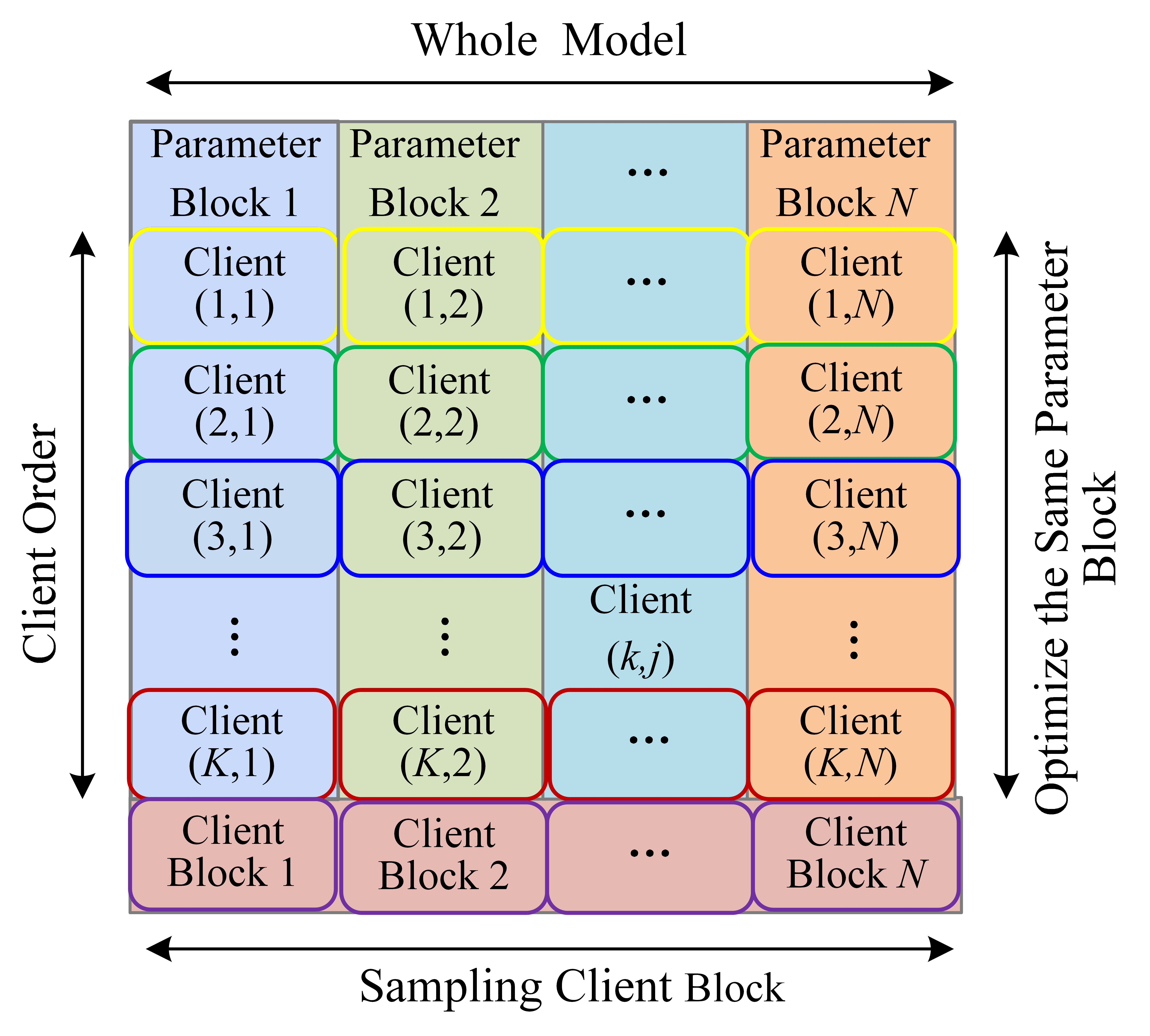}
	\vspace{-2mm}
	\caption{  
		The client parameter block allocation in FedBCGD. For the sake of convenience, we suppose $S=N\cdot K$ clients are sampled and divided into $N$ client blocks, i.e., $K$ clients for each client block. The clients in the $i$-th client block  are responsible for optimizing the upload parameter block $i$.}
	\label{fig:2}
	\vspace{-2mm}
\end{figure}
%The client below the Parameter Block is responsible for optimizing the Parameter Block.

We firstly divide the global model $\boldsymbol{x}$ into $N$ blocks of parameters and one shared block, each of which can have a different number of parameters,
\vspace{-4mm}
\begin{equation}
	\boldsymbol{x}=\big[\boldsymbol{x}_{(1)}^{\top}, \ldots, \boldsymbol{x}_{(N)}^{\top},\boldsymbol{x}_{s}^{\top}\big]^{\top}.
\end{equation}
We divide the sampled $S=N\cdot K$ clients into $N$ client blocks with $K$ clients in each client block (see Figure \hyperref[fig:2]{2}). These $N$ parameter blocks are distributed to the selected $N $ client blocks, where each parameter block will be optimized by $K$ clients. Due to significant differences in communication capabilities among different clients, parameter blocks with smaller parameter values can be assigned to clients with poorer communication capabilities, while parameter blocks with larger parameter values can be assigned to clients with better communication capabilities. This prevents clients with the smallest resources from becoming bottlenecks in FL. We define $\boldsymbol{x}_{k,j}$ as the local parameters of $k$-th client in $j$-th client block  (as client$_{k, j}$). Each client performs $T$ local stochastic gradient steps on its respective client block, by using a minibatch in each iteration:
\vspace{-2mm}
\begin{equation}
	\boldsymbol{x}_{k, j}^{r,t+1}=\boldsymbol{x}_{k, j}^{r,t}-\eta \nabla f_{k, j}\big(\boldsymbol{x}_{k, j}^{r,t};\zeta\big),
	\vspace{-1mm}
\end{equation}

where $\boldsymbol{x}_{k, j}^{t+1}$ is the $t\!+\!1$-th local update whole parameter of client$_{k, j}$, and $\boldsymbol{x}_{k, j,(j)}^{t+1}$ is the $j$-th parameter block of client$_{k, j}$. $\nabla f_{k, j,(j)}$ is  the $j$-th gradient block in client$_{k, j}$ (see Figure \hyperref[fig:2]{2}). The local client of FedBCGD is used to update all model parameters $\boldsymbol{x}$ and send the selected parameter block $\boldsymbol{x}_{(j)}$ and $\boldsymbol{x}_{s}$ to the server.

Below, we will describe the proposed server-side aggregation operation. For the $k$-th client of the $j$-th parameter block, it sends the parameter block $\boldsymbol{x}_{k, j,(j)}^{r, T}$ and $\boldsymbol{x}_{k, j,s}^{r, T}$ to server after $T$ local updates. The central server performs separate aggregation operations on $\boldsymbol{x}_{(j)}$ and $\boldsymbol{v}_{(j)}$ for the $j$-th parameter block in Lines 20-22 of Algorithm \hyperref[algorithm:1]{1}.
%\vspace{-2mm}
%\begin{align}
%&\boldsymbol{x}_{(j)}^r=\frac{1}{K} \sum_{k=1}^K \boldsymbol{x}_{k, j,(j)}^{r, T},\\
%&v_{(j)}^r=\lambda v_{(j)}^{r-1}+\boldsymbol{x}_{(j)}^{r}-\boldsymbol{x}_{(j)}^{r-1},\\
%&\boldsymbol{x}_{(j)}^r=\boldsymbol{x}_{(j)}^{r}+v_{(j)}^{r}.
%\end{align}
Next, we update the shared parameter block in Lines 24-26 of Algorithm \hyperref[algorithm:1]{1}.
%\vspace{-2mm}
%\begin{align}
%&\boldsymbol{x}_{s}^{r}=\frac{1}{NK}\sum_{j=1}^N\sum_{k=1}^K\boldsymbol{x}_{k, j,s}^{r, T},\\
%&v_{s}^r=\lambda v_{s}^{r-1}+\boldsymbol{x}_{s}^{r}-\boldsymbol{x}_{s}^{r-1},\\
%&\boldsymbol{x}_{s}^r=\boldsymbol{x}_{s}^{r}+v_{s}^{r}.
%\end{align}
%\begin{equation}	\!\!\!\boldsymbol{x}_{(j)}^r\!=\!\frac{1}{K} \sum_{k=1}^K \boldsymbol{x}_{k, j,(j)}^{r, T},\,
%		\boldsymbol{x}_{s}^{r}\!=\!\frac{1}{NK}\sum_{j=1}^N\sum_{k=1}^K\boldsymbol{x}_{k, j,s}^{r, T}.\!\!\!
%\end{equation}
Finally, all the parameter blocks are combined into a complete model,
%\vspace{-2mm}
%\begin{equation}
$\boldsymbol{x}^r=\big[\boldsymbol{x}_{(1)}^{r \top}, \ldots, \boldsymbol{x}_{(N)}^{r \top},\boldsymbol{x}_{s}^{r \top}\big]^{\top}$, and the momentum term is $\boldsymbol{v}^r=\big[\boldsymbol{v}_{(1)}^{r \top}, \ldots, \boldsymbol{v}_{(N)}^{r \top},\boldsymbol{v}_{s}^{r \top}\big]^{\top}$.
%\end{equation}
Before the next iteration starts, the client transfers all model parameters $\boldsymbol{x}^r$ to the selected client and tells the client which model parameter block needs to be uploaded. $v_{(j)}^r$ is the $j$-th block of the momentum term $v^r$, and $\lambda$ is the momentum parameter. The momentum term $v_{(j)}^r$ considers the model's continuous updates over time, making the updating process smoother. More specifically, it remembers and utilizes the direction and speed of previous model parameter updates, thereby accelerating the convergence speed of the model.
% When the number of participating clients is low, the momentum term can leverage past gradient information to increase the client participation rate and maintain the model's updates to a certain extent.
%\vspace{-2mm}
\subsection{Our FedBCGD+ Algorithm}

The FedBCGD+ algorithm is an extension of our FedBCGD algorithm based on the principles of variance reduction in SVRG \cite{johnson2013accelerating}. And its details are presented in the Appendix. Note that the server-side updates in FedBCGD+ are consistent with FedBCGD, while the new proposed client-side update of our FedBCGD+ algorithm is formulated as follows:
\vspace{-2mm}
\begin{equation}
	\begin{split}
		\!\!\!\!\boldsymbol{x}_{k, j}^{r, t+1}=&\underbrace{\boldsymbol{x}_{k, j}^{r, t}-\eta \nabla f_{k, j}\big(\boldsymbol{x}_{k, j}^{r, t} ; \zeta\big)}_{\text {Stochastic Gradient Descent}}+\underbrace{\eta \mathbf{c}-\eta \mathbf{c}_{k, j}}_{\text {Client Drift Control Variate  }}\\
		&+\underbrace{\eta \nabla f_{k, j}\big(\boldsymbol{x}^{r}\big)-\eta \nabla f_{k, j}\big(\boldsymbol{x}^r ; \zeta\big).}_{\text {Stochastic Variance Reduction}}
	\end{split}
\end{equation}
Each client-side update consists of a stochastic gradient descent term, one client drift control variate term and a variance reduction term, which is different from all existing works such as \cite{karimireddy2020scaffold}.

FedBCGD+ maintains a state for each client (the client control variate $\boldsymbol{c}_i$) and the server (the server control variate $c$).
Here, $\boldsymbol{c}_{k, j}^{+}\!=\!\nabla f_{k, j}(\boldsymbol{x}^r)$, and we need to send $\boldsymbol{x}_{k, j,(j)}^{r, T} , \Delta \boldsymbol{c}_{k,j,(j)}\!=\!\boldsymbol{c}_{k, j,(j)}^{+}\!-\!\boldsymbol{c}_{k, j,(j)}$, $\Delta \boldsymbol{c}_{k, j,s}=\boldsymbol{c}_{k, j,s}^{+}-\boldsymbol{c}_{k, j,s}$ to the server, $\boldsymbol{c}_i = \boldsymbol{c}_i^{+}$.
We update $\boldsymbol{c}$ on the server-side as follows:
\vspace{-2mm}
\begin{align}
	&\boldsymbol{c}_{(j)}=\boldsymbol{c}_{(j)}+\frac{1}{M} \sum_{k=1}^{K} \Delta \boldsymbol{c}_{k, j,(j)},
	\boldsymbol{c}_{s}=\boldsymbol{c}_{s}+\frac{1}{MN}\sum_{j=1}^{N} \sum_{k=1}^{K} \Delta \boldsymbol{c}_{k, j,s},\\
	&\boldsymbol{c}=\big[\boldsymbol{c}_{(1)}^{\top}, \ldots, \boldsymbol{c}_{(N)}^{\top},\boldsymbol{c}_{s}^{\top}\big]^{\top}.
\end{align}
\vspace{-2mm}
%The server update of our FedBCGDM+ algorithm can be written as follows:
%\begin{align}
%	&v_{(j)}^r=\lambda v_{(j)}^{r-1}+\boldsymbol{x}_{(j)}^{r-1}-\boldsymbol{x}_{(j)}^{r-2}, \\
%	&\boldsymbol{x}_{(j)}^r=\boldsymbol{x}_{(j)}^{r-1}+v_{(j)}^{r-1},\\
%	&\boldsymbol{c}_{(j)}=\boldsymbol{c}_{(j)}+\frac{K}{M} \sum_{k=1}^{K} \Delta \boldsymbol{c}_{k, j,(j)},\\
%	&\boldsymbol{x}^r=\big[\boldsymbol{x}_{(1)}^{r \top}, \ldots, \boldsymbol{x}_{(N)}^{r \top}\big]^{\top}, \boldsymbol{c}=\big[\boldsymbol{c}_{(1)}^{\top}, \ldots, \boldsymbol{c}_{(N)}^{\top}\big]^{\top}.\!\!\!
%\end{align}
%is similar to FedAvgM algorithm,

The key of our FedBCGD+ algorithm for improving the convergence speed is based on the following observation. The convergence of FL algorithms is hindered by two sources of high variance: (i) the global server aggregation step and multiple local updates, which are exacerbated by client heterogeneity, and (ii) the noise from local client-level stochastic gradients.

In the local update in Eq.\ (4), the first term involves stochastic gradient descent, the second term incorporates client heterogeneity control inspired by SCAFFOLD \cite{karimireddy2020scaffold}, and the third term adopts one stochastic variance reduction technique as in SVRG \cite{johnson2013accelerating} to reduce the variance of stochastic gradients.
By integrating these three components, our algorithm effectively addresses the challenges posed by heterogeneous clients and noisy local gradients, leading to a significant improvement in the convergence speed during the FL process. Compared with existing algorithms such as SCAFFOLD, and our FedBCGD,  FedBCGD+ has a faster convergence rate, as shown in the following  theoretical results. 

\section{Theoretical Guarantees}
In this section, we provide rigorous theoretical analysis for all the proposed algorithms, and the detailed proofs are included in the Appendix. The theoretical analysis of our FedBCGD algorithm is not a simple parallelization extension of the traditional BCD algorithm but an innovative theoretical analysis framework. Compared with related work, the two proposed algorithms have some theoretical advantages, including faster convergence rates and lower communication complexities. For the convenience of theoretical analysis, we ignore the shared  block in the algorithms.

\begin{table*}[t]
	\caption{Comparison of the average testing accuracy (\%) on CIFAR100, where the heterogeneity parameter is $\rho=0.6$, total communication floats are $1000d$, and the number of blacks is $N=5$.
		The number in brackets indicates the number of communication floats to reach the target accuracy. Note that centralised SGD refers to using SGD to train models on a single machine.}
	\label{tab:cross-device}
	\vspace{-2mm}
	\centering
	\setlength{\tabcolsep}{13pt}
	\begin{tabular}{llccccc}
		\hline & CIFAR100  & LeNet-5 (40\%) & VGG-11 (48\%) & ResNet-18 (54\%) & VGG-19 (45\%)   \\
		%\hline \multirow{6}{*}{$E=5$}
		\hline
		&Centralised  SGD & $53.7\pm0.2 $ & $56.3\pm0.3 $ & $62.2\pm0.1 $ & $58.9\pm0.1 $\\
		& FedAvg \cite{mcmahan2017communication} & $41.2\pm0.2 \;(558d) $ & $48.7\pm0.4\; (720d)$ & $54.2\pm0.2 \;(927d)$ & $47.6\pm0.1\; (735d)$\\
		& FedAvgM \cite{hsu2019measuring} & $48.2\pm0.5\; (277d) $ & $51.7\pm0.6\; (299d)$ & $ 61.8\pm0.8\; (398d) $ & $56.0\pm0.3\; (403d)$\\
		& FedAdam \cite{reddi2020adaptive} & $ 46.2\pm0.8 \;(391d)$ & $50.9\pm0.5 \;(597d)$ & $ 53.9\pm0.4 \;(\infty)$& $58.7\pm0.2\; (367d)$ \\
		& SCAFFOLD \cite{karimireddy2020scaffold} & $ 50.3\pm0.2 \;(214d) $ & $47.9\pm0.2 \;(\infty)$ & $52.3\pm0.2 \;(\infty)$ & $58.3\pm0.5\; (556d)$\\
		& FedDC \cite{gao2022feddc} & $ 53.2\pm0.3\; (302d)$ & $48.2\pm0.2\; (956d)$ & $46.6\pm0.1 \;(\infty)$ & $56.8\pm0.4 \;(321d)$\\
		%& \textbf{FedBCGD (ours)} & $ 42.3\pm0.1(118d)  $ & $48.9\pm0.2(217d)$ & $ 61.6\pm0.3(311d) $ & $54.9\pm0.2(\mathbf{156d})$\\
		&\textbf{FedBCGD (ours)} & $ \mathbf{55.7\pm0.4}\;(77d) $ & $\mathbf{62.2\pm0.4}\;(107d)$ & $\mathbf{68.1\pm0.5}\;(277d) $ & $61.1\pm0.3\;(206d)$\\
		& \textbf{FedBCGD+ (ours)} & $55.6\pm0.3\mathbf{\;(75d)}   $ & $58.7\pm0.3\mathbf{\;(105d)}$ & $ 65.1\pm1.8\mathbf{\;(154d)} $ & $\mathbf{63.6\pm0.4} \mathbf{\;(176d)}$\\
		\hline
	\end{tabular}
	\label{table:table 2}
\end{table*}

%\begin{comment}
\begin{table*}[t]
	\caption{Comparison of the average testing accuracy (\%) over the last 10\% rounds of each algorithm on CIFAR10, where the heterogeneity parameter is $\rho=0.6$, total communication floats are $1000d$, the number of blacks is set to $N=5$.}
	\label{tab:cross-device}
	\vspace{-2mm}
	\centering
	\setlength{\tabcolsep}{13pt}
	\begin{tabular}{llcccc}
		\hline & CIFAR10  & LeNet-5 (78\%) & VGG-11 (83\%) & ResNet-18 (88\%) & VGG-19 (84\%)  \\
		%\hline \multirow{6}{*}{$E=5$}
		\hline
		&Centralised  SGD & $83.1\pm0.2 $ & $87.4\pm0.3 $ & $90.1\pm0.1 $ & $88.6\pm0.1 $\\
		& FedAvg \cite{mcmahan2017communication} & $79.6\pm0.3 \;(498d)$ & $83.3\pm0.7 \;(630d)$ & $89.0\pm0.5 \;(698d)$ & $84.9\pm0.7 \;(499d)$\\
		& FedAvgM \cite{hsu2019measuring} & $81.1\pm0.6 \;(360d) $ & $83.7\pm0.4 \;(830d)$ & $ 89.1\pm0.7 \;(882d) $ & $87.4\pm0.5 \;(252d)$\\
		& FedAdam \cite{reddi2020adaptive} & $78.3\pm1.2 \;(860d) $ & $85.4\pm1.1 \;(478d) $ & $ 81.1 \pm1.3 \;(\infty)$& $87.5\pm0.9 \;(298d)$ \\
		& SCAFFOLD \cite{karimireddy2020scaffold} & $82.8 \pm0.7 \;(540d) $ & $86.9\pm0.6\;(278) $ & $ 89.0\pm0.4 \;(747d)$ & $85.5\pm0.5 \;(358d)$\\
		& FedDC \cite{gao2022feddc} & $83.0\pm0.2\; (280d) $ & $83.1\pm0.6 \;(866)$ & $ 88.0\pm0.6 \;(1985d) $ & $78.0\pm0.9 \;(\infty)$\\
		%&\textbf{FedBCGD (ours)} & $ 81.6\pm0.5(\mathbf{131d}) $ & $85.4\pm0.4(\mathbf{180d})$ & $ 91.2\pm0.5(\mathbf{ 277d}) $ & $79.1\pm0.7(\infty)$\\
		&\textbf{FedBCGD (ours)} & $ \mathbf{84.7\pm0.7}\;(249d) $ & $\mathbf{88.4\pm0.7}\; (292d)$ & $\mathbf{92.1\pm0.3}\; (398d)$ & $\mathbf{87.8\pm0.4}\; (\mathbf{117d})$\\
		& \textbf{FedBCGD+ (ours)} & $83.5 \pm0.3 \;\mathbf{(182d)}  $ & $88.3\pm0.4\;\mathbf{(209d)} $ & $90.3\pm0.5 \;\mathbf{(266d) } $ & $87.1 \pm0.4\;(207d)$\\
		\hline
	\end{tabular}
	\label{table:table 3}
\end{table*}
\vspace{-2mm}
%\end{comment}

\subsection{Theoretical Results of FedBCGD}
%We now compare FedBCGD with FedAvg. From Table  \hyperref[table:1]{1}, for the strongly convex setting, the main components of FedAvg's convergence rate are $\tilde{\mathcal{O}}\big(\frac{\sigma^2+G^2}{\mu S T \epsilon}\big)$, and the main components of FedBCGD's convergence rate are $\tilde{\mathcal{O}}\big(\frac{\sigma^2+G^2}{\mu K T \epsilon}\big)$. That is, our FedBCGD algorithm is equivalent to the case where FedAvg has a sampling rate of $K/M$ per round. From this perspective, FedBCGD is equivalent to reducing the communication consumption per round by reducing the sampling rate per round of FedAvg. From the perspective of communication complexities, FedAvg and FedBCGD have the same communication complexity, and are both $\tilde{\mathcal{O}}\big(\frac{\sigma^2+G^2}{\mu S T \epsilon}d\big)$, where $d$ is the parameter size of model.

%From Table \hyperref[table:1]{1}, we observe that the FedAvg's communication complexity is $\tilde{\mathcal{O}}\big(\frac{\sigma^2+G^2}{\mu S T \epsilon}d\!+\!\frac{\sigma+G}{\mu \sqrt{\epsilon}}d\big)$, while the communication complexity of FedBCGD is  $\tilde{\mathcal{O}}\big(\frac{\sigma^2+G^2}{\mu S T \epsilon}d\!+\!\frac{\sigma+G}{\alpha \mu \sqrt{\epsilon}}d\big)$. It is evident that the momentum of FedBCGD improves the second term of the convergence rate. In our subsequent experiments, FedBCGD demonstrates excellent performance.

\begin{theorem}[FedBCGD]
	For $\beta$-smooth functions $\left\{f_i\right\}$, which satisfy Assumptions 1-5 (see the Appendix for details), the output of FedBCGD has expected error smaller than $\epsilon$ for some values of $\eta, R$, where $R$ denotes the number of communication rounds, $Com$ is the communication complexity (i.e., the product of the number of communication rounds and the floats sent per round) satisfying:\\	
	\textbf{Strongly convex}: $\tilde{\eta}=\frac{\alpha \eta T}{4}$, $\tilde{\eta} \leq \frac{1}{8 \beta}$, and
	\vspace{-2mm}
	\begin{equation*}
		\begin{split}   R=\mathcal{O}\big(\frac{\sigma^2+G^2}{\mu K T  \epsilon}+\frac{\sigma+G}{\alpha \mu \sqrt{\epsilon}}+\frac{\beta}{ \mu}\log \frac{1}{\epsilon}\big),\qquad \\	Com=\mathcal{O}\big(\frac{\sigma^2+G^2}{\mu S T  \epsilon}d+\frac{\sigma+G}{\alpha \mu N \sqrt{\epsilon}}d+\frac{\beta}{ \mu N}\log \frac{1}{\epsilon}d\big),
		\end{split}
	\end{equation*}
	\vspace{-1mm}
	\textbf{Non-convex}: $\tilde{\eta}=\frac{1}{4} \alpha \eta T$, $\tilde{\eta} \leq \frac{1}{16 \beta}$, $F:=f\left(\boldsymbol{x}^0\right)-f^{\star}$,
	\vspace{-1mm}
	\begin{equation*}
		\begin{aligned}
			R=\mathcal{O}\big(\frac{\beta \sigma^2}{TK \epsilon^2}+\frac{\sqrt{\beta} G+\sqrt{\frac{\beta}{T}} \sigma}{\epsilon^{\frac{3}{2}}}+\frac{F\beta}{\epsilon}\big),\qquad\\
			Com=\mathcal{O}\big(\frac{\beta \sigma^2}{TS \epsilon^2}d+\frac{\sqrt{\beta} G+\sqrt{\frac{\beta}{T}} \sigma}{N\epsilon^{\frac{3}{2}}}d+\frac{F\beta}{N\epsilon}d\big).
		\end{aligned}
	\end{equation*}
	%where  % and $F:=f\left(\boldsymbol{x}^0\right)-f^{\star}$
\end{theorem}
From Table \hyperref[table:1]{1}, comparing the second term of communication complexity of FedAvg (i.e., $\mathcal{O}\big(\frac{\sigma+G}{ \mu \sqrt{\epsilon}}d\big)$), the term of FedBCGD is $\mathcal{O}\big(\frac{\sigma+G}{\alpha \mu N \sqrt{\epsilon}}d\big)$, which is $N$ times significantly lower.
As the number of blocks $N$ increases, FedBCGD can achieve a significantly lower communication complexity, and we will verify this in the experimental section (see Figure \hyperref[fig:4]{4}). The momentum parameter $\alpha$ here is equivalent to the server step size, and a larger server step size can accelerate convergence, as pointed out in  \cite{karimireddy2020scaffold}. 

%We can find that the communication complexity of our FedBCGD is mainly influenced by two factors: the variance of stochastic gradient $\sigma$ and client heterogeneity $G$.
%The participation of clients and the number of local iterations also impact the speed of FedBCGD due to two subjective factors.
%In addition, from the analysis of our algorithm convergence, the main reason for the slower convergence of FedBCGD compared to FedAvg is the fewer number of clients participating in each round for each parameter block. To accelerate FedBCGD, we adopt an approach with momentum to increase the client participation rate. 
%Furthermore, we will demonstrate and validate that methods that accelerate FedAvg or mitigate the impact of data heterogeneity are still effective for FedBCGD.
%%% Fanhua Shng

%\subsection{Theoretical Results of of FedBCGDM} 

%Can the introduction of the variance reduction technique and client drift control variate in client local training truly accelerate our FedBCGD algorithm? In the next subsection, we will analyze the convergence properties of our FedBCGDM+ algorithm.\\

%\vspace{-2mm}
\subsection{Theoretical Results of FedBCGD+} 
%Below we present the theoretical result for our FedBCGD+.
%\vspace{-2mm}
\begin{theorem}[FedBCGD+]
	For $\beta$-smooth functions $\left\{f_i\right\}$, which satisfy Assumptions 1-5, the output of FedBCGD+ has expected error smaller than $\epsilon$ for some values of $\eta, R$, where $R$ and $Com$  satisfy:\\	
	\textbf{Strongly convex}: $\tilde{\eta}=\frac{\alpha \eta T}{4}$, $\tilde{\eta} \leq \frac{1}{8 \beta}$, and
	%\vspace{-2mm}
	\begin{equation*}
		R=\mathcal{O}\big(\big( \frac{M}{K} \!+\!\frac{\beta}{\mu}\big)\log \frac{1}{\epsilon}\big),\\
		Com=\mathcal{O}\big(\big(\frac{M}{S}\!+\!\frac{\beta}{\mu N}\big)d \log \frac{1}{\epsilon}\big),
	\end{equation*}
	%\vspace{-1mm}
	\textbf{Non-convex}: $\tilde{\eta}=\frac{1}{4} \alpha \eta T$, $\tilde{\eta} \leq \frac{1}{16 \beta}$, $F:=f\left(\boldsymbol{x}^0\right)-f^{\star}$,
	%\vspace{-1mm}
	\begin{equation*}
		R=\mathcal{O}\big(\frac{\beta F}{\epsilon}\big(\frac{M}{K}\big)^{\frac{2}{3}}\big),\;\;Com=\mathcal{O}\big(\frac{\beta F}{\epsilon}\big(\frac{M}{S}\big)^{\frac{2}{3}}\frac{1}{N}^{\frac{1}{3}}d\big).
	\end{equation*}
	%where ${D}^2\!:=\!\big(\|\boldsymbol{x}^0\!-\!\boldsymbol{x}^{\star}\|^2\!+\!\frac{1}{2 M \beta^2}\! \sum_{i=1}^M\!\|\boldsymbol{c}_i^0\!-\!\nabla\! f_i(\boldsymbol{x}^{\star})\|^2\big)$. % and $F:=f\left(\boldsymbol{x}^0\right)-f^{\star}$
\end{theorem}

The communication complexity of FedBCGD is $\mathcal{O}\big(\frac{\sigma^2+G^2}{\mu S T  \epsilon}d+\frac{\sigma+G}{\alpha \mu N \sqrt{\epsilon}}d+\frac{\beta}{ \mu N}\log \frac{1}{\epsilon}d\big)$ in the strongly convex setting. The main influence on the communication complexity is determined by the two parameters, $G$ (client heterogeneity) and $\sigma$ (noise of stochastic gradients). FedBCGD+ resolves these issues, and can achieve the communication complexity of   $\mathcal{O}\big(\big(\frac{M}{S}\!+\!\frac{\beta}{\mu N}\big)d \log \frac{1}{\epsilon}\big)$. When $N$ = $\sqrt{{\beta}/{\mu}}$, and its communication complexity is $\mathcal{O}\left(\left(\frac{M}{S}+\sqrt{\frac{\beta}{\mu}}\right) d \log \frac{1}{\epsilon}\right)$, which significantly improves the best-known result (see Table \hyperref[table:1]{1} for details). When $\sigma$ = 0, the communication complexity of FedBCGD is also better than that of SCAFFOLD, $\mathcal{O}\big(\big(\frac{M}{S}+\frac{\beta}{\mu}\big) d \log \frac{1}{\epsilon}\big)$. Without client sampling ($S$ = $M$), the communication complexity of FedBCGD+ is $\mathcal{O}\big(\sqrt{\frac{\beta}{\mu}} d \log \frac{1}{\epsilon}\big)$, which is much better than that of FedLin \cite{mitra2021linear}, $\mathcal{O}\big(\frac{\beta}{\mu} d \log \frac{1}{\epsilon}\big)$. In the non-convex setting, the communication complexity of FedBCGD+ is $\mathcal{O}\big(\frac{\beta F}{\epsilon}\big(\frac{M}{S}\big)^{{2}/{3}}{N^{-{1}/{3}}}d\big)$, which also is the best-known result (see Table \hyperref[table:1]{1}). Without client sampling, the communication complexity of FedBCGD+ is $\mathcal{O}\big(\frac{\beta F}{\epsilon}{N^{-{1}/{3}}}d\big)$, which is much better than that of CE-LSGD \cite{patel2022towards}, $ \mathcal{O}\big(\frac{\beta F}{\epsilon}d\big)$. As the number of blocks $N$ increases, FedBCGD+ can also achieve a significantly lower communication complexity.

%, but we cannot surface the superiority of momentum in theory

\section{Experiments}
In this section, we conduct various experiments for  convex and non-convex problems, and more results are reported in the Appendix.

\begin{figure*}[htbp]
	\centering
	\begin{minipage}[b]{0.245\textwidth}
		\centering
		\subcaptionbox{LeNet-5, CIFAR10}{\includegraphics[width=\textwidth]{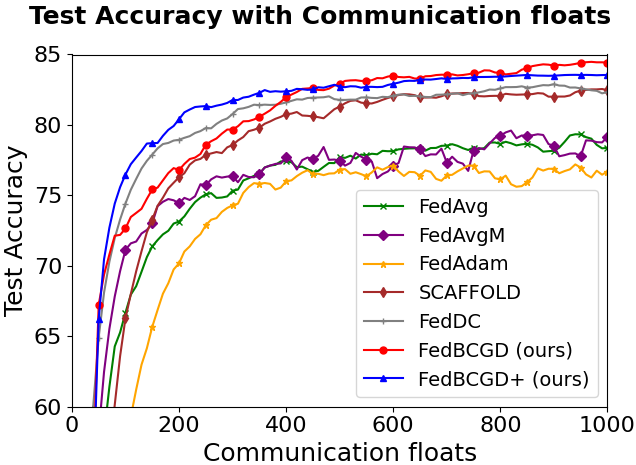}}
	\end{minipage}
	\hfill
	\begin{minipage}[b]{0.245\textwidth}
		\centering
		\subcaptionbox{LeNet-5, CIFAR100}{\includegraphics[width=\textwidth]{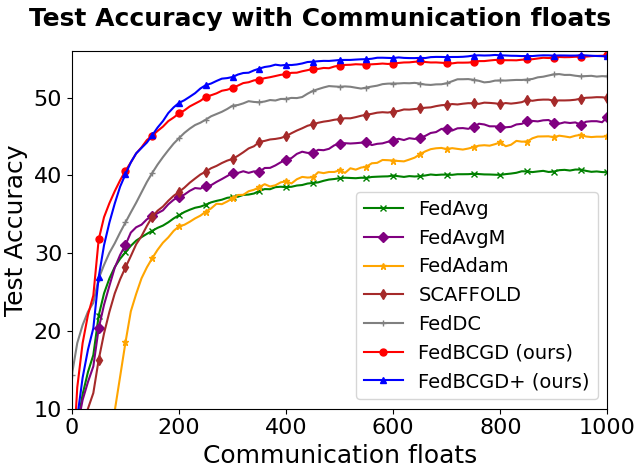}}
	\end{minipage}
	\hfill
	\begin{minipage}[b]{0.245\textwidth}
		\centering
		\subcaptionbox{VGG-11,CIFAR10}{\includegraphics[width=\textwidth]{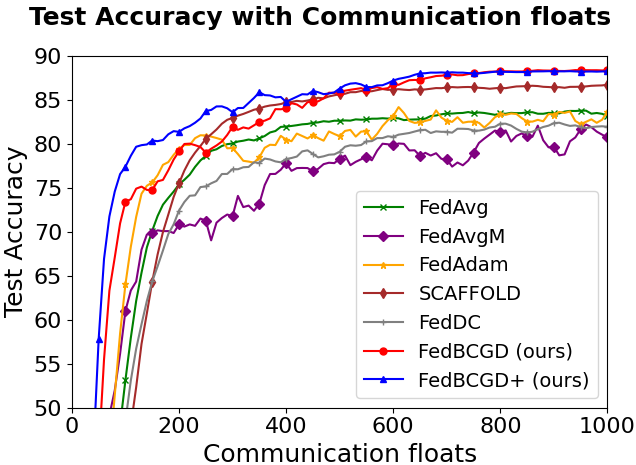}}
	\end{minipage}
	\hfill
	\begin{minipage}[b]{0.245\textwidth}
		\centering
		\subcaptionbox{VGG-11, CIFAR100}{\includegraphics[width=\textwidth]{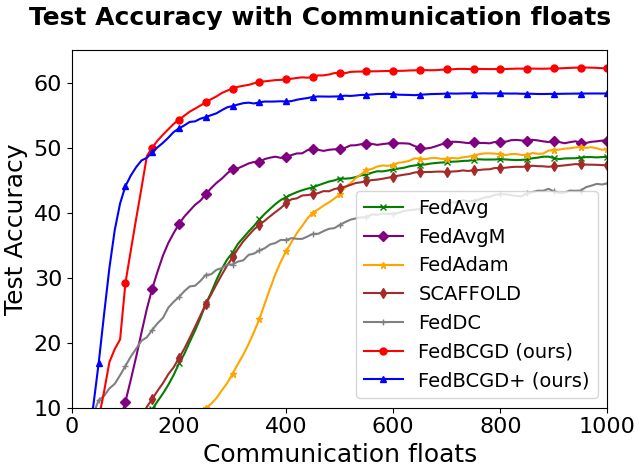}}
	\end{minipage}
	
	\vspace{0.2cm} % 调整上下图之间的垂直间距
	
	\begin{minipage}[b]{0.245\textwidth}
		\centering
		\subcaptionbox{ResNet-18, CIFAR10}{\includegraphics[width=\textwidth]{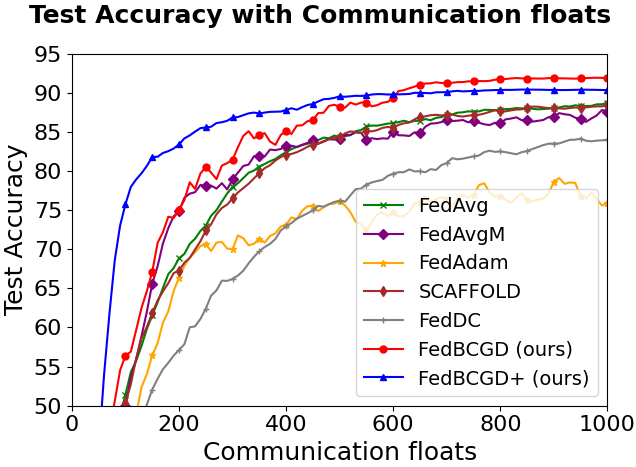}}
	\end{minipage}
	\hfill
	\begin{minipage}[b]{0.245\textwidth}
		\centering
		\subcaptionbox{ResNet-18,CIFAR100}{\includegraphics[width=\textwidth]{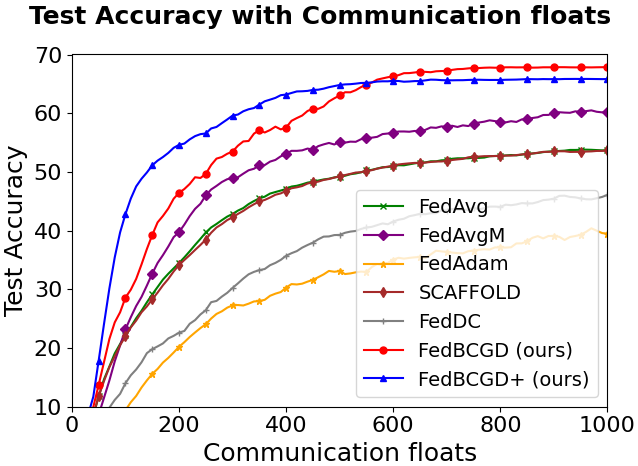}}
	\end{minipage}
	\hfill
	\begin{minipage}[b]{0.245\textwidth}
		\centering
		\subcaptionbox{VGG-19, CIFAR10}{\includegraphics[width=\textwidth]{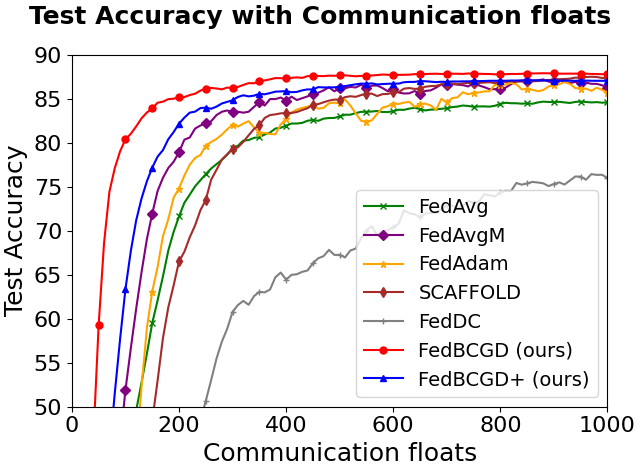}}
	\end{minipage}
	\hfill
	\begin{minipage}[b]{0.245\textwidth}
		\centering
		\subcaptionbox{VGG-19, CIFAR100}{\includegraphics[width=\textwidth]{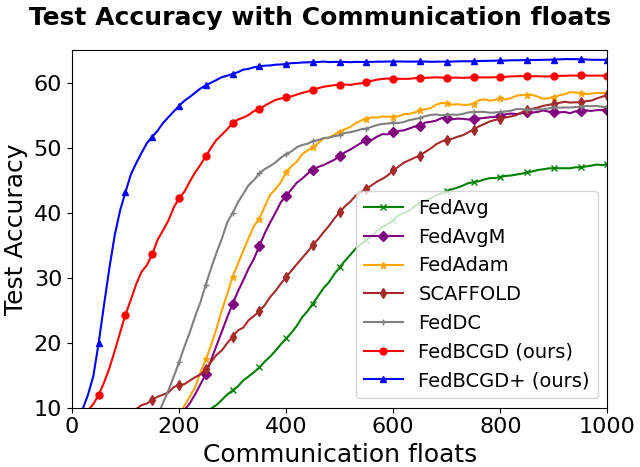}}
	\end{minipage}
	\vspace{-3mm}
	\caption{The convergence comparison of our FedBCGD and FedBCGD+, and other baselines on the CIFAR10 and CIFAR100 datasets with different neural network architectures, where, in 100 clients, partial (10\%) clients are used, $\rho\!=\!0.6$.}
	\label{figure:figure 4}
\end{figure*}

\subsection{Experimental Settings and Baselines}
\textbf{{Datasets:}} We evaluate our algorithms on the CIFAR10 \cite{krizhevsky2009learning}, CIFAR100 \cite{krizhevsky2009learning}, Tiny ImageNet \cite{le2015tiny} and EMNIST datasets. We set up a total of 100 clients in the FL experiment with a participation rate of 10\%. For the non-IID data setup, we model data heterogeneity by sampling label ratios $\rho$ from a Dirichlet distribution. \\
\textbf{{Models:}} To test the robustness of our algorithms, we use standard classifiers (including LeNet-5 \cite{lecun2015lenet}, VGG-11, VGG-19 \cite{simonyan2014very}, and ResNet-18 \cite{he2016deep}), Vision Transformer (ViT-Base) \cite{dosovitskiy2020image}.
%, and logistic regression \cite{menard2002applied}
We divided the parameters of the model into 5 blocks or more blocks and provide the detailed parameter block division of the model in the Appendix.\\
\textbf{{Methods:}} We compare FedBCGD and FedBCGD+ with many
SOTA FL baselines, including FedAvg \cite{mcmahan2017communication}, SCAFFOLD \cite{karimireddy2020scaffold}, FedAvgM \cite{hsu2019measuring}, FedDC \cite{gao2022feddc} , FedAdam \cite{reddi2020adaptive}, and TOP-k \cite{Aji_2017}, FedPAQ \cite{reisizadeh2020fedpaq}.
\textbf{Hyper-parameter Settings:}
The initial learning rate is searched in $\{0.01,0.03,0.05,0.1,0.2,0.3\}$, with a  decay of 0.998 and a weight decay of 0.001 for each round.

\subsection{Results on Non-Convex Problems}
\textbf{Results on Convolutional Neural Network:}
From Tables \hyperref[table:table 2]{2} and \hyperref[table:table 3]{3}, and Figure \hyperref[figure:figure 4]{3}, we have the following observations:
(i) Compared to FedAvg and its accelerated algorithms, FedBCGD significantly reduces the communication floats per round, converges faster, and achieves more robust final model performance. In the experiment of LeNet-5 on CIFAR100, FedBCGD ($77d$) achieve   7.3$\times$ speedup to reach 40\% accuracy, compared to FedAvg ($558d$).
(ii) FedBCGD+ further improves the convergence speed by client drift control and variance reduction, accelerating FedBCGD training process in experiments. In the experiment of ResNet-18 on CIFAR100, FedBCGD+ ($154d$) achieves  1.8$\times$ speedup to reach 54\% accuracy, compared to FedBCGD ($277d$). However, in terms of the final testing accuracy, it does not outperform FedBCGD. 
\begin{figure}[h]
	\centering
	\includegraphics[width=0.3\textwidth]{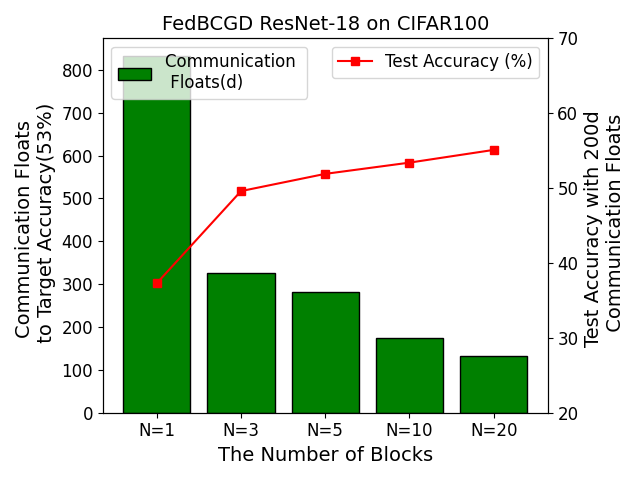}
	\vspace{-4mm}
	\caption{The acceleration comparison of FedBCGD with different numbers of blocks.}
	\label{fig:4}
\end{figure}
\begin{figure}[H]
	\centering
	\subcaptionbox{FedBCGD, CIFAR10}{\includegraphics[width=0.235\textwidth]{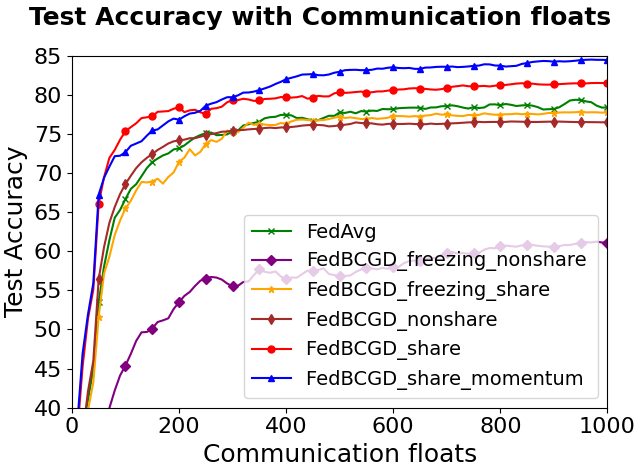}}
	\subcaptionbox{FedBCGD, CIFAR100}{\includegraphics[width=0.235\textwidth]{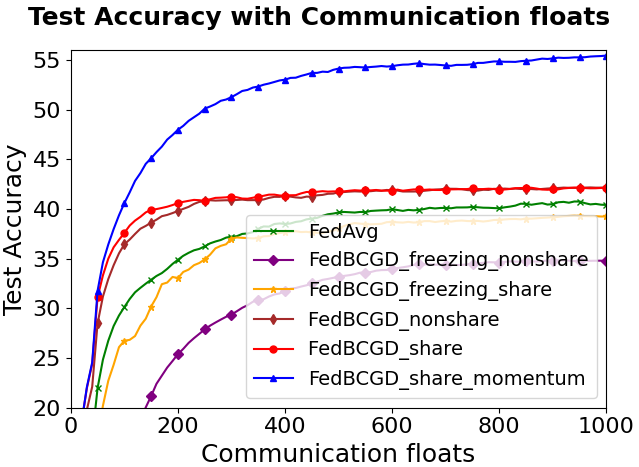}}
	\vspace{-3mm}	
	\caption{Accuracy  comparison of FedBCGD with LeNet-5 on CIFAR10 (a) and CIFAR100 (b), where heterogeneity is $\rho\!=\!0.6$. FedBCGD\_freezing\_nonshare is updated by using the local freezing parameter algorithm without the shared block. FedBCGD\_freezing\_share is FedBCGD\_freezing algorithm with shared parameters. FedBCGD\_nonshare  trains all parameters locally and only transmits parameter blocks  without shared parameters. FedBCGD\_share  has shared parameters. FedBCGD\_share\_momentum (i.e., FedBCGD)  has momentum acceleration. }
	\vspace{-2mm}
	\label{fig:fig1}
\end{figure}
\vspace{-2mm}
This means that FedBCGD+ has a faster convergence speed, requiring less communication floats at the specified accuracy, while the higher accuracy of our FedBCGD algorithm ultimately means that it has better generalization ability. And the generalization ability of our FedBCGD framework is better than those of other algorithms, e.g., FedAvg.
(iii) The final accuracy of FedBCGD is much higher than that of Centralised SGD, which means that our FedBCGD has better generalization performance. That is, FedBCGD and FedBCGD+ can jump from a poor local minimum and converge to sharp local minima.

Figure \hyperref[fig:4]{4} compares the effects of different block numbers under the same settings. When the number of blocks is 1, it degenerates into the FedAvgM algorithm. At the specified testing accuracy 53\%, when the number of blocks is 20, our FedBCGD algorithm requires the least communication floats. The FedBCGD algorithm with  20 blocks achieves the highest accuracy with the same communication floats $200d$.
As the number of blocks increases, the acceleration effect of the FedBCGD algorithm becomes more obvious.

From Figure \hyperref[fig:fig1]{5}, we can observe that freezing parameters in local training will cause client parameters to drift (purple line), resulting in poor performance. In addition, uploading parameters with shared parameters can improve convergence speed and final performance of the model (red line). Adding momentum compensation to client aggregation does accelerate convergence significantly (blue line).
\vspace{-1mm}
\begin{table}[th]
	\caption{Comparison of  each algorithm on CIFAR100 and CIFAR10.  Heterogeneity is  $\rho\!=\!0.1$, total communication floats are $1000d$, and the number of blocks in ResNet-18 is $N=5$.}
	\label{tab:cross-device}
	\vspace{-2mm}
	\setlength{\tabcolsep}{5pt}	\begin{tabular}{llccc}
		\hline & $\rho=0.1$ &  CIFAR100 (45\%)  &  CIFAR10 (78\%) \\
		\hline
		& FedAvg \cite{mcmahan2017communication}   & $45.8\pm0.3\;(741d)$& $78.1\pm0.4\;(952d)$ \\
		& FedAvgM \cite{hsu2019measuring} &  $ 48.3 \pm0.6\;(769d)$  & $ 78.6 \pm0.8\;(997d)$  \\
		& FedAdam \cite{reddi2020adaptive}  &  $49.9\pm0.5\;(610d)$   & $ 71.4\pm1.1\;(\infty)$  \\
		& SCAFFOLD \cite{karimireddy2020scaffold}  &  $44.3\pm0.3\;(\infty)$  & $76.3\pm1.4\;(\infty)$ \\
		& FedDC \cite{gao2022feddc}  & $ 46.6\pm0.8\;(278d)$  & $ 79.1\pm0.8\;(948d)$ \\
		%& \textbf{FedBCGD (ours)}   & $ 48.8\pm0.7(200d) $  & $ 84.3\pm0.8(\mathbf{199d}) $ \\
		& \textbf{FedBCGD (ours)}  & $59.5\pm0.3\;(\mathbf{147d}) $ &   $\mathbf{86.2\pm0.9}\;(\mathbf{212d}) $ \\
		& \textbf{FedBCGD+ (ours)}   & $\mathbf{59.9\pm0.4}\;(200d)$ & $80.2\pm1.3\;(768d)$ \\
		\hline
	\end{tabular}
	\label{table:4}
\end{table}
\vspace{-1mm}
\begin{table}[h]	\caption{The test accuracy comparison of each algorithm with ViT-Base on CIFAR100 and Tiny ImageNet. Heterogeneity is  $\rho\!=\!0.6$, total communication floats are $100d$, $N=6$.}
	\label{tab:cross-device}
	\vspace{-2mm}
	\setlength{\tabcolsep}{4pt}
	\begin{tabular}{llccc}
		\hline 
		& $\rho=0.6$ &  CIFAR100 \;(88\%) & Tiny Imagenet\; (70\%) \\
		\hline
		& Centralised SGD  & $81.5\pm0.3$  &$76.7\pm0.2$   \\
		& FedAvg \cite{mcmahan2017communication}  & $90.4\pm0.1\;(24d)$  &$71.2\pm0.1\;(67d)$   \\
		& FedAvgM \cite{mcmahan2017communication}  & $88.7\pm0.3\;(32d)$ &$76.7\pm0.4\;(10d)$  \\
		& FedAdam \cite{reddi2020adaptive}  & $87.6\pm0.2\;(\infty)$ &$ 65.5\pm0.6\;(\infty)$ \\
		& SCAFFOLD \cite{karimireddy2020scaffold}  & $88.2\pm0.3\;(88d)$ &$56.8\pm1.1\;(\infty)$ \\
		& FedDC \cite{gao2022feddc}&  $85.8\pm0.4\;(25d)$  &$55.0\pm1.2\;(\infty)$  \\
		%& \textbf{FedBCGD (ours)} & $90.7\pm0.4(18d)$   &$81.3\pm0.2(5d)$ \\
		& \textbf{FedBCGD (ours)} & $\mathbf{92.0\pm0.2}\;(\mathbf{7d}) $ &$\mathbf{83.5\pm0.2}\;(5.8d)$  \\
		& \textbf{FedBCGD+ (ours)}  & $90.6\pm0.3 \;(14d)$ &$81.3\pm0.2\;(\mathbf{4.6d})$  \\
		\hline
	\end{tabular}
	\label{table:5}
\end{table}

From results in Table \hyperref[table:4]{4}, we compare the convergence speed of our algorithms and baseline algorithms under high levels of data heterogeneity. It can be observed that when data heterogeneity is high (e.g., $\rho\!=\!0.1$), FedAvg converges slowly and struggles to reach the optimal point. In contrast, our algorithms consistently converge and achieve better model generalization. Moreover, under high data heterogeneity, FedBCGD+ slightly outperforms FedBCGD, demonstrating the effectiveness of the variance control strategy in our FedBCGD+ algorithm. 
%We conducted many experiments on the ResNet-18 network due to its stable convergence characteristics.

In our experiments we  get a phenomenon that our algorithm, FedBCGD and FedBCGD+,  may generalizes better than Centralized SGD when the client data is not highly heterogeneous. The same phenomenon was also found in the literature \cite{gu2023and,li2023effectiveness}. For highly non-convex problems, gradient decent and SGD methods are usually prone to fall into  local minima, whereas the distributed methods  local SGD  are  more prone to jump out of the local and sharp minimum and usually have better generalization ability \cite{gu2023and}.

In Table \hyperref[table:6]{6}, the FedBCGD algorithm outperforms  TOP-k and FedPAQ in terms of convergence speed and final generalization accuracy. The convergence can be further accelerated when the quantization strategy of QSGD is added to the block of FedBCGD.

\begin{table}[h]	\caption{The test accuracy comparison of each algorithm with LeNet-5 on CIFAR100 and CIFAR10. Here, the heterogeneity is  $\rho\!=\!0.6$, total communication floats are $200d$, $N=5$.}
	\label{tab:cross-device}
	\vspace{-2mm}
	\setlength{\tabcolsep}{1.5pt}
	\begin{tabular}{llccc}
		\hline 
		& $\rho=0.6$ &  CIFAR100 (40\%) & CIFAR10 (70\%) \\
		\hline
		%& Centralised SGD  & $81.5\pm0.3$  &$76.7\pm0.2$   \\
		& FedAvg \cite{mcmahan2017communication}  & $35.4\pm0.1\;(\infty)$  &$73.2\pm0.1\;(133d)$   \\
		& TOP-k \cite{Aji_2017} & $42.2\pm0.5\;(112d)$ &$74.5\pm0.4\;(92d)$  \\
		& FedPAQ \cite{reisizadeh2020fedpaq}  & $43.3\pm0.2\;(110d)$ &$75.2\pm0.4\; (121d)$ \\
		& \textbf{FedBCGD (ours)} & $48.7\pm0.2\;(91d) $ &$\mathbf{77.2\pm0.2}\;(65d)$  \\
		& \textbf{FedBCGD+ (ours)}  & $49.6\pm0.3 \;(89d)$ &$80.6\pm0.2\;(57d)$  \\
		& \textbf{QSGD\cite{alistarh2017qsgd}+FedBCGD (ours)}  & $52.2\pm0.4 \;(61d)$ &$82.6\pm0.1\;(32d)$  \\
		& \textbf{QSGD\cite{alistarh2017qsgd}+FedBCGD+ (ours)}  & $\mathbf{53.1\pm0.2} \;(\mathbf{56d})$ &$\mathbf{83.2\pm0.3}\;(\mathbf{29d})$  \\
		\hline
	\end{tabular}
	\label{table:6}
\end{table}

\vspace{-1mm}
\begin{figure}[H]
	\centering
	\subcaptionbox{ViT-Base, CIFAR100}{\includegraphics[width=0.236\textwidth]{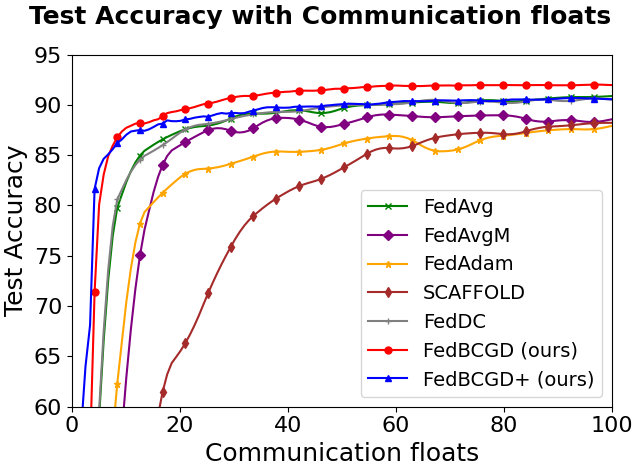}}
	\subcaptionbox{ViT-Base, Tiny ImageNet}{\includegraphics[width=0.236\textwidth]{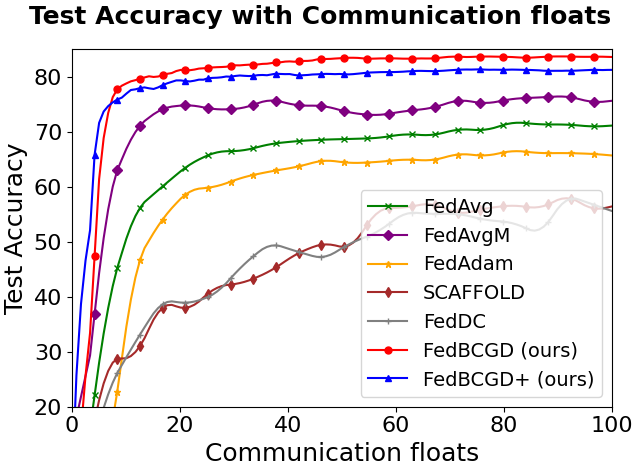}}
	\vspace{-3mm} 
	\caption {The test accuracy varies with the communication floats with ViT-Base on the CIFAR100 and Tiny ImageNet datasets, where $E\!=\!1$ and $\rho\!=\!0.6$, $N=6$.}
	\label{fig:6}
\end{figure}
\textbf{Results on Vision Transformer:}
To verify the effectiveness of our algorithm on large models, we adopt the most classic ViT-Base model on the Tiny ImageNet and CIFAR100 datasets. For the initialization of the model, we used the pretrained model downloaded from the official website. We divide the ViT-Base model into six parameter blocks. From the experimental results in Table \hyperref[table:5]{5} and Figure \hyperref[fig:6]{6}, we can observe that our FedBCGD algorithm can achieve the best results on the CIFAR100 dataset, and has more than 3$\times$ faster convergence speed, compared to FedAvg. The FedBCGD algorithm can achieve the best results on the Tiny ImageNet dataset, and attains more than 11.5$\times$ faster convergence speed. This can verify that FedBCGD can achieve excellent convergence speed on both Vision Transformer models and big datasets.

\textbf{Effectiveness of $\lambda$:} We tested FedBCGD using ResNet-18 on CIFAR100 dataset with momentum parameter $\lambda$ taking the values of \{0.4, 0.5, 0.6, 0.7, 0.8, 0.9\} and $\rho=0.6$. 
We note that setting $\lambda$ too small or too large impairs the convergence and generalization ability of FedBCGD. As shown in Figure \hyperref[fig:7]{7}, when $\lambda$ is relatively small, with $\lambda$ = 0.4, the FedBCGD algorithm converges quickly, but the final generalization is not good. When we enlarge the value of $\lambda$, $\lambda$ = 0.8, the convergence is slower but the final generalization is good. Empirically, we find the best performance is achieved when the $\lambda$ is set to around 0.8.
\begin{figure}[H]
	\centering
	\subcaptionbox{ Test accuracy}{\includegraphics[width=0.235\textwidth]{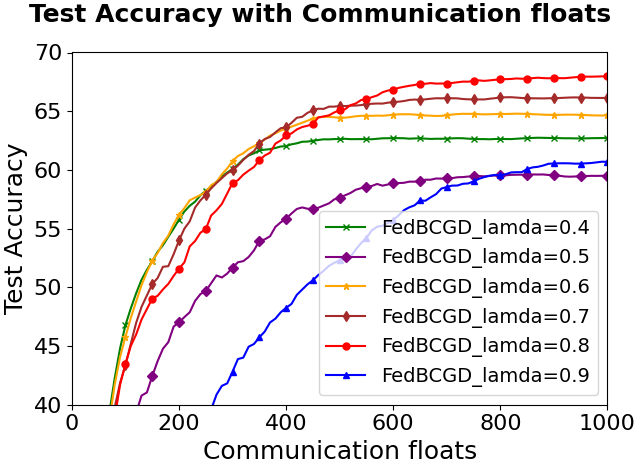}}
	\subcaptionbox{ Train loss}{\includegraphics[width=0.235\textwidth]{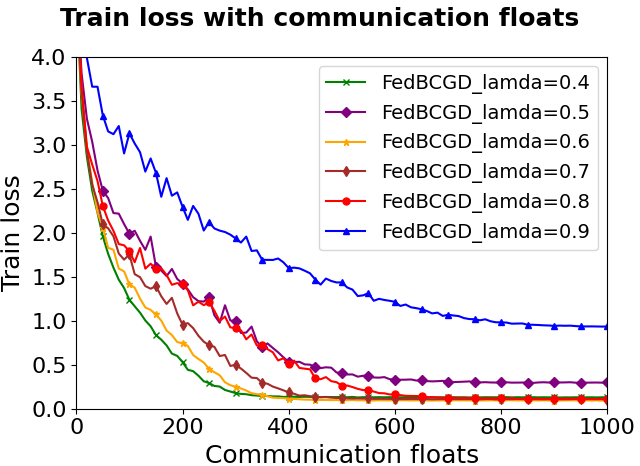}}
	\vspace{-3mm}
	\caption{Test accuracy (a) and training loss (b) with ResNet-18 on CIFAR100, where $E\!=\!5$ and $\rho\!=\!0.6$, $N=5$.}
	\label{fig:7}
\end{figure}

%\begin{comment}
\subsection{Results on Convex Problems}
%In the above experiment, the FedBCGD+ algorithm is not the best algorithm, which makes  a gap between complex deep learning experiments and theory.
%We use a logistic regression model to verify the consistency between FedBCGD+'s practice and theory results.
%The following two experiments are used to verify that the experimental results of FedBCGDM+ can match its theoretical results.
We conducted the classification tests on the EMNIST (byclass) dataset on classical logistic regression problems:
\vspace{-2mm}
\begin{align}
	f(x)=\frac{1}{N} \sum_{i=1}^N \log \left(1+\exp \left(-b_i a_i^{\top} x\right)\right)+\frac{\lambda}{2}\|x\|^2,
\end{align}
where $a_i \in \mathbb{R}^d$ and $b_i \in\{-1,+1\}$ are the data samples, and $N$ is their total number. We set the regularization parameter $\lambda=10^{-4} L$, where $L$ is the smoothness constant.

\begin{figure}[H]
	\centering
	\subcaptionbox{Train loss}{\includegraphics[width=0.23\textwidth]{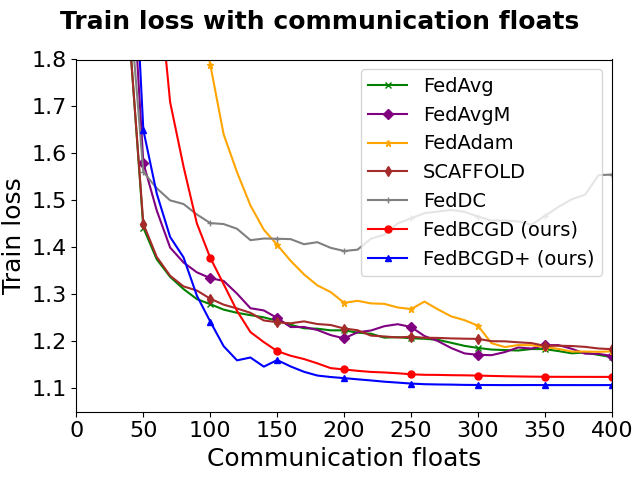}}
	\subcaptionbox{Test loss}{\includegraphics[width=0.23\textwidth]{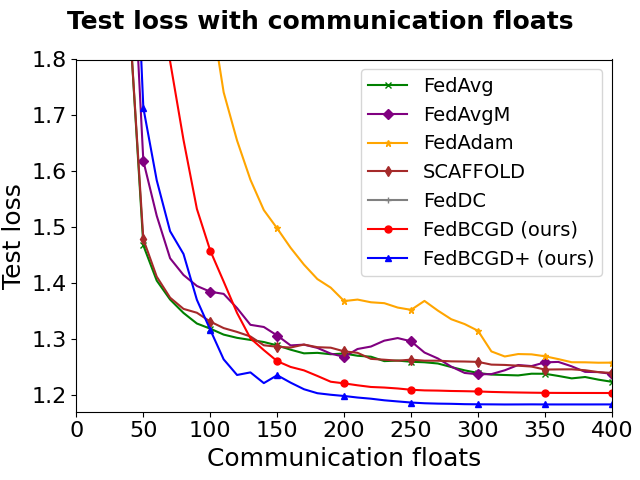}}
	\vspace{-3mm}
	\caption{Logistic regression with $E\!=\!1$ and $\rho\!=\!0.1$, $N=5$.}
	\label{fig:8}
\end{figure}

From Figure \hyperref[fig:8]{8} (a,b), we observe that our FedBCGD and FedBCGD+ algorithms demonstrate faster convergence speed. Particularly, under the strong convexity, our FedBCGD+ algorithm exhibits even faster convergence compared to our FedBCGD, which aligns with our theoretical analysis.

%\end{comment}

\begin{comment}
	\textbf{ERM with Non-Convex Loss:}
	We also apply our algorithms to solve the regularized Empirical Risk Minimization (ERM) problem with non-convex sigmoid loss:
	\vspace{-2mm}
	\begin{align}
		\min _{x \in \mathbb{R}^d} \frac{1}{n} \sum_{i=1}^n f_i(x)+\frac{\lambda}{2}\|x\|^2,
	\end{align}
	where $f_i(x)=1 /\left[1+\exp \left(b_i a_i^{\top} x\right)\right]$.  Here, we consider binary classification on EMNIST. Note that we only consider classifying the first class in EMNIST.
	
	From the results of the ERM problem in Figure \hyperref[fig:7]{8} (b), we observe that our algorithms exhibit much faster convergence speeds than other algorithms. Moreover, in the case of high client data heterogeneity, FedBCGD+ demonstrates faster convergence than FedBCGD, which is consistent with our theoretical results.
\end{comment}
\section{Conclusion}
This paper proposed the first federated block coordinate gradient descent method for horizontal FL. Moreover, we presented an accelerated version by using variance reduction and client parameter block drift control. In particular, we analyzed the convergence properties of the proposed algorithms, which show that our algorithms have significantly lower communication complexities than existing methods, and they also attain the best-known convergence rates for both convex and non-convex problems. Various experimental results verified our theoretical results and effectiveness of all the proposed algorithms. In the future, it is worthwhile to pay attention to how to more rationally divide model into blocks and how to choose the optimal parameter block to upload for clients.

%\newpage
{
	\section*{Acknowledgments}
	This work was supported by the National Natural Science Foundation of China (No. 62276182), National Key Research and Development Program of China (No. 2023YFF0906204) and Peng Cheng Lab Program (No. PCL2023A08).}	

%\vfill
%\eject 
%\vfill\eject

\balance
\bibliographystyle{ACM-Reference-Format}

\bibliography{sample-base}
\clearpage
%\setcounter{page}{1}

%\maketitlesupplementary
%\usepackage{subcaption}
\onecolumn
%\section{Rationale}
\label{sec:rationale}

\section{Appendix A: Basic Assumptions and Notations}

\begin{algorithm}[tb]
	\caption{FedBCGD}
	\begin{algorithmic}[1] %[1] enables line numbers
		\STATE $\textbf{Initialize } \boldsymbol{x}_{i}^{0,0}=\boldsymbol{x}^{i n i t}$, $\forall i \in [M]$.
		\STATE \textbf{Divide} the model parameters $\boldsymbol{x}$ into $N\!+\!1$ blocks.
		\FOR{$r=0,...,R$}
		\STATE{\textbf{Client:}}
		\STATE{$\textbf{Sample} \text{ clients } \mathcal{ S} \subseteq\{1, \ldots, M\}$,  $|\mathcal{S}|=N\cdot K$;}
		\STATE{ \textbf{Divide} \text{the sampled clients into} $N$ \text{client blocks};}
		\STATE $\textbf{Communicate }(\boldsymbol{x}^r) \text { to all clients } i \in \mathcal{S}$;
		\FOR{$j=1, \ldots, N$ client blocks in parallel}	
		\FOR{$k=1, \ldots, K$ clients in parallel}	
		\FOR{$t=1, \ldots, T$ local update}	
		\STATE \!Compute batch gradient $\nabla\! f_{k, j}\big(\boldsymbol{x}_{k, j}^{r,t} ; \zeta\big)$,
		\STATE$\boldsymbol{x}_{k, j}^{r, t+1}=\boldsymbol{x}_{k, j}^{r, t}-\eta \nabla f_{k, j}\big(\boldsymbol{x}_{k, j}^{r, t} ; \zeta\big)$;				
		\ENDFOR
		\STATE Send $\boldsymbol{x}_{k, j,(j)}^{r, T}$, $\boldsymbol{x}_{k, j,s}^{r, T}$ to server;
		\ENDFOR
		\ENDFOR
		\STATE \textbf{Server:}
		\FOR{$j=1, \ldots, N$ Blocks in parallel}	
		\STATE Block $j$ computes,
		\STATE$\boldsymbol{x}_{(j)}^r=\frac{1}{K} \sum_{k=1}^K \boldsymbol{x}_{k, j,(j)}^{r, T}$;
		$v_{(j)}^r=\lambda v_{(j)}^{r-1}+\boldsymbol{x}_{(j)}^{r}-\boldsymbol{x}_{(j)}^{r-1}$;
		\STATE$\boldsymbol{x}_{(j)}^r=\boldsymbol{x}_{(j)}^{r}+v_{(j)}^{r},$
		\ENDFOR
		\STATE$\boldsymbol{x}_{s}^{r}=\frac{1}{NK}\sum_{j=1}^N\sum_{k=1}^K\boldsymbol{x}_{k, j,s}^{r, T}$;	$v_{s}^r=\lambda v_{s}^{r}+\boldsymbol{x}_{s}^{r-1}-\boldsymbol{x}_{s}^{r-1}$;
		\STATE$\boldsymbol{x}_{s}^r=\boldsymbol{x}_{s}^{r}+v_{s}^{r} ;\boldsymbol{x}^r=\big[\boldsymbol{x}_{(1)}^{r \top}, \ldots, \boldsymbol{x}_{(N)}^{r \top},\boldsymbol{x}_{s}^{r \top}\big]^{\top}$;
		\STATE $\boldsymbol{v}^r=\big[\boldsymbol{v}_{(1)}^{r \top}, \ldots, \boldsymbol{v}_{(N)}^{r \top},\boldsymbol{v}_{s}^{r \top}\big]^{\top}$;
		\ENDFOR
	\end{algorithmic}
	\label{algorithm:1}
\end{algorithm}
\subsection{Basic Assumptions }
Before giving our theoretical results, we first present the common assumptions.
\begin{Assumption}[Convexity] $f_i$ is $\mu$-strongly-convex for all $i \in[M]$, i.e.,
	\begin{equation}
		f_i(\boldsymbol{y}) \geq f_i(\boldsymbol{x})+\left\langle\nabla f_i(\boldsymbol{x}), \boldsymbol{y}-\boldsymbol{x}\right\rangle+\frac{\mu}{2}\|\boldsymbol{y}-\boldsymbol{x}\|^2
	\end{equation}
	
	for all $\boldsymbol{x}, \boldsymbol{y}$ in its domain and $i \in[M]$. We allow $\mu=0$, which corresponds to general convex functions.
\end{Assumption}

\begin{Assumption}[Smoothness]
	The gradient of the loss function is Lipschitz continuous with constant $\beta$, for all $\boldsymbol{x}_1, \boldsymbol{x}_2 \in \mathbb{R}^d$
	
	\begin{equation}
		\left\|\nabla f\left(\boldsymbol{x}_1\right)-\nabla f\left(\boldsymbol{x}_2\right)\right\| \leq \beta\left\|\boldsymbol{x}_1-\boldsymbol{x}_2\right\|.
	\end{equation}
\end{Assumption}

\begin{Assumption} Let $\zeta$ be a mini-batch drawn uniformly at random from all samples. We assume that the data is distributed so that, for all $\boldsymbol{x} \in \mathbb{R}^d$
	\begin{equation}
		\mathbb{E}_{\zeta \mid \boldsymbol{x}}\left[\nabla f_i\left(\boldsymbol{x} ; \zeta\right)\right]=\nabla f_i(\boldsymbol{x}).
	\end{equation}
	We also can get:
	
	\begin{equation}
		\mathbb{E}_{\zeta \mid \boldsymbol{x}}\left[\left\|\nabla f_i\left(\boldsymbol{x} ; \zeta_i\right)-\nabla f_i(\boldsymbol{x})\right\|^2\right] \leq \sigma^2.
	\end{equation}
\end{Assumption}

\begin{Assumption}[Bounded heterogeneity]
	The dissimilarity of $f_i(\boldsymbol{x})$ and $f(\boldsymbol{x})$ is bounded as follows:
	\begin{equation}
		\frac{1}{M} \sum_{i=1}^M\left\|\nabla f_i(\boldsymbol{x})-\nabla f(\boldsymbol{x})\right\|^2 \leq G^2.
	\end{equation}
\end{Assumption}

\begin{Assumption}[Stochastic gradient smoothness]
	The gradient of the loss function is Lipschitz continuous with constant $\beta$, for all $\boldsymbol{x}_1, \boldsymbol{x}_2 \in \mathbb{R}^d$
	
	\begin{equation}
		\left\|\nabla f\left(\boldsymbol{x}_1;\zeta\right)-\nabla f\left(\boldsymbol{x}_2;\zeta\right)\right\| \leq \beta\left\|\boldsymbol{x}_1-\boldsymbol{x}_2\right\|.
	\end{equation}
\end{Assumption}

Assumption 2 bounds the variance of stochastic gradients, which is common in stochastic optimization analysis \cite{bubeck2015convex}. Assumption 3 bounds the gradient difference between global and local loss functions, which is a widely-used approach to characterize client heterogeneity in federated optimization literature \cite{li2020federated,reddi2020adaptive}. Assumption 5 is a necessary assumption in stochastic gradient noise reduction, an assumption that is used only in the proof of the convergence speed of the FedBCGD+ algorithm.

\subsection{Notation}

We first define the notations to be used in analyzing the convergence properties of our algorithms. 

1. $\boldsymbol{x}^r$ is the $r$ communication rounds global model.

2. $\boldsymbol{x}_{(j)}^r$ is the $j$-th block of $\boldsymbol{x}$, so that $\boldsymbol{x}^r=\left[\boldsymbol{x}_{(1)}^{r \top}, \ldots, \boldsymbol{x}_{(N)}^{r \top}\right]^{\top}$. Note that $\boldsymbol{x}_{(j)}^r$ is a virtual vector. It is realized at a hub $j$ every $r$ iterations, but we will study the evolution of this virtual vector in every iteration.

3. $\boldsymbol{x}_{k, j}^r \in \mathbb{R}^{d}$ are the local versions of the coordinates of the weight vector $\boldsymbol{x}_{(j)}^r$ that each client $k$ if hub $j$ updates.

4. $\boldsymbol{x}^{\star}$ is is the minimum value of the function $f(\boldsymbol{x})$.

5. $\boldsymbol{x}_{k, j,(j)}$ is the $j$-th block of $\boldsymbol{x}_{k, j}$ at client $k$ in silo $j$, so that $\boldsymbol{x}_{(j)}=\frac{1}{K} \sum_{k=1}^K \boldsymbol{x}_{k, j,(j)}$.

6. $\boldsymbol{y}_{k, j}^{r, t}$ is the local parameter vector that client $j$ in silo $k$ at iteration $t$.

7. $\nabla_{(j)} f_{k, j}\left(\boldsymbol{y}_{k, j} ; \zeta\right)$ is the partial derivative of $f(\boldsymbol{x})$ with respect to coordinate block $j$, computed at client $k$ in silo $j$ using the coordinates and rows at client $k$ corresponding to minibatch $\zeta$.

8. $\boldsymbol{G}^r=\left[\left(\boldsymbol{G}_{(1)}^r\right)^{\top}, \ldots,\left(\boldsymbol{G}_{(N)}^r\right)^{\top}\right]^{\top}$, where $\mathbf{G}_{(j)}^r=\frac{1}{K} \sum_{k=1}^K \sum_{t=1}^T \nabla_{(j)} f_{k, j}\left(y_{k, j}^{r, t} ; \zeta\right)$.

It should be noted that components on $\boldsymbol{x}$, i.e., $\boldsymbol{x}_{(j)}$ are realized every $T$ iterations when the hubs communicate with clients and with other hubs, but we will study the evolution of these virtual vectors at each iteration. Therefore, based on the above definitions, assumptions and our algorithms, we can express the evolution of the virtual global parameter/weight vector in the following forms:
\begin{equation}
	\boldsymbol{x}^r=\left[\begin{array}{c}
		\boldsymbol{x}_{(1)}^r \\
		\boldsymbol{x}_{(2)}^r \\
		\vdots \\
		\boldsymbol{x}_{(N)}^r
	\end{array}\right]=\frac{1}{K}\left[\begin{array}{c}
		\sum_{k=1}^K x_{k, 1,(1)}^r \\
		\sum_{k=1}^K x_{k, 2,(2)}^r \\
		\vdots \\
		\sum_{k=1}^K x_{k, N,(N)}^r
	\end{array}\right]
\end{equation}

\begin{equation}
	\boldsymbol{x}^{r+1}=\boldsymbol{x}^r-\frac{\eta}{K}\left[\begin{array}{c}
		\sum_{k=1}^K \sum_{t=1}^T \nabla_{(1)} f_{k, 1}\left(y_{k, 1}^{r, t} ; \zeta\right) \\
		\sum_{k=1}^K \sum_{t=1}^T \nabla_{(2)} f_{k, 2}\left(y_{k, 2}^{r, t} ; \zeta\right) \\
		\vdots \\
		\sum_{k=1}^K \sum_{t=1}^T \nabla_{(N)} f_{k, N}\left(y_{k, N}^{r, t} ; \zeta\right).
	\end{array}\right]
\end{equation}

In this case, we update all coordinates of the global weight vector $\boldsymbol{x}^r$, virtually at each time step $t$. We have the virtual gradient at each time instant $t$ as:

\begin{equation}
	\mathbf{G}^{\mathbf{r}} = \frac{1}{K}\left[\begin{array}{c}
		\sum_{k=1}^K \sum_{t=1}^T \nabla_{(1)} f_{k, 1}\left(y_{k, 1}^{r, t} ; \zeta\right) \\
		\sum_{k=1}^K \sum_{t=1}^T \nabla_{(2)} f_{k, 2}\left(y_{k, 2}^{r, t} ; \zeta\right) \\
		\vdots \\
		\sum_{k=1}^K \sum_{t=1}^T \nabla_{(N)} f_{k, N}\left(y_{k, N}^{r, t} ; \zeta\right),
	\end{array}\right]
\end{equation}

\begin{equation}
	\mathbb{E}_{\mathcal{S}}\left[\mathbf{G}^{\mathbf{r}}\right]=\frac{1}{M}\left[\begin{array}{c}
		\sum_{i=1}^M \sum_{t=1}^T \nabla_{(1)} f_i\left(y_i^{r, t}\right) \\
		\sum_{i=1}^M \sum_{t=1}^T \nabla_{(2)} f_i\left(y_i^{r, t}\right) \\
		\vdots \\
		\sum_{i=1}^M \sum_{t=1}^T \nabla_{(N)} f_i\left(y_i^{r, t}\right).
	\end{array}\right]
\end{equation}
We optimize the objective function of the tiered decentralized coordinate descent approach with periodic averaging. The objective is to train a global model $\boldsymbol{x}^r$, which is a $d$-vector that can be decomposed as follows:
\begin{equation}
	\boldsymbol{x}^r=\left[\boldsymbol{x}_{(1)}^{r \top}, \ldots, \boldsymbol{x}_{(N)}^{r \top}\right]^{\top}
\end{equation}
where each $\boldsymbol{x}_{(j)}^r$ is the block of $\boldsymbol{x}^r$, or coordinates, for block $j, r$ is communication rounds. The goal of the training algorithm is to minimize an objective function with following structures.
\section{Appendix B: Theoretical Results of FedBCGD,  FedBCGD+}

In this section, we only present the main theoretical results of  the proposed  FedBCGD, FedBCGD+ algorithms in Theorems 1-2, respectively.  The detailed proofs of Theorems 1-2 are given in Appendices  respectively. 
\setcounter{theorem}{0}

Moreover, we provide the convergence properties of the proposed FedBCGD algorithm. In addition, we also present the detailed proof for the theoretical results in the next subsection.

\begin{theorem}[Convergence rates of FedBCGD]
	Suppose that each function $\left\{f_i\right\}$ satisfies Assumptions $1, 2$, and $3$. Then, in each of the following cases, there exist weights $\left\{w_r\right\}$ and local step-sizes $\eta$, the output of FedBCGD (i.e. $\overline{\boldsymbol{z}}^R$) satisfies the following inequalities.
	
	\textbf{1. Case of strongly convex}: $f_i$ satisfies Assumption 1 for $\mu>0$, $\tilde{\eta}=\frac{\alpha\eta T}{4}$, 
	$\tilde{\eta}\leq\frac{1}{\beta}$ then
	
	\begin{equation}
		\begin{aligned}
			\mathbb{E}\left[f\left(\bar{z}^R\right)\right]-f\left(x^{\star}\right) \leq&\left\|x^0-x^{\star}\right\|^2 \mu \exp \left(-\frac{\alpha \mu R}{\beta}\right)+\frac{128\left[\left(1-\frac{K}{M}\right) \frac{1}{K}\right] G^2+32 \frac{\sigma^2}{K T}}{\mu R} \\
			& +\frac{\left(384 \beta G^2+\frac{192 \beta}{T} \sigma^2\right)}{\alpha^2 \mu^2 R^2}+\frac{\left(6144 \beta^2 G^2+\frac{3072}{T} \beta^2 \sigma^2\right)}{\alpha^2 \mu^3 R^3}
		\end{aligned}
	\end{equation}
	
	\textbf{2. Case of general convex}: Each $f_i$ satisfies Assumption 1 for $\mu=0$, $\tilde{\eta}=\frac{\alpha\eta T}{4},$ $\tilde{\eta}\leq\frac{1}{\beta}$then
	\begin{equation}
		\begin{aligned}
			& \mathbb{E}\left[f\left(\bar{z}^R\right)\right]-f\left(x^{\star}\right) \\
			\leq& \frac{\beta^{\frac{3}{2}} d_0}{\alpha R}+\frac{\left(6144 \beta^2 G^2+\frac{3072}{T} \beta^2 \sigma^2\right)}{\alpha^2 R}+\frac{\left[32\left[\left(1-\frac{K}{M}\right) \frac{1}{K}\right] G^2+32 \frac{\sigma^2}{K T}\right]^{\frac{1}{2}} d_0^{\frac{1}{2}}}{\sqrt{R}} \\
			& +\frac{\left(384 \beta G^2+\frac{192 \beta}{T } \sigma^2\right)^{\frac{1}{3}} d_0^{\frac{2}{3}}}{\alpha^{\frac{2}{3}} R^{\frac{2}{3}}}.
		\end{aligned}
	\end{equation}
	\textbf{3. Case of non-convex}: Each $f_i$ satisfies Assumption 2 and $\tilde{\eta}=\frac{\alpha\eta T}{4},$ $\tilde{\eta}\leq\frac{1}{\beta}$, then
	\begin{equation}
		\begin{aligned}
			& \frac{1}{R} \sum_{r=1}^R\left\|\nabla f\left(x^r\right)\right\|^2 \leq \frac{16 \beta d_0}{T K \alpha R}+\frac{2 \sqrt{d_0}}{\sqrt{R T M}}\left(\frac{8 \beta}{K}\left(1-\frac{K}{M}\right) G^2+\frac{8 \beta \sigma^2}{T K}\left(1-\frac{K}{M}\right)+\frac{8 \beta}{T M} \sigma^2\right)^{\frac{1}{2}} \\
			& +2\left(\frac{d_0}{R}\right)^{\frac{2}{3}}\left[\frac{384 \beta^2}{\alpha^2} G^2+\frac{92 \beta^2}{T} \frac{\sigma^2}{\alpha^2}+\frac{\left(16 \gamma^2 \beta^2\right)}{T M} \sigma^2+\frac{\left(16 \gamma^2 \beta^2\right) \sigma^2}{T K}\left(1-\frac{K}{M}\right)+\frac{16 \gamma^2 \beta^2}{K}\left(1-\frac{K}{M}\right) G^2\right]^{\frac{1}{3}} \\
			& +2\left(\frac{d_0}{R}\right)^{\frac{3}{4}}\left[\frac{4608}{\alpha^2} \frac{\beta^3}{K}\left(1-\frac{K}{M}\right) G^2+\frac{1152 \beta^3}{K T \alpha^2}\left(1-\frac{K}{M}\right) \sigma^2\right]^{\frac{1}{4}} \\
			& +2\left(\frac{d_0}{R}\right)^{\frac{4}{5}}\left[\frac{9216}{\alpha^2} \frac{\gamma^2 \beta^4}{K}\left(1-\frac{K}{M}\right) G^2+\frac{2304 \beta^4}{K} \frac{\gamma^2}{\alpha^2 T}\left(1-\frac{K}{M}\right) \sigma^2\right]^{\frac{1}{5}}.
		\end{aligned}
	\end{equation}
\end{theorem}
\begin{theorem}[Convergence rates of FedBCGD+]
	Suppose that each function $\left\{f_i\right\}$ satisfies Assumptions $ 1, 2$, and $3$. Then, in each of the following cases, there exist weights $\left\{w_r\right\}$ and local step-sizes $\eta$, the output of FedBCGD+ (i.e., $\overline{\boldsymbol{z}}^R$) satisfies the following inequalities.
	
	\textbf{1. Case of strongly convex}: Each $f_i$ satisfies Assumption 1 for $\mu>0$,$\tilde{\eta}=\frac{\alpha\eta T}{4}$, $\tilde{\eta}\leq \min \left(\frac{1}{81 \beta}, \frac{S}{15 \mu N  }\right)$ then

	\begin{equation}
		\mathbb{E}\left[f\left(\overline{\boldsymbol{z}}^R\right)\right]-f\left(\boldsymbol{x}^{\star}\right) \leq \tilde{\mathcal{O}}\left(\frac{M \mu}{K} \tilde{D}^2 \exp \left(-\min \left\{\frac{M}{30 K}, \frac{\mu}{162 \beta}\right\} R\right)\right) .
	\end{equation}

	\textbf{2. Case of General convex}: Each $f_i$ satisfies Assumption 1 for $\mu=0$, $\tilde{\eta}\leq\frac{1}{\beta}$ then

	\begin{align}
		\mathbb{E}\left[f\left(\bar{z}^{R}\right)\right]-f\left(x^{\star}\right) \leq \mathcal{O}\left(\sqrt{\frac{M}{K}} \frac{\beta \tilde{D}^2}{R}\right).
	\end{align}
	
	\textbf{3. Case of non-convex}: Each $f_i$ satisfies Assumption 2 and $\tilde{\eta}=\frac{1}{4}\alpha T$, $\tilde{\eta} \leq \frac{1}{24 \beta}\left(\frac{K}{M}\right)^{\frac{2}{3}}$  then
	\begin{align}
		\mathbb{E}\left[\left\|\nabla f\left(\bar{z}^{R}\right)\right\|^2\right] \leq \mathcal{O}\left(\frac{\beta F}{R}\left(\frac{M}{K}\right)^{\frac{2}{3}}\right),
	\end{align}
	where $\tilde{D}^2:=\left(\left\|\boldsymbol{x}^0-\boldsymbol{x}^{\star}\right\|^2+\frac{1}{2 N \beta^2} \sum_{i=1}^N\left\|\boldsymbol{c}_i^0-\nabla f_i\left(\boldsymbol{x}^{\star}\right)\right\|^2\right)$ and $F:=\left(f\left(\boldsymbol{x}_0\right)-f\left(\boldsymbol{x}^{\star}\right)\right)$.
\end{theorem}

\section{Appendix C: Main Lemmas}

In this section, we prove some main lemmas, which play key roles for the proofs of  Theorems 1-3.

\begin{lemma} 
	The following holds for any $\beta$-smooth and $\mu$-strongly convex function $h$, and any $\boldsymbol{x}, \boldsymbol{y}, \boldsymbol{z}$ in the domain of $h$ :
	\begin{equation}
		\langle\nabla h(\boldsymbol{x}), \boldsymbol{z}-\boldsymbol{y}\rangle \geq h(\boldsymbol{z})-h(\boldsymbol{y})+\frac{\mu}{4}\|\boldsymbol{y}-\boldsymbol{z}\|^2-\beta\|\boldsymbol{z}-\boldsymbol{x}\|^2.
	\end{equation}
\end{lemma}

$Proof.$
Given any $\boldsymbol{x}, \boldsymbol{y}$, and $\boldsymbol{z}$, we get the following two inequalities using smoothness and strong convexity of $h:$
\begin{align}
	& \langle\nabla h(\boldsymbol{x}), \boldsymbol{z}-\boldsymbol{x}\rangle \geq h(\boldsymbol{z})-h(\boldsymbol{x})-\frac{\beta}{2}\|\boldsymbol{z}-\boldsymbol{x}\|^2, \\
	& \langle\nabla h(\boldsymbol{x}), \boldsymbol{x}-\boldsymbol{y}\rangle \geq h(\boldsymbol{x})-h(\boldsymbol{y})+\frac{\mu}{2}\|\boldsymbol{y}-\boldsymbol{x}\|^2.
\end{align}

Furthermore, applying the relaxed triangle inequality, we can get
\begin{equation}
	\frac{\mu}{2}\|\boldsymbol{y}-\boldsymbol{x}\|^2 \geq \frac{\mu}{4}\|\boldsymbol{y}-\boldsymbol{z}\|^2-\frac{\mu}{2}\|\boldsymbol{x}-\boldsymbol{z}\|^2.
\end{equation}
Combining all the inequalities together, we have
\begin{equation}
	\langle\nabla h(\boldsymbol{x}), \boldsymbol{z}-\boldsymbol{y}\rangle \geq h(\boldsymbol{z})-h(\boldsymbol{y})+\frac{\mu}{4}\|\boldsymbol{y}-\boldsymbol{z}\|^2-\frac{\beta+\mu}{2}\|\boldsymbol{z}-\boldsymbol{x}\|^2 .
\end{equation}
The lemma follows since $\beta \geq \mu$.

\begin{lemma} 
	[Bounding heterogeneity] Recall our bound on the gradient dissimilarity:
	\begin{equation}
		\frac{1}{M} \sum_{i=1}^M\left\|\nabla f_i(x)-\nabla f(x)\right\|^2 \leq G^2.
	\end{equation}
	If $\left\{f_i\right\}$ are convex, we can relax the assumption to
	\begin{equation}
		\frac{1}{M} \sum_{i=1}^M\left\|\nabla f_i(\boldsymbol{x})\right\|^2 \leq G^2+2 \beta\left(f(\boldsymbol{x})-f^{\star}\right).
	\end{equation}
\end{lemma} 
$Proof.$
According to the inequality $\frac{1}{n} \sum_{i=1}^n\left\|\mathbf{a}_i-\overline{\mathbf{a}}\right\|_2^2=\frac{1}{n} \sum_{i=1}^n\left\|\mathbf{a}_i\right\|^2-\|\overline{\mathbf{a}}\|^2$ for $\mathbf{a}_i \in \mathbb{R}^d, \overline{\mathbf{a}}=\frac{1}{n} \sum_{i=1}^n \mathbf{a}_i$,
\begin{equation}
	\frac{1}{M} \sum_{i=1}^M\left\| \nabla f_i(x)- \nabla f(x) \right\|^2 \leq G^2, 
\end{equation}
\begin{equation}
	\begin{split}
		& \frac{1}{M} \sum_{i=1}^M\left\|\nabla f_i(x)\right\|^2 \leq\|\nabla f(x)\|^2+G^2 \\
		& \leq\left\|\nabla f(x)-\nabla f\left(x^{\star}\right)\right\|^2+G^2 \\
		& \leq \frac{1}{M} \sum_{i=1}^M\left\|\nabla f_i(x)-\nabla f_i\left(x^{\star}\right)\right\|^2+G^2 \\
		& \leq 2 \beta\left(f(x)-f^{\star}\right)+G^2.
	\end{split}
\end{equation}

\begin{lemma} 
	\textbf{(Relaxed triangle inequality).} Let $\left\{\boldsymbol{v}_1, \ldots, \boldsymbol{v}_\tau\right\}$ be $\tau$ vectors in $\mathbb{R}^d$. Then the following inequalities are true:
	
	1. $\left\|\boldsymbol{v}_i+\boldsymbol{v}_j\right\|^2 \leq(1+a)\left\|\boldsymbol{v}_i\right\|^2+\left(1+\frac{1}{a}\right)\left\|\boldsymbol{v}_j\right\|^2$ for any $a>0$, and
	
	2. $\left\|\sum_{i=1}^\tau \boldsymbol{v}_i\right\|^2 \leq \tau \sum_{i=1}^\tau\left\|\boldsymbol{v}_i\right\|^2$.
\end{lemma} 

\begin{lemma}
	$K$ is the number of selected clients in block $j$ and $M$ is the total number of clients. The following inequalities can be obtained.
	\begin{equation}	\mathbb{E}\left\|\frac{1}{K} \sum_{i=1}^K \nabla f_i(x)\right\|^2 \leq \mathbb{E}\|\nabla f(x)\|^2+\mathbb{E}\left(1-\frac{K}{M}\right) \frac{1}{K M} \sum_{i=1}^M\left\|\nabla f_i(x)\right\|^2,
	\end{equation}
	\begin{equation}	\mathbb{E}\left\|\frac{1}{K} \sum_{i=1}^K \nabla f_i(x)\right\|^2 \leq \frac{1}{M} \sum_{i=1}^M\left\|\nabla f_i(x)\right\|^2.
	\end{equation}
\end{lemma}

$Proof.$ Define $\mathbb{I}_i$ as the random variable which indicates client $i$ is selected in the $r$-th global epoch.
\begin{equation}
	\begin{split}
		& \quad\;\mathbb{E}\left\|\frac{1}{K} \sum_{i=1}^K \nabla f_i(x)\right\|^2 \\
		& =\mathbb{E}\left\|\frac{1}{K} \sum_{i=1}^M \nabla f_i(x) \mathbb{I}_i\right\| \\
		& =\mathbb{E}\left\langle\frac{1}{K} \sum_{i=1}^M \nabla f_i(x) \mathbb{I}_i, \frac{1}{K} \sum_{i=1}^M \nabla f_j(x) \mathbb{I}_j\right\rangle \\
		& =\mathbb{E} \frac{1}{K^2}\left[\sum_{i, j \in[M], i \neq j}\left\langle\nabla f_i(x), \nabla f_j(x)\right\rangle \mathbb{E}\left[\mathbb{I}_i\mathbb{I}_j\right]+\sum_{i \in[M]}\left\langle\nabla f_i(x), \nabla f_i(x)\right\rangle \mathbb{E}\left[\mathbb{I}_i\right]\right] \\
		& =\mathbb{E} \frac{1}{K^2}\left[\sum_{i, j \in[M], i \neq j} \frac{K(K-1)}{M(M-1)}\left\langle\nabla f_i(x), \nabla f_j(x)\right\rangle+\sum_{i \in[M]} \frac{K}{M}\left\langle\nabla f_i(x), \nabla f_i(x)\right\rangle\right] \\
		& =\mathbb{E} \frac{1}{K^2}\left[\sum_{i, j \in[M]} \frac{K(K-1)}{M(M-1)}\left\langle\nabla f_i(x), \nabla f_j(x)\right\rangle+\sum_{i \in[M]} \frac{K(M-K)}{M(M-1)}\left\langle\nabla f_i(x), \nabla f_i(x)\right\rangle\right] \\
		& \leq \mathbb{E}\|\nabla f(x)\|^2+\mathbb{E}\left(1-\frac{K}{M}\right) \frac{1}{K M} \sum_{i \in[M]}\left\|\nabla f_i(x)\right\|^2 \\
		&\leq \frac{1}{M} \sum_{i \in[M]}\left\|\nabla f_i(x)\right\|^2.
	\end{split}
\end{equation}

We will now proceed to the second part of our lemma's exposition.
\begin{equation}
	\mathbb{E}\left\|\frac{1}{K} \sum_{i=1}^K \nabla f_i(x)\right\|^2 \leq \frac{1}{M} \sum_{i=1}^M\left\|\nabla f_i(x)\right\|^2. \\
\end{equation}
\vspace{-2mm}
$Proof:$
\vspace{-2mm}
\begin{equation}
	\begin{split}
		& \mathbb{E}\left\|\frac{1}{K} \sum_{i=1}^K \nabla f_i(x)\right\|^2=\mathbb{E}\left\|\frac{1}{K} \sum_{i=1}^M \nabla f_i(x) \mathbb{I}_i\right\| \\
		& =\mathbb{E}\left\langle\frac{1}{K} \sum_{i=1}^M \nabla f_i(x) \mathbb{I}_i, \frac{1}{K} \sum_{i=1}^M \nabla f_j(x) \mathbb{I}_j\right\rangle \\
		& =\mathbb{E} \frac{1}{K^2}\left[\sum_{i, j \in[M], i \neq j}\left\langle\nabla f_i(x), \nabla f_j(x)\right\rangle \mathbb{E}\left[\mathbb{I}_i \mathbb{I}_j\right]+\sum_{i \in[M]}\left\langle\nabla f_i(x), \nabla f_i(x)\right\rangle \mathbb{E}\left[\mathbb{I}_i\right]\right] \\
		& =\mathbb{E} \frac{1}{K^2}\left[\sum_{i, j \in[M], i \neq j} \frac{K(K-1)}{M(M-1)}\left\langle\nabla f_i(x), \nabla f_j(x)\right\rangle+\sum_{i \in[M]} \frac{K}{M}\left\langle\nabla f_i(x), \nabla f_i(x)\right\rangle\right] \\
		& =\mathbb{E} \frac{1}{K^2}\left[\sum_{i, j \in[M]} \frac{K(K-1)}{M(M-1)}\left\langle\nabla f_i(x), \nabla f_j(x)\right\rangle+\sum_{i \in[M]} \frac{K(M-K)}{M(M-1)}\left\langle\nabla f_i(x), \nabla f_i(x)\right\rangle\right]\\
		& \leq \frac{M^2}{K^2} \frac{K(K-1)}{M(M-1)} \mathbb{E}\|\nabla f(x)\|^2+\mathbb{E} \frac{1}{K^2}\left[\frac{K}{M}-\frac{K(K-1)}{M(M-1)}\right] \sum_{i \in[M]}\left\|\nabla f_i(x)\right\|^2 \\
		& \leq \frac{M}{K} \frac{(K-1)}{(M-1)} \mathbb{E}\|\nabla f(x)\|^2+\frac{1}{K}\left[1-\frac{(K-1)}{(M-1)}\right] \frac{1}{M} \sum_{i \in[M]} \mathbb{E}\left\|\nabla f_i(x)\right\|^2 \\
		& \leq \frac{M}{K} \frac{(K-1)}{(M-1)} \frac{1}{M} \sum_{i \in[M]} \mathbb{E}\left\|\nabla f_i(x)\right\|^2+\frac{1}{K}\left[1-\frac{(K-1)}{(M-1)}\right] \frac{1}{M} \sum_{i \in[M]} \mathbb{E}\left\|\nabla f_i(x)\right\|^2 \\
		& =\left(\frac{M}{K} \frac{(K-1)}{(M-1)}+\frac{1}{K}\left[1-\frac{(K-1)}{(M-1)}\right]\right) \frac{1}{M} \sum_{i \in[M]} \mathbb{E}\left\|\nabla f_i(x)\right\|^2 \\
		& =\frac{1}{M} \sum_{i \in[M]} \mathbb{E}\left\|\nabla f_i(x)\right\|^2.
	\end{split}
\end{equation}

\begin{lemma}[Bounded drift]
	\begin{equation}
		\sum_{i=1}^M \sum_{t=1}^T \mathbb{E}\left\|y_i^{r, t}-x^r\right\|^2 \leq 6 T^3 \eta^2 \sum_{i=1}^M\left\|\nabla f_i\left(x^r\right)\right\|^2+3 M T^2 \eta^2 \sigma^2.
	\end{equation}
\end{lemma}
$Proof.$

\begin{equation}
	\begin{split}
		& \mathbb{E}\left\|y_i^{r, t-1}-x^r-\eta \nabla f_i\left(y_i^{r, t-1} ; \zeta\right)\right\|^2 \\
		& \leq \mathbb{E}\left\|y_i^{r, t-1}-x^r-\eta \nabla f_i\left(y_i^{r, t-1}\right)\right\|^2+\eta^2 \sigma^2 \\
		&\stackrel{\rm{a}}{\leq}\left(1+\frac{1}{T-1}\right) \mathbb{E}\left\|y_i^{r, t-1}-x^r\right\|^2+T \eta^2\left\|\nabla f_i\left(y_i^{r, t-1}\right)\right\|^2+\eta^2 \sigma^2 \\
		& =\left(1+\frac{1}{T-1}\right) \mathbb{E}\left\|y_i^{r, t-1}-x^r\right\|^2+T \eta^2\left\|\nabla f_i\left(y_i^{r, t-1}\right)-\nabla f_i\left(x^r\right)+\nabla f_i\left(x^r\right)\right\|^2+\eta^2 \sigma^2 \\
		& \leq\left(1+\frac{1}{T-1}\right) \mathbb{E}\left\|y_i^{r, t-1}-x^r\right\|^2+2 T \eta^2\left\|\nabla f_i\left(y_i^{r, t-1}\right)-\nabla f_i\left(x^r\right)\right\|^2+2 T \eta^2\left\|\nabla f_i\left(x^r\right)\right\|^2+\eta^2 \sigma^2 \\
		& \leq\left(1+\frac{1}{T-1}+2 T \eta^2 \beta^2\right) \mathbb{E}^2\left\|y_i^{r, t-1}-x^r\right\|\left\|^2+2 T \eta^2\right\| \nabla f_i\left(x^r\right) \|^2+\eta^2 \sigma^2 \\
		& \leq\left(1+\frac{2}{(T-1)}\right) \mathbb{E}^2\left\|y_i^{r, t-1}-x^r\right\|^2+2 T \eta^2\left\|\nabla f_i\left(x^r\right)\right\|^2+\eta^2 \sigma^2,
	\end{split}
\end{equation}
where the inequality $\stackrel{\rm{a}}{\leq}$  follows directly from Lemma 3. Let $ 2 T \eta^2 \beta^2 \leq \frac{1}{(T-1)}$, and unrolling the above recursion, we have
\begin{equation}
	\begin{split}
		& \mathbb{E}\left\|y_i^{r, t-1}-x^r\right\|^2 \\
		& \leq \sum_{\tau=1}^{t-1}\left(2 T \eta^2\left\|\nabla f_i\left(x^r\right)\right\|^2+\eta^2 \sigma^2\right)\left(1+\frac{2}{(T-1)}\right)^\tau \\
		& \leq\left(2 T \eta^2\left\|\nabla f_i\left(x^r\right)\right\|^2+\eta^2 \sigma^2\right) 3 T .
	\end{split}
\end{equation}
So, we can get
\begin{equation}
	\begin{split}
		& \sum_{i=1}^M \sum_{t=1}^T \mathbb{E}\left\|y_k^{r, t}-x^r\right\|^2 \leq \sum_{i=1}^M \mathbb{E}\left(2 T \eta^2\left\|\nabla f_i\left(x^r\right)\right\|^2+ \eta^2 \sigma^2\right) 3 T^2 \\
		& \leq 6 T^3 \eta^2 \sum_{i=1}^M \mathbb{E}\left\|\nabla f_i\left(x^r\right)\right\|^2+3  M T^2 \eta^2 \sigma^2.
	\end{split}
\end{equation}

\begin{lemma}
	The variance of $\mathbf{G}^r$ can be bounded by the following inequality
	\begin{equation}
		\mathbb{E} \sum_{j=1}^N\left\|\frac{1}{K} \sum_{k=1}^K \sum_{t=1}^T \nabla_{(j)} f_{k, j}\left(y_{k, j}^{r, t} ; \zeta\right)\right\|^2 \leq 2 \frac{T}{K} \sum_{k=1}^K \sum_{t=1}^T\left\|\nabla f_i\left(y_i^{r, t}\right)\right\|^2+2 \frac{T}{K} \sigma^2.
	\end{equation}
\end{lemma}
$Proof.$
\begin{equation}
	\begin{split}
		& \mathbb{E} \sum_{j=1}^N\left\|\frac{1}{K} \sum_{k=1}^K \sum_{t=1}^T \nabla_{(j)} f_{k, j}\left(y_{k, j}^{r, t} ; \zeta\right)\right\|^2 \\
		& \leq \frac{T}{M} \sum_{k=1}^K \sum_{t=1}^T \sum_{j=1}^N\left\|\nabla_{(j)} f_i\left(y_i^{r, t} ; \zeta\right)\right\|^2 \\
		& \leq \frac{T}{K} \sum_{k=1}^K \sum_{t=1}^T\left\|\nabla f_i\left(y_i^{r, t} ; \zeta\right)\right\|^2 \\
		& \leq 2 \frac{T}{K} \sum_{k=1}^K \sum_{t=1}^T\left\|\nabla f_i\left(y_i^{r, t}\right)\right\|^2+2 \frac{T}{K} \sigma^2.
	\end{split}
\end{equation}

\begin{lemma}
	[Linear convergence rate)] For every non-negative sequence $\left\{d_{r-1}\right\}_{r \geq 1}$ and any parameters $\mu>0$, $\eta_{\max } \in(0,1 / \mu], c \geq 0, R \geq \frac{1}{2 \eta_{\max } \mu}$, there exists a constant step-size $\eta \leq \eta_{\max }$ and weights $w_r:=(1-$ $\mu \eta)^{1-r}$ such that for $W_R:=\sum_{r=1}^{R+1} w_r$	
	\begin{equation}
		\Psi_R:=\frac{1}{W_R} \sum_{r=1}^{R+1}\left(\frac{w_r}{\eta}(1-\mu \eta) d_{r-1}-\frac{w_r}{\eta} d_r+c \eta w_r\right)=\tilde{\mathcal{O}}\left(\mu d_0 \exp \left(-\mu \eta_{\max } R\right)+\frac{c}{\mu R}\right).
	\end{equation}
\end{lemma}

\begin{lemma}
	[Sub-linear convergence rate] For every non-negative sequence $\left\{d_{r-1}\right\}_{r \geq 1}$ and any parameters $\eta_{\max } \geq 0, c \geq 0, R \geq 0$, there exists a constant step-size $\eta \leq \eta_{\max }$ and weights $w_r=\overline{1}$ such that,
	\begin{equation}
		\Psi_R:=\frac{1}{R+1} \sum_{r=1}^{R+1}\left(\frac{d_{r-1}}{\eta}-\frac{d_r}{\eta}+c_1 \eta+c_2 \eta^2\right) \leq \frac{d_0}{\eta_{\max }(R+1)}+\frac{2 \sqrt{c_1 d_0}}{\sqrt{R+1}}+2\left(\frac{d_0}{R+1}\right)^{\frac{2}{3}} c_2^{\frac{1}{3}}.
	\end{equation}
\end{lemma}

\begin{lemma}
	[Separating mean and variance] Let $\left\{\Xi_1, \ldots, \Xi_\tau\right\}$ be $\tau$ random variables in $\mathbb{R}^d$ which are not necessarily independent. First suppose that their mean is $\mathbb{E}\left[\Xi_i\right]=\xi_i$ and variance is bounded as $\mathbb{E}\left[\left\|\Xi_i-\xi_i\right\|^2\right] \leq \sigma^2$. Then, the following holds
	
	\begin{equation}
		\mathbb{E}\left[\left\|\sum_{i=1}^\tau \Xi_i\right\|^2\right] \leq\left\|\sum_{i=1}^\tau \xi_i\right\|^2+\tau^2 \sigma^2.
	\end{equation}
	
	Now instead suppose that their conditional mean is $\mathbb{E}\left[\Xi_i \mid \Xi_{i-1}, \ldots \Xi_1\right]=\xi_i$ i.e. the variables $\left\{\Xi_i-\xi_i\right\}$ form a martingale difference sequence, and the variance is bounded by $\mathbb{E}\left[\left\|\Xi_i-\xi_i\right\|^2\right] \leq \sigma^2$ as before. Then we can show the tighter bound
	\begin{equation}
		\mathbb{E}\left[\left\|\sum_{i=1}^\tau \Xi_i\right\|^2\right] \leq 2\left\|\sum_{i=1}^\tau \xi_i\right\|^2+2 \tau \sigma^2.
	\end{equation}
\end{lemma}

\section{Appendix D: Proof of Theorem 1}

\subsection{1. The rate of strongly convex and smooth convergence: }
We outline the FEDBCGD algorihtm in Algorithm 1. In round $r$, we perform the following updates:
\begin{align}
	v^r & =\lambda v^{r-1}+\Delta x^{r-1}, \Delta x^{r-1}=\eta \mathbf{G}^r, \\
	x^r & =x^{r-1}+v^{r-1}.
\end{align}
Before giving the convergence analysis of Theorem 3, we first present the following lemma.
\begin{lemma}
	Let $z^r=x^r+\gamma\left(x^r-x^{r-1}\right), \gamma=\frac{\lambda}{1-\lambda}$, we can get
	\begin{align}
		z^{r+1}=z^r-\frac{1}{1-\lambda} \eta \mathbf{G}^r.
	\end{align}
\end{lemma}
$Proof.$

\begin{align}
	\begin{split}
		& z^{r+1}=x^{r+1}+\gamma\left(x^{r+1}-x^r\right) \\
		& =x^r+v^{r+1}+\gamma\left(v^{r+1}\right) \\
		& =z^r-\gamma\left(v^r\right)+v^{r+1}+\gamma\left(v^{r+1}\right) \\
		& =z^r-\gamma v^r+(1+\gamma) v^{r+1} \\
		& =z^r-\gamma v^r+(1+\gamma)\left(\lambda v^r-\eta \mathbf{G}^r\right) \\
		& \stackrel{\rm{a}}{=} z^r+(-\gamma+(1+\gamma) \beta) v^r+(1+\gamma)\left(-\eta \mathbf{G}^r\right) \\
		& =z^r-\eta(1+\gamma) \mathbf{G}^r \\
		& =z^r-\frac{1}{1-\lambda} \eta \mathbf{G}^r,
	\end{split}
\end{align}
with the equality $\stackrel{\rm{a}}{=}$  , we let $(-\gamma+(1+\gamma) \lambda)=0, \gamma=\frac{\lambda}{1-\lambda}$.
We complete the proof.
\subsection{The proof of Theorem 1}
$Proof.$
We can then apply $z^{r+1}=z^r-\alpha \eta \mathbf{G}^r, \alpha=\frac{1}{1-\lambda}$ to bound the second moment of the server update as
\begin{align}
	\begin{split}
		& \mathbb{E}\left\|z^{r+1}-x^{\star}\right\|^2=\mathbb{E}\left\|z^{r+1}-x^{\star}\right\|^2 \\
		& \leq \mathbb{E}\left\|z^r-x^{\star}\right\|^2+\eta \alpha \underbrace{\mathbb{E}\left\langle-\mathbf{G}^r, z^r-x^{\star}\right\rangle}_{C_1}+\eta^2 \alpha^2 \underbrace{\mathbb{E}\left\|\mathbf{G}^r\right\|^2}_{C_2}.
	\end{split}
\end{align}

The term $C_1$ can be bounded by using perturbed strong-convexity (Lemma 1) with $h=f_k, \boldsymbol{x}=y_k^{r, t}, \boldsymbol{y}=\boldsymbol{x}^{\star}$, and $\boldsymbol{z}=z^r$ to get

\begin{align}
	\begin{split}
		& C_1=-\mathbb{E}\left\langle\mathbf{G}^r, z^r-x^{\star}\right\rangle\\
		&=-\sum_{j=1}^N\left\langle\frac{1}{M} \sum_{i=1}^M \sum_{t=1}^T \nabla_{(j)} f_i\left(y_i^{r, t}\right), z_{(j)}^r-x_{(j)}^{\star}\right\rangle \\
		& =-\left\langle\frac{1}{M} \sum_{k=1}^M \sum_{t=1}^T \nabla f_i\left(y_i^{r, t}\right), z^r-x^{\star}\right\rangle \\
		& \leq-\frac{1}{M} \sum_{k=1}^M \sum_{i=1}^T\left(f_i\left(z^r\right)-f_i\left(x^{\star}\right)-\beta\left\|y_i^{r, t}-z^r\right\|^2+\frac{\mu}{4}\left\|z^r-x^{\star}\right\|^2\right) \\
		& \leq T\left(-f\left(z^r\right)+f\left(x^{\star}\right)-\frac{\mu}{4}\left\|z^r-x^{\star}\right\|^2\right)+\frac{\beta}{M} \sum_{i=1}^M \sum_{t=1}^T\left\|y_i^{r, t}-z^r\right\|^2.
	\end{split}
\end{align}

The term $C_2$ can be bounded by using Lemma 6 in $ \stackrel{\rm{a}}{\leq}$, Lemma 3 in $ \stackrel{\rm{b}}{\leq}$, Lemma 4 and Lemma 2 in $ \stackrel{\rm{c}}{\leq}$.

\begin{equation}
	\begin{split}
		& C_2=\mathbb{E}\left\|\mathbf{G}^r\right\|^2 \\
		& =\sum_{j=1}^N \mathbb{E}\left\|\frac{1}{K} \sum_{k=1}^K \sum_{t=1}^T \nabla_{(j)} f_{k, j}\left(y_{k, j}^{r, t} ; \zeta\right)\right\|^2 \\
		& =\sum_{j=1}^N \mathbb{E}\left\|\frac{1}{K} \sum_{k=1}^K \sum_{t=1}^T \nabla_{(j)} f_{k, j}\left(y_{k, j}^{r, t} ; \zeta\right)-\frac{1}{K} \sum_{k=1}^K \sum_{t=1}^T \nabla_{(j)} f_{k, j}\left(x^r\right)+\frac{1}{K} \sum_{k=1}^K \sum_{t=1}^T \nabla_{(j)} f_{k, j}\left(x^r\right)\right\|^2 \\
		& \leq \sum_{j=1}^N 2 \mathbb{E}\left\|\frac{1}{K} \sum_{k=1}^K \sum_{t=1}^T \nabla_{(j)} f_{k, j}\left(y_{k, j}^{r, t} ; \zeta\right)-\frac{1}{K} \sum_{k=1}^K \sum_{t=1}^T \nabla_{(j)} f_{k, j}\left(x^r\right)\right\|^2+\sum_{j=1}^N 2 \mathbb{E}\left\|\frac{1}{K} \sum_{k=1}^K \sum_{t=1}^T \nabla_{(j)} f_{k, j}\left(x^r\right)\right\|^2 \\
		& \stackrel{\rm{a}}{\leq} 2 \frac{T}{M} \sum_{j=1}^N \sum_{i=1}^M \sum_{t=1}^T\left\|\nabla_{(j)} f_i\left(y_i^{r, t} ; \zeta\right)-\nabla_{(j)} f_i\left(x^r\right)\right\|^2+\sum_{j=1}^N 2 T^2 \mathbb{E}\left\|\frac{1}{K} \sum_{k=1}^K \nabla_{(j)} f_{k, j}\left(x^r\right)\right\|^2\\
		& \leq 4 \frac{T}{M} \sum_{i=1}^M \sum_{t=1}^T\left\|\nabla f_i\left(y_i^{r, t}\right)-\nabla f_i\left(x^r\right)\right\|^2+\sum_{j=1}^N 2 T^2 \mathbb{E}\left\|\frac{1}{K} \sum_{k=1}^K \nabla_{(j)} f_{k, j}\left(x^r\right)\right\|^2+4 \frac{T}{K} \sigma^2 \\
		& \leq 4 \frac{T}{M} \sum_{i=1}^M \sum_{t=1}^T\left\|\nabla_{(j)} f_i\left(y_i^{r, t}\right)-\nabla_{(j)} f_i\left(x^r\right)\right\|^2+\sum_{j=1}^N 2 T^2 \mathbb{E}\left\|\frac{1}{K} \sum_{k=1}^K \nabla_{(j)} f_{k, j}\left(x^r\right)\right\|^2+4 \frac{T}{K} \sigma^2 \\
		& \leq 4 \frac{T \beta^2}{M} \sum_{i=1}^M \sum_{t=1}^T\left\|y_i^{r, t}-x^r\right\|^2+\sum_{j=1}^N 2 T^2 \mathbb{E}\left\|\frac{1}{K} \sum_{k=1}^K \nabla_{(j)} f_{k, j}\left(x^r\right)\right\|^2+4 \frac{T}{K} \sigma^2 \\
		& \leq 4 \frac{\beta^2}{M} \sum_{i=1}^M \sum_{t=1}^T\left\|y_i^{r, t}-x^r\right\|^2+\sum_{j=1}^N 2 \mathbb{E}\left\|\frac{T}{K} \sum_{k=1}^K \nabla_{(j)} f_{k, j}\left(x^r\right)\right\|^2+4 \frac{T}{K} \sigma^2\\
		& \leq 4 \frac{T \beta^2}{M} \sum_{i=1}^M \sum_{t=1}^T\left\|y_i^{r, t}-x^r\right\|^2+2 T^2 \sum_{j=1}^N \mathbb{E}\left\|\frac{1}{K} \sum_{k=1}^K \nabla_{(j)} f_{k, j}\left(x^r\right)-\nabla_{(j)} f\left(x^r\right)+\nabla_{(j)} f\left(x^r\right)\right\|^2+4 \frac{T}{K} \sigma^2 \\
		&  \stackrel{\rm{b}}{\leq} 4 \frac{T \beta^2}{M} \sum_{k=1}^M \sum_{t=1}^T\left\|y_i^{r, t}-x^r\right\|^2+2 T^2\left\|\nabla f\left(x^r\right)\right\|^2+2\left(1-\frac{K}{M}\right) T^2 \frac{1}{K M} \sum_{i=1}^M\left\|\nabla f_i\left(x^r\right)\right\|^2+4 \frac{T}{K} \sigma^2 \\
		&  \stackrel{\rm{c}}{\leq} 4 \frac{T \beta^2}{M} \sum_{k=1}^M \sum_{t=1}^T\left\|y_i^{r, t}-x^r\right\|^2+4 T^2 \beta\left(f(x^r)-f\left(x^{\star}\right)\right)+2\left(1-\frac{K}{M}\right) \frac{T^2}{K}\left(G^2+2 \beta\left(f(x^r)-f\left(x^{\star}\right)\right)\right)+4 \frac{T}{K} \sigma^2,
	\end{split}
\end{equation}

Combining the bounds on $C_1$ and $C_2$ in the original inequality, we can get

\begin{align}
	\begin{split}
		& \mathbb{E}\left\|z^{r+1}-x^{\star}\right\|^2=\mathbb{E}\left\|z^r-x^{\star}\right\|^2+\eta \alpha \underbrace{\mathbb{E}\left\langle-\mathbf{G}^r, z^r-x^{\star}\right\rangle}_{C_1}+\eta^2 \alpha^2 \underbrace{\mathbb{E}\left\|\mathbf{G}^r\right\|^2}_{C_2} \\
		& \leq \mathbb{E}\left\|z^r-x^{\star}\right\|^2+\alpha \eta T\left(-f\left(z^r\right)+f\left(x^{\star}\right)-\frac{\mu}{4}\left\|z^r-x^{\star}\right\|^2\right)+\frac{\alpha \eta \beta}{M} \sum_{k=1}^M \sum_{t=1}^T\left\|y_k^{r, t}-z^r\right\|^2 \\
		& +\alpha^2 \eta^2 \frac{4T \beta^2}{M} \sum_{k=1}^M \sum_{t=1}^T\left\|y_k^{r, t}-z^r\right\|^2+4 T^2 \beta \alpha^2 \eta^2\left(f\left(z^r\right)-f\left(x^{\star}\right)\right)  \\
		& +2\left(1-\frac{K}{M}\right) T^2 \alpha^2 \eta^2 \frac{1}{K}\left(G+2 \beta\left(f\left(z^r\right)-f\left(x^{\star}\right)\right)\right) +4\alpha^2 \eta^2  \frac{T}{K} \sigma^2   \\
		& \stackrel{\rm{a}}{\leq} \mathbb{E}\left\|z^r-x^{\star}\right\|^2+\alpha \eta T\left(-f\left(z^r\right)+f\left(x^{\star}\right)-\frac{\mu}{4}\left\|z^r-x^{\star}\right\|^2\right)\\
		&+\left(\frac{\alpha \eta \beta}{M}
		+\frac{4 T \beta^2 \alpha^2 \eta^2}{M}\right) \sum_{k=1}^M \sum_{t=1}^T\left\|y_k^{r, t}-z^r\right\|^2 
		+2\left(1-\frac{K}{M}\right) T^2 \alpha^2 \eta^2 \frac{1}{K} G  \\
		&+\left(4 T^2 \beta \alpha^2 \eta^2+\left(1-\frac{K}{M}\right) 4 T^2 \alpha^2 \eta^2 \frac{1}{K} \beta\right)\left(f\left(z^r\right)-f\left(x^{\star}\right)\right)
		+4\alpha^2 \eta^2  \frac{T}{K} \sigma^2  \\
		& \stackrel{\rm{b}}{\leq} \mathbb{E}\left\|z^r-x^{\star}\right\|^2+\alpha \eta T\left(-f\left(z^r\right)+f\left(x^{\star}\right)-\frac{\mu}{4}\left\|z^r-x^{\star}\right\|^2\right) \\
		& +\left(\frac{\alpha \eta \beta}{M}+\frac{4T \beta^2 \alpha^2 \eta^2}{M}\right)\left(6 T^3 \eta^2 \sum_{i=1}^M\left\|\nabla f_i\left(z^r\right)\right\|^2+3 M  T^2 \eta^2 \sigma^2\right)  \\
		& +2\left(1-\frac{K}{M}\right) T^2 \alpha^2 \eta^2 \frac{1}{K} G^2+\left(4 T^2 \beta \alpha^2 \eta^2+\left(1-\frac{K}{M}\right) 4 T^2 \alpha^2 \eta^2 \frac{1}{K} \beta\right)\left(f\left(z^r\right)-f\left(x^{\star}\right)\right)
		+4\alpha^2 \eta^2  \frac{T}{K} \sigma^2  \\
		& \stackrel{\rm{c}}{\leq} \mathbb{E}\left\|z^r-x^{\star}\right\|^2+\alpha \eta T\left(-f\left(z^r\right)+f\left(x^{\star}\right)-\frac{\mu}{4}\left\|z^r-x^{\star}\right\|^2\right) +\left(\alpha \eta \beta+4 T \beta^2 \alpha^2 \eta^2\right) 6 T^3 \eta^2 2 \beta\left(f\left(z^r\right)-f\left(x^{\star}\right)\right)\\
		&+\left(\alpha \eta \beta+4 T \beta^2 \alpha^2 \eta^2\right) 6 T^3  \eta^2 G^2+\left(\alpha \eta \beta+4 T \beta^2 \alpha^2 \eta^2\right) 3 T^2 \eta^2 \sigma^2  \\
		& +2\left(1-\frac{K}{M}\right) T^2 \alpha^2 \eta^2 \frac{1}{K} G^2+\left(4 T^2 \beta \alpha^2 \eta^2+\left(1-\frac{K}{M}\right) 4 T^2 \alpha^2 \eta^2 \frac{1}{K} \beta\right)\left(f\left(z^r\right)-f\left(x^{\star}\right)\right)+4\alpha^2 \eta^2  \frac{T}{K} \sigma^2  \\
		& \leq \mathbb{E}\left\|z^r-x^{\star}\right\|^2+\alpha \eta T \frac{\mu}{4}\left\|z^r-x^{\star}\right\|^2 \\
		& +\left[-\alpha \eta T+6 T^3 \alpha^2 \eta^2 2 \beta\left(\eta \beta+4T \beta^2 \eta^2\right)+\left(4 T^2 \beta \alpha^2 \eta^2+\left(1-\frac{K}{M}\right) 4 T^2 \alpha^2 \eta^2 \frac{1}{K} \beta\right)\right]\left(f\left(z^r\right)-f\left(x^{\star}\right)\right)  \\
		& +2\left[\left(1-\frac{K}{M}\right) T^2 \frac{1}{K}\right] \alpha^2 \eta^2 G^2+\left(6 \alpha \beta T^3 \alpha \eta^3+24 \beta^2 T^4 \alpha^2 \eta^4\right) G^2  \\
		& +4\alpha^2 \eta^2  \frac{T}{K} \sigma^2+\left(\alpha \eta \beta+4 T \beta^2 \alpha^2 \eta^2\right) 3  T^2  \eta^2 \sigma^2  ,
	\end{split}
\end{align}
where the inequality $ \stackrel{\rm{b}}{\leq}$ follows Lemma 5, the inequality $ \stackrel{\rm{c}}{\leq}$ holds due to Lemma 2. 
Next, we put $\left(f\left(x^r\right)-f\left(x^*\right)\right)$ term in left.

\begin{align}	
	\begin{split}
		& {\left[\alpha \eta T-6 T^3 \alpha^2 \eta^2 \beta\left(\eta \beta+4 T \beta^2 \eta^2\right)-\left(4 T^2 \beta \alpha^2 \eta^2+\left(1-\frac{K}{M}\right) 4 T^2 \alpha^2 \eta^2 \frac{1}{K} \beta\right)\right]\left(f\left(z^r\right)-f\left(x^{\star}\right)\right)} \\
		& \leq \mathbb{E}\left\|z^r-x^{\star}\right\|^2-\alpha \eta T \frac{\mu}{4}\left\|z^r-x^{\star}\right\|^2+2\left[\left(1-\frac{K}{M}\right) T^2 \frac{1}{K}\right] \alpha^2 \eta^2 G^2+\left(6 \alpha \beta T^3 \alpha \eta^3+24 \beta^2 T^4 \alpha^2 \eta^4\right) G^2 \\
		& + 4\alpha^2 \eta^2  \frac{T}{K} \sigma^2+\left(\alpha \eta \beta+4 T \beta^2 \alpha^2 \eta^2\right) 3 T^2 \eta^2 \sigma^2.
	\end{split}  
\end{align}

\begin{align}
	\begin{split}
		& \left(f\left(z^r\right)-f\left(x^{\star}\right)\right) \\
		&\stackrel{\rm{a}}{\leq} \frac{\left(1-\mu \frac{\alpha \eta T}{4}\right)}{\frac{\alpha \eta T}{4}} \mathbb{E}\left\|z^r-x^*\right\|^2-\frac{4}{\alpha \eta T} \mathbb{E}\left\|z^{r+1}-x^*\right\|^2 \\
		& +2\left[\left(1-\frac{K}{M}\right) T^2 \frac{1}{K}\right] \alpha^2 \eta^2 G^2+\left(6 \beta T^3 \alpha \eta^3+24 \beta^2 T^4 \alpha^2 \eta^4\right) G^2 +4\alpha^2 \eta^2  \frac{T}{K} \sigma^2+\left(\alpha \eta \beta+4 T \beta^2 \alpha^2 \eta^2\right) 3T^2  \eta^2 \sigma^2  \\
		& \leq \frac{(1-\mu \tilde{\eta})}{\tilde{\eta}} \mathbb{E}\left\|z^r-x^{\star}\right\|^2-\frac{1}{\tilde{\eta}} \mathbb{E}\left\|z^{r+1}-x^{\star}\right\|^2 \\
		& +32\left[\left(1-\frac{K}{M}\right) \frac{1}{K}\right] \tilde{\eta} G^2+\frac{384 \beta \tilde{\eta}^2 G^2}{\alpha^2}+\frac{6144 \beta^2 \tilde{\eta}^3 G^2}{\alpha^2} +64 \tilde{\eta} \frac{\sigma^2}{TK} +\frac{192 \beta}{T\alpha^2}   \tilde{\eta}^2 \sigma^2+\frac{3072}{T\alpha^2} \beta^2  \tilde{\eta}^3 \sigma^2   \\
		& \leq \frac{(1-\mu \tilde{\eta})}{\tilde{\eta}} \mathbb{E}\left\|z^r-x^{\star}\right\|^2-\frac{1}{\tilde{\eta}} \mathbb{E}\left\|z^{r+1}-x^{\star}\right\|^2+\left[32\left[\left(1-\frac{K}{M}\right) \frac{1}{K}\right] G^2+64\frac{\sigma^2}{TK}\right] \tilde{\eta} \\ 
		& +\left(\frac{384 }{\alpha^2}  \beta G^2+\frac{192 \beta}{T\alpha^2} \sigma^2\right) \tilde{\eta}^2+\left(\frac{6144  }{\alpha^2}\beta^2 G^2+\frac{3702}{T\alpha^2} \beta^2  \sigma^2\right) \tilde{\eta}^3,
	\end{split} 
\end{align}

where the inequality $ \stackrel{\rm{a}}{\leq}$ follows  $\left[\alpha \eta T-6 T^3 \alpha^2 \eta^2 2 \beta\left(\eta \beta+4 T \beta^2 \eta^2\right)-\left(4 T^2 \beta \alpha^2 \eta^2+\left(1-\frac{K}{M}\right) 4 T^2 \alpha^2 \eta^2 \frac{1}{K} \beta\right)\right] \geq$ $\frac{1}{4} \alpha \eta T$. In the last inequalities, we let $\tilde{\eta}=\frac{\alpha \eta T}{4} $,$\tilde{\eta} \leq \frac{1}{8 \beta}$. With Lemma 7, we can get

\begin{align}
	\begin{split}
		&\mathbb{E}\left[f\left(\bar{z}^R\right)\right]-f\left(x^{\star}\right) \leq\left\|x^0-x^{\star}\right\|^2 \mu \exp \left( -\frac{\alpha\mu R}{\beta}\right)+\frac{32\left[\left(1-\frac{K}{M}\right) \frac{1}{K} \right] G^2+64  \frac{\sigma^2}{KT} }{\mu R} \\
		& +\frac{\left(384 \beta G^2+\frac{192 \beta}{T}  \sigma^2\right)}{\alpha^2\mu^2 R^2}+\frac{\left(6144 \beta^2 G^2+\frac{3072}{T} \beta^2 \sigma^2\right)}{\alpha^2\mu^3 R^3} .
	\end{split}
\end{align}

\subsection{2. The convergence rate of general convex and smooth case:}

For general convex  case, we have $\mu=0$, then the following inequality holds:
\begin{align}
	\begin{split}
		& \left(f\left(z^r\right)-f\left(x^{\star}\right)\right) \\
		& \leq \frac{1}{\tilde{\eta}} \mathbb{E}\left\|z^r-x^{\star}\right\|^2-\frac{1}{\tilde{\eta}} \mathbb{E}\left\|z^{r+1}-x^{\star}\right\|^2+\left[32\left[\left(1-\frac{K}{M}\right) \frac{1}{K}\right] G^2+64  \frac{\sigma^2}{KT}\right] \tilde{\eta} \\ & +\left(\frac{384 }{\alpha^2}  \beta G^2+\frac{192 \beta}{T\alpha^2} \sigma^2\right) \tilde{\eta}^2+\left(\frac{6144  }{\alpha^2}\beta^2 G^2+\frac{3072}{T\alpha^2} \beta^2  \sigma^2\right) \tilde{\eta}^3 . 
	\end{split}
\end{align}
With Lemma $8, \tilde{\eta} \leq \frac{1}{\left(192 T \beta^3+64 \beta\right)} \leq \frac{1}{64 \beta}$, we can get,

\begin{align}
	\begin{split}
		&\mathbb{E}\left[f\left(\bar{z}^R\right)\right]-f\left(x^{\star}\right) \\
		& \leq \frac{\beta^{\frac{3}{2}} d_0}{\alpha R}+\frac{\left(6144 \beta^2 G^2+\frac{3072}{T} \beta^2 \sigma^2\right)}{\alpha^2 R}+\frac{\left[32\left[\left(1-\frac{K}{M}\right) \frac{1}{K}\right] G^2+64 \frac{\sigma^2}{KT}\right]^{\frac{1}{2}} d_0^{\frac{1}{2}}}{\sqrt{R}} \\
		& +\frac{\left(\frac{384}{\alpha^2} \beta G^2+\frac{192 \beta}{T \alpha^2} \sigma^2\right)^{\frac{1}{3}} d_0^{\frac{2}{3}}}{\alpha^{\frac{2}{3}} R^{\frac{2}{3}}} .
	\end{split}
\end{align}

\subsection{3. The convergence rate of non-convex and smooth case:}

From the smoothness of the function, we can obtain,
\begin{align}
	\begin{split}
		& \mathbb{E} f\left(z^{r+1}\right) \leq \mathbb{E} f\left(z^r\right)+\mathbb{E}\left\langle\nabla f\left(z^r\right), z^{r+1}-z^r\right\rangle+\frac{\beta}{2} \mathbb{E}\left\|z^{r+1}-z^r\right\|^2 \\
		& \leq \mathbb{E} f\left(z^r\right)+\alpha \eta \underbrace{\mathbb{E}\left\langle\nabla f\left(z^r\right),-\mathbf{G}^r\right\rangle}_{D_1}+\frac{\beta}{2} \eta^2 \alpha^2 \underbrace{\mathbb{E}\left\|\mathbf{G}^r\right\|^2}_{D_2}.
	\end{split}
\end{align}

Next we will perform an upper bound analysis on $D_1$,

\begin{align}
	\begin{split}
		& D_1=-\mathbb{E}\left\langle\nabla f\left(z^r\right), \mathbf{G}^r\right\rangle \\
		& =-\mathbb{E}\left\langle\nabla f\left(z^r\right)-\nabla f\left(x^r\right), \mathbf{G}^r\right\rangle-\mathbb{E}\left\langle\nabla f\left(x^r\right), \mathbf{G}^r\right\rangle \\
		& \leq \frac{1}{2 a} \mathbb{E}\left\|\nabla f\left(z^r\right)-\nabla f\left(x^r\right)\right\|^2+\frac{a}{2}\left\|\mathbb{E}\left[\mathbf{G}^r\right]\right\|^2-\frac{1}{T}\left\langle T \nabla f\left(x^r\right), \mathbb{E}\left[\mathbf{G}^r\right]\right\rangle \\
		& \leq \frac{1}{2 a} \mathbb{E}\left\|\nabla f\left(z^r\right)-\nabla f\left(x^r\right)\right\|^2-\left(\frac{1}{4 T}-\frac{a}{2}\right)\left\|\mathbb{E}\left[\mathbf{G}^r\right]\right\|^2-\frac{1}{2} T\left\|\nabla f\left(x^r\right)\right\|^2 \\
		& +\frac{1}{2 T}\left\|T \nabla f\left(x^r\right)-\frac{1}{M} \sum_{i=1}^M \sum_{t=1}^T \nabla_{(j)} f_i\left(y_i^{r, t}\right)\right\|^2   \\
		& \leq \frac{\beta^2}{2 a} \mathbb{E}\left\|z^r-x^r\right\|^2-\left(\frac{1}{4 T}-\frac{a}{2}\right)\left\|\mathbb{E}\left[\mathbf{G}^r\right]\right\|^2-\frac{1}{2} T\left\|\nabla f\left(x^r\right)\right\|^2+\frac{\beta^2}{2 M} \sum_{i=1}^M \sum_{t=1}^T\left\|x^r-y_i^{r, t}\right\|^2.
	\end{split}
\end{align}

Next we will find the upper bound constraint on $\mathbb{E}\left\|z^r-x^r\right\|^2$
\begin{align}
	\begin{split}
		& z^r=x^r+\gamma\left(x^r-x^{r-1}\right) \\
		& \left\|z^r-x^r\right\|^2 \leq \gamma^2\left\|x^r-x^{r-1}\right\|^2 \\
		& \left\|x^r-x^{r-1}\right\|^2=\gamma^2\left\|\eta^2 \mathbf{G}^r+\beta v^{r-1}\right\|^2 \\
		& \leq \eta^2 \gamma^2 \underbrace{\left\|\sum_{s=0}^r \beta^{r-s} \mathbf{G}^s\right\|^2}_{T_1}.
	\end{split}
\end{align}

For the first term $T_1$, taking the total expectation, we get

\begin{align}
	\begin{split}
		& \mathbb{E}\left[T_1\right] \leq\left(\sum_{s=0}^r \beta^{r-s}\right) \sum_{s=0}^r \beta^{r-s} \mathbb{E}\left[\left\|\mathbf{G}^s\right\|^2\right] \\
		& \leq\left(\sum_{s=0}^r \beta^{r-s}\right) \sum_{s=0}^r \beta^{r-s} \mathbb{E}\left[\left\|\mathbf{G}^s\right\|^2\right] \\
		& \leq \frac{1}{1-\beta} \sum_{s=0}^r \beta^{r-s} \mathbb{E}\left[\left\|\mathbf{G}^s\right\|^2\right].
	\end{split}
\end{align}

Finally, we can get,
\begin{align}
	\mathbb{E}\left\|z^r-x^r\right\|^2 \leq \frac{\eta^2 \gamma^2}{1-\beta} \sum_{s=0}^r \beta^{r-s} \mathbb{E}\left[\left\|\mathbf{G}^s\right\|^2\right],
\end{align}

and
\begin{align}
	\begin{split}
		& \sum_{r=1}^R\left\|\sum_{s=0}^r \beta^{r-s} \mathbf{G}^s\right\|^2 \leq \frac{1}{1-\beta} \sum_{r=1}^R \mathbb{E}\left[\left\|\mathbf{G}^r\right\|^2\right] \sum_{s=0}^r \beta^{R-s} \\
		& \leq \frac{1}{(1-\beta)^2} \sum_{r=1}^R \mathbb{E}\left[\left\|\mathbf{G}^r\right\|^2\right].
	\end{split}
\end{align}

Next we will perform an upper bound analysis on $D_2$,

\begin{align}
	\begin{split}
		& D_2=\mathbb{E}\left\|\mathbf{G}^r\right\|^2=\sum_{j=1}^N \mathbb{E}\left\|\frac{1}{K} \sum_{k=1}^K \sum_{t=1}^T \nabla_{(j)} f_{k, j}\left(y_{k, j}^{r, t} ; \zeta\right)\right\|^2 \\
		& \leq \frac{1}{M^2} \mathbb{E}\left\|\sum_{i=1}^M \sum_{t=1}^T \nabla f_i\left(y_i^{r, t} ; \zeta\right)\right\|^2+\mathbb{E} \frac{1}{K M}\left(1-\frac{K}{M}\right) \sum_{i=1}^M\left\|\sum_{t=1}^T \nabla f_i\left(y_i^{r, t} ; \zeta\right)\right\|^2 \\
		& \leq \frac{1}{M^2} \mathbb{E}\left\|\sum_{i=1}^M \sum_{t=1}^T \nabla f_i\left(y_i^{r, t}\right)\right\|^2+\mathbb{E} \frac{1}{K M}\left(1-\frac{K}{M}\right)\left[3 T \beta^2 \sum_{i=1}^M \sum_{t=1}^T\left\|y_i^{r, t}-x\right\|^2+M T^2 G^2+M T^2\|\nabla f(x)\|^2\right] \\
		& +\frac{T}{M} \sigma^2+\frac{T}{K}\left(1-\frac{K}{M}\right) \sigma^2 .
	\end{split}
\end{align}

Combining the bounds on $D_1$, $D_2$ in the original inequality, we can get

\begin{align}
	\begin{split}
		& \mathbb{E} f\left(z^{r+1}\right) \leq \mathbb{E} f\left(z^r\right)+\frac{\alpha \eta \beta^2}{2 a} \mathbb{E}\left\|z^r-x^r\right\|^2-\alpha \eta\left(\frac{1}{4 T}-\frac{a}{2}\right)\left\|\mathbb{E}\left[\mathbf{G}^r\right]\right\|^2-\frac{\alpha \eta T}{2}\left\|\nabla f\left(x^r\right)\right\|^2 \\
		& +\frac{\alpha \eta \beta^2}{2 M} \sum_{i=1}^M \sum_{t=1}^T\left\|y_i^{r, t}-x\right\|^2+\frac{\beta}{2} \eta^2 \alpha^2 \mathbb{E}\left\|\mathbf{G}^{\mathbf{r}}\right\|^2 .
	\end{split}
\end{align}

Summing the left and right sides of the above inequality from 1 to $R$ simultaneously, we have
\begin{align}
	\begin{split}
		& \mathbb{E} f\left(z^{R+1}\right) \leq \mathbb{E} f\left(z^0\right)+\frac{\alpha \eta \beta^2}{2 a} \sum_{r=1}^R \mathbb{E}\left\|z^r-x^r\right\|^2-\alpha \eta\left(\frac{1}{4 T}-\frac{a}{2}\right) \sum_{r=1}^R\left\|\mathbb{E}\left[\mathbf{G}^r\right]\right\|^2-\frac{\alpha \eta T}{2} \sum_{r=1}^R\left\|\nabla f\left(x^r\right)\right\|^2 \\
		& +\frac{\alpha \eta \beta^2}{2 M} \sum_{r=1}^R \sum_{i=1}^M \sum_{t=1}^T\left\|y_i^{r, t}-x\right\|^2+\frac{\beta}{2} \eta^2 \alpha^2 \sum_{r=1}^R \mathbb{E}\left\|\mathbf{G}^{\mathbf{r}}\right\|^2 \\
		& \leq \mathbb{E} f\left(z^0\right)+\frac{\alpha \eta \beta^2}{2 a} \frac{\eta^2 \gamma^2}{(1-\beta)^2} \sum_{r=1}^R \mathbb{E}\left[\left\|\mathbf{G}^r\right\|^2\right]-\alpha \eta\left(\frac{1}{4 T}-\frac{a}{2}\right) \sum_{r=1}^R\left\|\mathbb{E}\left[\mathbf{G}^r\right]\right\|^2-\frac{\alpha \eta T}{2} \sum_{r=1}^R\left\|\nabla f\left(x^r\right)\right\|^2 \\
		& +\frac{\alpha \eta \beta^2}{2 M} \sum_{r=1}^R \sum_{i=1}^M \sum_{t=1}^T\left\|y_i^{r, t}-x\right\|^2+\frac{\beta}{2} \eta^2 \alpha^2 \sum_{r=1}^R \mathbb{E}\left\|\mathbf{G}^r\right\|^2 \\
		& \leq \mathbb{E} f\left(z^0\right) +\left(\frac{\alpha \eta \beta^2}{2 a} \frac{\eta^2 \gamma^2}{(1-\beta)^2}+\frac{\beta}{2} \eta^2 \alpha^2\right) \sum_{r=1}^R\left[\frac{1}{M^2} \mathbb{E}\left\|\sum_{i=1}^M \sum_{t=1}^T \nabla f_i\left(y_i^{r,t}\right)\right\|\right.\\
		&\left.
		+\mathbb{E} \frac{1}{K M}\left(1-\frac{K}{M}\right)\left[3 T \beta^2 \sum_{i=1}^M \sum_{t=1}^T\left\|y_i^{r,t}-x\right\|^2+M T^2 G^2+M T^2\|\nabla f(x)\|^2\right]+\frac{T}{M} \sigma^2+\left(1-\frac{K}{M}\right) \frac{T}{K} \sigma^2\right]  \\
		& -\alpha \eta\left(\frac{1}{4 T}-\frac{a}{2}\right) \sum_{r=1}^R\left\|\mathbb{E}\left[\mathbf{G}^r\right]\right\|^2-\frac{\alpha \eta T}{2} \sum_{r=1}^R\left\|\nabla f\left(x^r\right)\right\|^2+\frac{\alpha \eta \beta^2}{2 M} \sum_{r=1}^R \sum_{i=1}^M \sum_{t=1}^T\left\|y_i^{r, t}-x\right\|^2 \\
		& \leq \mathbb{E} f\left(z^0\right)+\sum_{r=1}^R\left[\frac{C_1}{M^2} \mathbb{E}\left\|\sum_{i=1}^M \sum_{t=1}^T \nabla f_i\left(y_i^{r ,t}\right)\right\|^2\right.\\
		&\left.
		+\mathbb{E} \frac{C_1}{K M}\left(1-\frac{K}{M}\right)\left[3 T \beta^2 \sum_{i=1}^M \sum_{t=1}^T\left\|y_i^{r, r}-x\right\|^2+M T^2 G^2+M T^2\|\nabla f(x)\|^2\right]+C_1 \frac{T}{M} \sigma^2+C_1\left(1-\frac{K}{M}\right) \frac{T \sigma^2}{K}\right]  \\
		& -\alpha \eta\left(\frac{1}{4 T}-\frac{a}{2}\right) \sum_{r=1}^R\left\|\mathbb{E}\left[\mathbf{G}^r\right]\right\|^2-\frac{\alpha \eta T}{2} \sum_{r=1}^R\left\|\nabla f\left(x^r\right)\right\|^2+\frac{\alpha \eta \beta^2}{2 M} \sum_{r=1}^R \sum_{i=1}^M \sum_{t=1}^T\left\|y_i^{r, t}-x\right\|^2, 
	\end{split} 
\end{align}
let
\begin{align}
	\left(\frac{\alpha \eta \beta^2}{2 a} \frac{\eta^2 \gamma^2}{(1-\beta)^2}+\frac{\beta}{2} \eta^2 \alpha^2\right)=C_1 ,
\end{align}

\begin{align}
	\begin{split}
		& \leq \mathbb{E} f\left(z^0\right)+\left(\frac{\alpha \eta \beta^2}{2 M}+\frac{3 T \beta^2 C_1}{K M}\left(1-\frac{K}{M}\right)\right) \sum_{r=1}^R \sum_{i=1}^M \sum_{t=1}^T \mathbb{E}\left\|y_i^{r, t}-x\right\|^2 \\
		& +\frac{R C_1 T^2 G^2}{K}\left(1-\frac{K}{M}\right)+\left(\left(1-\frac{K}{M}\right) \frac{C_1 T^2}{K}-\frac{\alpha \eta T}{2}\right) \sum_{r=1}^R\left\|\nabla f\left(x^r\right)\right\|^2 \\
		&+\left(\frac{C_1}{M^2}-\alpha \eta\left(\frac{1}{4 T}-\frac{a}{2}\right)\right) \sum_{r=1}^R\left\|\mathbb{E}\left[\mathbf{G}^r\right]\right\|^2 
		+C_1 \frac{T R}{M} \sigma^2+C_1 \frac{T R \sigma^2}{K}\left(1-\frac{K}{M}\right) \\
		& \leq \mathbb{E} f\left(z^0\right)+\left(\frac{\alpha \eta \beta^2}{2}+\frac{3 T \beta^2 C_1}{K}\left(1-\frac{K}{M}\right)\right) \sum_{r=1}^R\left[6 T^3 \eta^2 \frac{1}{M} \sum_{i=1}^M\left\|\nabla f_i\left(x^r\right)\right\|^2+3 T^2 \eta^2 \sigma^2\right] \\
		& +\frac{R C_1 T^2 G^2}{K}\left(1-\frac{K}{M}\right)+\left(\frac{C_1 T^2}{K}\left(1-\frac{K}{M}\right)-\frac{\alpha \eta T}{2}\right) \sum_{r=1}^R\left\|\nabla f\left(x^r\right)\right\|^2  \\
		& +\left(\frac{C_1}{M^2}-\alpha \eta\left(\frac{1}{4 T}-\frac{a}{2}\right)\right) \sum_{r=1}^R\left\|\mathbb{E}\left[\mathbf{G}^r\right]\right\|^2+C_1 \frac{T R}{M} \sigma^2+C_1 \frac{T R \sigma^2}{K} \\
		& \leq \mathbb{E} f\left(z^0\right)+\left(\frac{\alpha \eta \beta^2}{2}+\frac{3 T \beta^2 C_1}{K}\left(1-\frac{K}{M}\right)\right) \sum_{r=1}^R\left[12 T^3 \eta^2 G^2+12 T^3 \eta^2\left\|\nabla f\left(x^r\right)\right\|^2+3 T^2 \eta^2 \sigma^2\right] \\
		& +\frac{R C_1 T^2 G^2}{K}\left(1-\frac{K}{M}\right)+\left(\frac{C_1 T^2}{K}\left(1-\frac{K}{M}\right)-\frac{\alpha \eta T}{2}\right) \sum_{r=1}^R\left\|\nabla f\left(x^r\right)\right\|^2  \\
		& +\left(\frac{C_1}{M^2}-\alpha \eta\left(\frac{1}{4 T}-\frac{a}{2}\right)\right) \sum_{r=1}^R\left\|\mathbb{E}\left[\mathbf{G}^r\right]\right\|^2+C_1 \frac{T R}{M} \sigma^2+C_1 \frac{T R \sigma^2}{K}\left(1-\frac{K}{M}\right) .
	\end{split}
\end{align}
Moving $\left\|\nabla f\left(x^r\right)\right\|^2$ to the left, we can obtain
\begin{align}
	\begin{split}
		& \left(\frac{\alpha \eta T}{2}-\frac{C_1 T^2}{K}-12 T^3 \eta^2\left(\frac{\alpha \eta \beta^2}{2}+\frac{3 T \beta^2 C_1}{K}\left(1-\frac{K}{M}\right)\right)\right) \sum_{r=1}^R\left\|\nabla f\left(x^r\right)\right\|^2 \\
		& \leq \mathbb{E} f\left(z^0\right)+12 T^3 \eta^2 G^2\left(\frac{\alpha \eta \beta^2}{2}+\frac{3 T \beta^2 C_1}{K}\left(1-\frac{K}{M}\right)\right) R+\frac{R C_1 T^2 G^2}{K}\left(1-\frac{K}{M}\right)  \\
		& +3 T^2 \eta^2 \sigma^2\left(\frac{\alpha \eta \beta^2}{2}+\frac{3 T \beta^2 C_1}{K}\left(1-\frac{K}{M}\right)\right) R+C_1 \frac{T R}{M} \sigma^2+C_1 \frac{T R \sigma^2}{K}\left(1-\frac{K}{M}\right) .
	\end{split}
\end{align}

Let $\left(\frac{\alpha \eta T}{2}-\frac{C_1 T^2}{K}-12 T^3 \eta^2\left(\frac{\alpha \eta \beta^2}{2}+\frac{3 T \beta^2 C_1}{K}\left(1-\frac{K}{M}\right)\right)\right) \leq \frac{\alpha \eta T}{4}$,$\tilde{\eta}=\frac{1}{4} \alpha \eta T$,$\tilde{\eta} \leq \frac{1}{16 \beta}$, we can get,

\begin{align}
	\begin{split}
		& \tilde{\eta} \frac{1}{R} \sum_{r=1}^R\left\|\nabla f\left(x^r\right)\right\|^2 \leq \frac{\mathbb{E} f\left(z^0\right)}{R}+12 T^3 \eta^2 G^2\left(\frac{\alpha \eta \beta^2}{2}+\frac{3 T \beta^2 C_1}{K}\left(1-\frac{K}{M}\right)\right)+\frac{C_1 T^2 G^2}{K}\left(1-\frac{K}{M}\right) \\
		& +3 T^2 \eta^2 \sigma^2\left(\frac{\alpha \eta \beta^2}{2}+\frac{3 T \beta^2 C_1}{K}\left(1-\frac{K}{M}\right)\right)+C_1 \frac{T}{M} \sigma^2+C_1 \frac{T \sigma^2}{K}\left(1-\frac{K}{M}\right) .
	\end{split}
\end{align}

Moving $\tilde{\eta}$ to the left, we can obtain
\begin{align}
	\begin{split}
		& \frac{1}{R} \sum_{r=1}^R\left\|\nabla f\left(x^r\right)\right\|^2 \\
		& \leq \frac{\mathbb{E} f\left(z^0\right)}{\tilde{\eta} R}+\frac{192 T \tilde{\eta}}{\alpha^2} G^2\left(\frac{2 \tilde{\eta} \beta^2}{T}+\frac{3 T \beta^2}{K}\left(16 \frac{\tilde{\eta}^3 \gamma^2 \beta^2}{T^2}+8 \frac{\beta \tilde{\eta}^2}{T^2}\right)\left(1-\frac{K}{M}\right)\right)\\
		&+\frac{\left(16 \tilde{\eta}^2 \gamma^2 \beta^2+8 \beta \tilde{\eta}\right) G^2}{K}\left(1-\frac{K}{M}\right)  \\
		& +\frac{48 \tilde{\eta} \sigma^2}{\alpha^2}\left(\frac{2 \tilde{\eta} \beta^2}{T}+\frac{3 T \beta^2}{K}\left(16 \frac{\tilde{\eta}^3 \gamma^2 \beta^2}{T^2}+8 \frac{\beta \tilde{\eta}^2}{T^2}\right)\left(1-\frac{K}{M}\right)\right)  \\
		& +\left(16 \frac{\tilde{\eta}^2 \gamma^2 \beta^2}{T^2}+8 \frac{\beta \tilde{\eta}}{T^2}\right) \frac{T}{M} \sigma^2+\left(16 \frac{\tilde{\eta}^2 \gamma^2 \beta^2}{T^2}+8 \frac{\beta \tilde{\eta}}{T^2}\right)\left(1-\frac{K}{M}\right) \frac{T \sigma^2}{K} \\
		& \leq \frac{\mathbb{E} f\left(z^0\right)}{\tilde{\eta} R}+G^2\left(\frac{384 \beta^2 \tilde{\eta}^2}{\alpha^2}+\frac{9216}{\alpha^2} \frac{\gamma^2 \beta^4 \tilde{\eta}^4}{K}\left(1-\frac{K}{M}\right)
		\frac{4608}{\alpha^2} \frac{\beta^3 \tilde{\eta}^3}{K}\left(1-\frac{K}{M}\right)\right)\\
		&+\frac{\left(16 \tilde{\eta}^2 \gamma^2 \beta^2+8 \beta \tilde{\eta}\right) G^2}{K}\left(1-\frac{K}{M}\right)  \\
		& +\left(\frac{92 \beta^2}{T} \frac{\tilde{\eta}^2 \sigma^2}{\alpha^2}+\frac{2304 \beta^4}{K} \frac{\tilde{\eta}^4 \gamma^2 \sigma^2}{\alpha^2 T}\left(1-\frac{K}{M}\right)+\frac{1152 \beta^3}{K} \frac{\tilde{\eta}^3 \sigma^2}{T \alpha^2}\left(1-\frac{K}{M}\right)\right)  \\
		& +\frac{\left(16 \tilde{\eta}^2 \gamma^2 \beta^2+8 \beta \tilde{\eta}\right)}{T M} \sigma^2+\frac{\left(16 \tilde{\eta}^2 \gamma^2 \beta^2+8 \beta \tilde{\eta}\right) \sigma^2}{T K}\left(1-\frac{K}{M}\right)  \\
		& \leq \frac{\mathbb{E} f\left(z^0\right)}{\tilde{\eta} R}+\frac{8 \beta}{T M} \sigma^2 \tilde{\eta}+\frac{8 \beta}{T K}\left(1-\frac{K}{M}\right) \sigma^2 \tilde{\eta}+\frac{8 \beta}{K}\left(1-\frac{K}{M}\right) G^2 \tilde{\eta} \\
		& +\frac{384 \beta^2}{\alpha^2} G^2 \tilde{\eta}^2+\frac{92 \beta^2}{\alpha^2 T} \sigma^2 \tilde{\eta}^2+\frac{\left(16 \gamma^2 \beta^2\right)}{T M} \sigma^2 \tilde{\eta}^2+\frac{16 \gamma^2 \beta^2}{T K}\left(1-\frac{K}{M}\right) \sigma^2 \tilde{\eta}^2+\frac{16 \gamma^2 \beta^2}{K}\left(1-\frac{K}{M}\right) G^2 \tilde{\eta}^2  \\
		& +\frac{4608}{\alpha^2} \frac{\beta^3}{K}\left(1-\frac{K}{M}\right) G^2 \tilde{\eta}^3+\frac{1152 \beta^3}{K T \alpha^2}\left(1-\frac{K}{M}\right) \sigma^2 \tilde{\eta}^3  \\
		& +\frac{9216}{\alpha^2} \frac{\gamma^2 \beta^4}{K}\left(1-\frac{K}{M}\right) G^2 \tilde{\eta}^4+\frac{2304 \beta^4}{K} \frac{\gamma^2}{\alpha^2 T}\left(1-\frac{K}{M}\right) \sigma^2 \tilde{\eta}^4  .
	\end{split}
\end{align}

With Lemma 8,$\tilde{\eta} \leq \frac{1}{16 \beta}$, we can get
\begin{align}
	\begin{split}
		& \frac{1}{R} \sum_{r=1}^R\left\|\nabla f\left(x^r\right)\right\|^2 \leq \frac{16 \beta d_0}{T K \alpha R}+\frac{2 \sqrt{d_0}}{\sqrt{R T M}}\left(\frac{8 \beta}{K}\left(1-\frac{K}{M}\right) G^2+\frac{8 \beta \sigma^2}{T K}\left(1-\frac{K}{M}\right)+\frac{8 \beta}{T M} \sigma^2\right)^{\frac{1}{2}} \\
		& +2\left(\frac{d_0}{R}\right)^{\frac{2}{3}}\left[\frac{384 \beta^2}{\alpha^2} G^2+\frac{92 \beta^2}{T} \frac{\sigma^2}{\alpha^2}+\frac{\left(16 \gamma^2 \beta^2\right)}{T M} \sigma^2+\frac{\left(16 \gamma^2 \beta^2\right) \sigma^2}{T K}\left(1-\frac{K}{M}\right)+\frac{16 \gamma^2 \beta^2}{K}\left(1-\frac{K}{M}\right) G^2\right]^{\frac{1}{3}}  \\
		& +2\left(\frac{d_0}{R}\right)^{\frac{3}{4}}\left[\frac{4608}{\alpha^2} \frac{\beta^3}{K}\left(1-\frac{K}{M}\right) G^2+\frac{1152 \beta^3}{K T \alpha^2}\left(1-\frac{K}{M}\right) \sigma^2\right]^{\frac{1}{4}}  \\
		& +2\left(\frac{d_0}{R}\right)^{\frac{4}{5}}\left[\frac{9216}{\alpha^2} \frac{\gamma^2 \beta^4}{K}\left(1-\frac{K}{M}\right) G^2+\frac{2304 \beta^4}{K} \frac{\gamma^2}{\alpha^2 T}\left(1-\frac{K}{M}\right) \sigma^2\right]^{\frac{1}{5}} .
	\end{split}
\end{align}

\section{Appendix E: Proof of Theorem 2.}

\subsection{1. The rate of strongly convex and smooth convergence: }

We update the local control variates only for clients $i \in \mathcal{S}^r$
\begin{align}
	\boldsymbol{c}_i^r= \begin{cases}\tilde{\boldsymbol{c}}_i^r & \text { if } i \in \mathcal{S}^r \\ \boldsymbol{c}_i^{r-1} & \text { otherwise }\end{cases}.
\end{align}
Compute the new global parameters and global control variate using only updates from the clients $i \in \mathcal{K}_{j}^r$:

\begin{align}
	\boldsymbol{c}^r_{(j)}=\boldsymbol{c}^r_{(j)}+\frac{K}{M} \sum_{k=1}^K \Delta \boldsymbol{c}^r_{k, j,(j)},
\end{align}

\begin{align}
	\boldsymbol{c}_{(j)}^r=\frac{1}{M} \sum_{i=1}^M \boldsymbol{c}_{i,(j)}^r=\frac{1}{M}\left(\sum_{i \in \mathcal{K}_{j}^r} \boldsymbol{c}_{i,(j)}^r+\sum_{i \notin \mathcal{K}_{j}^r} \boldsymbol{c}_{i,(j)}^{r-1}\right),
\end{align}
\begin{equation}
	\boldsymbol{c}^r=\left[\boldsymbol{c}_{(1)}^{r\top}, \ldots, \boldsymbol{c}_{(N)}^{r\top}\right]^{\top}.
\end{equation}

We define client-drift to be how much the clients move from their starting point:
\begin{align}
	& \mathcal{E}_r=\frac{1}{M T} \sum_{i=1}^M \sum_{t=1}^T\left(\left\|y_i^{r, t}-z^r\right\|^2\right) .
\end{align}
Because we are sampling the clients, not all the client control-variates get updated every round. This leads to some 'lag' which we call control-lag:
\begin{align}
	& \mathcal{C}_r=\frac{1}{M}\sum_{i=1}^M\left\|\mathbb{E} [c_i^r]-\nabla f_i\left(z^r\right)\right\|^2.
\end{align}
With Lemma 10, we have 
\begin{align}	\mathbb{E}\left\|z^{r+1}-x^{\star}\right\|^2 \leq \mathbb{E}\left\|z^r-x^{\star}\right\|^2+2\eta \alpha \underbrace{\mathbb{E}\left\langle-\mathbf{G}^r, z^r-x^{\star}\right\rangle}_{E_1}+\eta^2 \alpha^2 \underbrace{\mathbb{E}\left\|\mathbf{G}^r\right\|^2}_{E_2}.
\end{align}
Before giving the convergence analysis of Theorem 1, we first present the following lemma.
\begin{lemma}
	We can get the bound of $E_2$
	\begin{align}
		\mathbb{E}\left\|\mathbf{G}^r\right\|^2 \leq\left(\frac{4 T^2}{M T}\right) \sum_{i=1}^M \sum_{t=1}^T \mathbb{E}\left\|y_i^{r, t}-z^r\right\|^2+\left(\frac{8 T^2}{M}\right) \sum_{i=1}^M\left\|\mathbb{E}c^r_i-\nabla f_i\left(x^r\right)\right\|^2+\left(\frac{4 T^2}{M}\right) 2 \beta\left(f\left(z^r\right)-f^{\star}\right).
	\end{align}
\end{lemma}
$Proof.$
%The term $E_2$ can be bounded by using Lemma 6 in (a), Lemma 3 in (b), Lemma 4 and Lemma 2 in (c), which seems to $A_2$.
\begin{align}
	\begin{split}
		& E_2=\mathbb{E}\left\|\mathbf{G}^r\right\|^2\\
		&=\sum_{j=1}^N \mathbb{E}\left\|\frac{1}{K} \sum_{k=1}^K \sum_{t=1}^T \nabla_{(j)} f_{k, j}\left(y_{k, j}^{r, t} ; \zeta\right)+c^r_{(j)}-c^r_{k, j,(j)}+\nabla_{(j)} f_{k, j}\left(x^r\right)-\nabla_{(j)} f_{k, j}\left(x^r ; \zeta\right)\right\|^2 \\
		& =\sum_{j=1}^N \mathbb{E}\left\|\frac{1}{K} \sum_{k=1}^K \sum_{t=1}^T \nabla_{(j)} f_{k, j}\left(y_{k, j}^{r, t} ; \zeta\right)+c^r_{(j)}-c^r_{k, j,(j)}+\nabla_{(j)} f_{k, j}\left(x^r\right)-\nabla_{(j)} f_{k, j}\left(x^r ; \zeta\right)\right\|^2 \\
		& \leq \frac{T}{M} \sum_{i=1}^M \sum_{t=1}^T \mathbb{E}\left\|\nabla f_i\left(y_i^{r, t} ; \zeta\right)-\nabla f_i\left(z^r ; \zeta\right)+c^r-c_i^r+\nabla f_i\left(z^r\right)\right\|^2 \\
		& \leq \frac{T}{M} \sum_{k=1}^M \sum_{t=1}^T \mathbb{E}\left\|\nabla f_i\left(y_i^{r, t} ; \zeta\right)-\nabla f_i\left(z^r ; \zeta\right)+c^r-c_i^r+\nabla f_i\left(z^r\right)+\nabla f_i\left(x^{\star}\right)-\nabla f_i\left(x^{\star}\right)\right\|^2\\
		& \leq\left(\frac{4 T}{M}\right) \sum_{i=1}^M \sum_{t=1}^T \mathbb{E}\left\|y_i^{r, t}-z^r\right\|^2+\left(\frac{4 T^2}{M}\right) \sum_{i=1}^M \mathbb{E}\left\|c^r_i-\nabla f_i\left(x^r\right)\right\|^2 \\
		& +\left(4 T^2\right) \mathbb{E}\left\|c^r\right\|^2+\left(\frac{4 T^2}{M}\right) \sum_{i=1}^M \mathbb{E}\left\|f_i\left(z^r\right)-\nabla f_i\left(x^{\star}\right)\right\|^2  \\
		& \leq\left(\frac{4 T}{M}\right) \sum_{i=1}^M \sum_{i=1}^T \mathbb{E}\left\|y_i^{r, t}-z^r\right\|^2+\left(\frac{4 T^2}{M}\right) \sum_{i=1}^M\left\|\mathbb{E} c_i^r-\nabla f_i\left(z^r\right)\right\|^2 \\
		& +\left(4 T^2\right)\left\|\mathbb{E} c^r\right\|^2+\left(\frac{4 T^2}{M}\right) 2 \beta\left(f\left(z^r\right)-f^{\star}\right).
	\end{split} 
\end{align}

\begin{lemma}
	We can get the bound of $E_1$
	\begin{align}
		& E_1\leq-f\left(z^r\right)+f\left(x^{\star}\right)+\frac{\beta}{M} \sum_{i=1}^M \sum_{i=1}^T\left(\left\|y_i^{r, t}-z^r\right\|^2\right)-\frac{\mu}{4} T\left\|z^r-x^{\star}\right\|^2.
	\end{align}
	
\end{lemma}
$Proof. $
The term $E_1$ can be bounded by using perturbed strong-convexity (Lemma 1) with $h=f_i, \boldsymbol{x}=$ $y_i^{r, t}, \boldsymbol{y}=\boldsymbol{x}^{\star}$, and $\boldsymbol{z}=z^r$ .
Next we will calculate the upper bound for $E_1$.
\begin{align}
	\begin{split}
		& E_1=-\mathbb{E}\left\langle\mathbf{G}^r, z^r-x^{\star}\right\rangle=-\sum_{j=1}^N\left\langle\frac{1}{M} \sum_{i=1}^M \sum_{t=1}^T \nabla_{(j)} f_i\left(y_i^{r, t}\right), z_{(j)}^r-x_{(j)}^{\star}\right\rangle \\
		& =-\left\langle\frac{1}{M} \sum_{i=1}^{M,} \sum_{t=1}^T \nabla f_i\left(y_i^{r, t}\right), z^r-x^{\star}\right\rangle \\
		& \leq-\frac{1}{M} \sum_{i=1}^M \sum_{i=1}^T\left(f_i\left(z^r\right)-f_i\left(x^{\star}\right)-\beta\left\|y_i^{r, t}-z^r\right\|^2+\frac{\mu}{4}\left\|z^r-x^{\star}\right\|^2\right) \\
		& \leq-f\left(z^r\right)+f\left(x^{\star}\right)+\frac{\beta}{M} \sum_{i=1}^M \sum_{i=1}^T\left(\left\|y_i^{r, t}-z^r\right\|^2\right)-\frac{\mu}{4} T\left\|z^r-x^{\star}\right\|^2.
	\end{split}
\end{align}
We will now bound the final source of error which is the client-drift.
\begin{lemma}
	$f_i$ satisfies Assumptions 1-4. Then, we can bound the drift as
	\begin{align}
		\frac{1}{T M} \sum_{t=1}^T \sum_{i=1}^M \mathbb{E}\left\|y_i^{r, t}-z^r\right\|^2 \leq 18 T^2 \beta \eta^2\left(f\left(z^r\right)-f\left(x^{\star}\right)\right)+18 T^2 \eta^2 \mathcal{C}_{r-1}.
	\end{align}
\end{lemma}
$Proof.$ First, we observe that if $T=1, \mathcal{E}_r=0$ since $\boldsymbol{y}_{i}^{r,0}=\boldsymbol{z}^r$ for all $i \in[M]$ and that $\Xi_{r-1}$ and the right hand side are both positive. Thus the Lemma is trivially true if $T=1$ and we will henceforth assume $T \geq 2$. Starting from the update rule  for $i \in[M]$ and $t \in[T]$
\begin{align}
	\begin{split}
		& \frac{1}{M} \sum_{i \in M} \mathbb{E}\left\|y_i^{r, t}-z^r\right\|^2 \\
		& =\frac{1}{M} \sum_{i \in M} \mathbb{E}\left\|y_i^{r, t-1}+\eta \nabla f_i\left(y_i^{r, t} ; \zeta\right)-\eta \nabla f_i\left(z^r ; \zeta\right)+\eta c^r-\eta c_i^r+\eta \nabla f_i\left(z^r\right)-z^r\right\|^2 \\
		& \leq(1+a) \frac{1}{M} \sum_{i \in M} \mathbb{E}\left\|y_i^{r, t-1}-z^r+\eta \nabla f_i\left(y_i^{r, t-1} ; \zeta\right)-\eta \nabla f_i\left(z^r ; \zeta\right)\right\|^2\\
		&+\left(1+\frac{1}{a}\right) \eta^2 \frac{1}{M} \sum_{i \in M} \mathbb{E}\left\|c^r-c_i^r+\nabla f_k\left(z^r\right)\right\|^2  \\
		& \leq(1+a) \frac{1}{M} \sum_{i \in M} \mathbb{E}\left\|y_i^{r, t-1}-z^r\right\|^2+\left(1+\frac{1}{a}\right) \eta^2 \frac{1}{M} \sum_{i \in M} \mathbb{E}\left\|c^r-c_i^r+\nabla f_i\left(z^r\right)\right\|^2.
	\end{split}
\end{align}

Once again using our relaxed triangle inequality to expand the other term $\frac{1}{M} \sum_{i\in M} \mathbb{E}\left\|c^r-c_i^r+\nabla f_i\left(x^r\right)\right\|^2$, we get 

\begin{align}
	\begin{split}
		& \frac{1}{M} \sum_{i \in M} \mathbb{E}\left\|c^r-c_i^r+\nabla f_i\left(x^r\right)\right\|^2 \\
		& =\frac{1}{M} \sum_{i=1}^M \mathbb{E}\left\|c^r-c_i^r+\nabla f_i\left(x^r\right)-\nabla f_i\left(x^{\star}\right)+\nabla f_i\left(x^{\star}\right)\right\|^2 \\
		& \leq 3\|\mathbb{E} c^r\|^2+\frac{3}{M} \sum_{i=1}^M\left\|\mathbb{E} c_i-\nabla f_i\left(x^{\star}\right)\right\|^2+\frac{3}{M} \sum_{i=1}^M\left\|\nabla f_i\left(x^r\right)-\nabla f_i\left(x^{\star}\right)\right\|^2 \\
		& \leq \frac{6}{M} \sum_{i=1}^M\left\|\mathbb{E} c_i^r-\nabla f_i\left(x^{\star}\right)\right\|^2+6 \beta\left(f\left(x^r\right)-f\left(x^{\star}\right)\right).
	\end{split}
\end{align}

The last step used the smoothness of $f_i$. Combining the bounds on  in the original inequality and using $a=\frac{1}{T-1}$, we have 
\begin{align}
	\begin{split}
		&\frac{1}{M} \sum_{i=1}^M \mathbb{E}\left\|y_i^{r, t-1}-z^r\right\|^2 \leq \frac{\left(1+\frac{1}{T-1}\right)}{M} \sum_{i=1}^M \mathbb{E}\left\|y_i^{r, t-1}-z^r\right\|^2+6 \eta^2 T \beta\left(f\left(z^r\right)-f\left(x^{\star}\right)\right)\\
		&+\frac{6 T \eta^2}{M} \sum_{i=1}^M\left\|\mathbb{E} c_i^r-\nabla f_i\left(x^{\star}\right)\right\|^2  .
	\end{split}
\end{align}

$
\text {Unrolling the recursion, we get the following for any } t \in\{1, \ldots, T\},
$
\begin{align}
	\begin{split}
		& \frac{1}{M} \sum_{i=1}^M \mathbb{E}\left\|y_i^{r, t-1}-x^r\right\|^2 \leq\left(6 T \beta \eta^2\left(f\left(z^r\right)-f\left(x^{\star}\right)\right)+6 T \eta^2 \mathcal{C}_{r-1}\right)\left(\sum_{\tau=0}^{t-1}\left(1+\frac{1}{T-1}\right)^r\right) \\
		& \leq\left(6 T \beta \eta^2\left(f\left(z^r\right)-f\left(x^{\star}\right)\right)+6 T \eta^2 \mathcal{C}_{r-1}\right)(T-1)\left(\left(1+\frac{1}{T-1}\right)^T-1\right) \\
		& \leq\left(6 T \beta \eta^2\left(f\left(z^r\right)-f\left(x^{\star}\right)\right)+6 T \eta^2 \mathcal{C}_{r-1}\right) 3 T \\
		& \leq 18 T^2 \beta \eta^2\left(f\left(z^r\right)-f\left(x^{\star}\right)\right)+18 T^2 \eta^2 \mathcal{C}_{r-1}.
	\end{split}
\end{align}
The inequality $(T-1)\left(\left(1+\frac{1}{T-1}\right)^T-1\right) \leq 3 T$ can be verified for $T=2,3$ manually. For $T \geq 4$, $(T-1)\left(\left(1+\frac{1}{T-1}\right)^T-1\right)<T\left(\exp \left(\frac{T}{T-1}\right)-1\right) \leq T\left(\exp \left(\frac{4}{3}\right)-1\right)<3 T$.

\begin{align}
	\mathcal{C}_r=\frac{1}{M} \sum_{i=1}^M\left\|\mathbb{E}\left[c_i^r\right]-\nabla f_i\left(x^{\star}\right)\right\|^2.
\end{align}

Again averaging over $t$,

\begin{align}
	\frac{1}{T M} \sum_{t=1}^T \sum_{i=1}^M \mathbb{E}\left\|y_i^{r, t}-x^r\right\|^2 \leq 18 T^2 \beta \eta^2\left(f\left(z^r\right)-f\left(x^{\star}\right)\right)+18 T^2 \eta^2 \mathcal{C}_{r-1}.
\end{align}

\begin{lemma}
	For updates of FedBCGD+ with the control update  and Assumptions 3-4, the following holds true for any $\tilde{\eta} \in[0,1 / \beta]:$
	\begin{align}
		& \mathbb{E}\left[\mathcal{C}_r\right] 
		\leq\left(1-\frac{K}{M}\right) \mathcal{C}_{r-1}+\frac{K}{M}\left(4 \beta\left(\mathbb{E}\left[f\left(z^{r-1}\right)\right]-f\left(x^{\star}\right)\right)\right).
	\end{align}
\end{lemma}

$Proof.$
We define client-drift to be how much the clients move from their starting point:
\begin{align}
	\mathcal{E}_r:=\frac{1}{T M} \sum_{t=1}^T \sum_{i=1}^M \mathbb{E}\left\|y_i^{r, t}-z^r\right\|^2.
\end{align}

$
\text {Plugging the above expression in the definition of} \mathcal{C}_r \text { we get }
$

\begin{align}
	\begin{split}
		& \mathcal{C}_r=\frac{1}{M} \sum_{i=1}^M\left\|\mathbb{E}\left[\mathbf{c}_i^r\right]-\nabla f_i\left(x^{\star}\right)\right\|^2 \\
		& =\frac{1}{M} \sum_{i=1}^M\left\|\left(1-\frac{K}{M}\right)\left(\mathbb{E}\left[\mathbf{c}_i^{r-1}\right]-\nabla f_i\left(x^{\star}\right)\right)+\frac{K}{M}\left(\left[\nabla f_i\left(z^r\right)\right]-\nabla f_i\left(x^{\star}\right)\right)\right\|^2 \\
		& \leq\left(1-\frac{K}{M}\right) \mathcal{C}_{r-1}+\frac{K}{M^2} \sum_{i=1}^M \mathbb{E}\left\|\nabla f_i\left(z^r\right)-\nabla f_i\left(x^{\star}\right)\right\|^2.
	\end{split}
\end{align}

The final step applied Jensen's inequality twice. We can then further simplify using the relaxed triangle inequality as follows:
\begin{align}
	\begin{split}
		& \mathbb{E}\left[\mathcal{C}_r\right] \leq\left(1-\frac{K}{M}\right) \mathcal{C}_{r-1}+\frac{K}{M^2} \sum_{i=1}^M \mathbb{E}\left\|\nabla f_i\left(z^r\right)-\nabla f_i\left(x^{\star}\right)\right\|^2 \\
		& \leq\left(1-\frac{K}{M}\right) \mathcal{C}_{r-1}+\frac{K}{M^2} \sum_{i=1}^M \mathbb{E}\left\|\nabla f_i\left(z^{r-1}\right)-\nabla f_i\left(x^{\star}\right)\right\|^2 \\
		& \leq\left(1-\frac{K}{M}\right) \mathcal{C}_{r-1}+\frac{K}{M^2} \sum_{i=1}^M \mathbb{E}\left\|\nabla f_{i}\left(z^{r-1}\right)-\nabla f_k\left(x^{\star}\right)\right\|^2 \\
		& \leq\left(1-\frac{K}{M}\right) \mathcal{C}_{r-1}+\frac{K}{M}\left(4 \beta\left(\mathbb{E}\left[f\left(z^{r-1}\right)\right]-f\left(x^{\star}\right)\right)\right).
	\end{split}
\end{align}

The last two inequalities follow from smoothness of $\left\{f_i\right\}$ and the definition 

\begin{align}
	\mathcal{E}_r:=\frac{1}{T M} \sum_{t=1}^T \sum_{i=1}^M \mathbb{E}\left\|y_i^{r, t}-z^r\right\|^2.
\end{align}

\begin{lemma}
	\begin{align}
		\left\|z^{r+1}-x^{\star}\right\|^2+9  \tilde{\eta}^2 \frac{M}{K} \mathcal{C}_r \leq\left(1-\frac{\tilde{\eta} \mu}{2}\right)\left\|z^r-x^{\star}\right\|^2+\left(1-\frac{\mu \tilde{\eta}}{2}\right) 9 \tilde{\eta}^2 \frac{M}{K} \mathcal{C}_{r-1}
	\end{align}
\end{lemma}
With $E_1$ and $E_2$,we can get
\begin{align}
	\begin{split}
		&\mathbb{E}\left\|z^{r+1}-x^{\star}\right\|^2 \leq \mathbb{E}\left\|z^r-x^{\star}\right\|^2+2 \eta \alpha \underbrace{\mathbb{E}\left\langle-\mathbf{G}^r, z^r-x^{\star}\right\rangle}_{E_1}+\eta^2 \alpha^2 \underbrace{\mathbb{E}\left\|\mathbf{G}^r\right\|^2}_{E_2}\\
		& \leq\left(1-\frac{\eta \alpha \mu}{2}\right) T\left\|z^r-x^{\star}\right\|^2+2 \eta \alpha T\left(-f\left(z^r\right)+f\left(x^{\star}\right)\right) \\
		& +\left[\frac{2 \eta \alpha \beta}{M}+\eta^2 \alpha^2\left(\frac{4 T}{M}\right)\right] \sum_{i=1}^M \sum_{t=1}^T\left(\left\|y_k^{r, t}-z^r\right\|^2\right) +\eta^2 \alpha^2\left(\frac{8 T^2}{M}\right) \sum_{i=1}^M\left\|\mathbb{E} c_i^r-\nabla f_i\left(z^r\right)\right\|^2  \\
		& \leq\left(1-\frac{\eta \alpha \mu T}{2}\right)\left\|z^r-x^{\star}\right\|^2+2 \eta \alpha T\left(-f\left(z^r\right)+f\left(x^{\star}\right)\right)+\left[2 \eta \alpha T \beta+4 T^2 \eta^2 \alpha^2\right] \frac{1}{M T} \sum_{i=1}^M \sum_{t=1}^T\left(\left\|y_k^{r, t}-z^r\right\|^2\right) \\
		& +8 T^2 \eta^2 \alpha^2\left(\frac{1}{M}\right) \sum_{i=1}^M\left\|\mathbb{E} c_i^r-\nabla f_i\left(z^r\right)\right\|^2 ,
	\end{split} 
\end{align}

with $\tilde{\eta}=\alpha\eta  T$, we have 
\begin{align}
	\begin{split}
		& \leq\left(1-\frac{\tilde{\eta} \mu}{2}\right)\left\|z^r-x^{\star}\right\|^2+2 \tilde{\eta}\left(-f\left(z^r\right)+f\left(x^{\star}\right)\right)+\left[2 \tilde{\eta} \beta+4 \tilde{\eta}^2\right] \frac{1}{M T} \sum_{i=1}^M \sum_{t=1}^T\left(\left\|y_i^{r, t}-z^r\right\|^2\right) \\
		& +8 \tilde{\eta}^2\left(\frac{1}{M}\right) \sum_{i=1}^M\left\|\mathbb{E} c_i^r-\nabla f_i\left(z^r\right)\right\|^2  \\
		& \leq\left(1-\frac{\tilde{\eta} \mu}{2}\right)\left\|z^r-x^{\star}\right\|^2+2 \tilde{\eta}\left(-f\left(z^r\right)+f\left(x^{\star}\right)\right)+\left[2 \tilde{\eta} \beta+4\tilde{\eta}^2\right] \mathcal{E}_r +8 \tilde{\eta}^2 \mathbb{E}\left[\mathcal{C}_r\right].
	\end{split}
\end{align}

We can use Lemma 13 (scaled by $9 \tilde{\eta}^2 \frac{N}{S}$ ) to bound the control-lag
%\begin{align}
%	3 \beta \alpha^2 \tilde{\eta} \mathcal{E}_r \leq 54 \tilde{\eta}^3 \beta^2\left(f\left(z^r\right)-f\left(x^{\star}\right)\right)+54 \tilde{\eta}^3 \beta \mathcal{C}_{r-1}
%\end{align}

\begin{equation}
	3 \beta \tilde{\eta} \mathcal{E}_r \leq \frac{54 \tilde{\eta}^3 \beta^2}{\alpha^2}\left(f\left(z^r\right)-f\left(x^{\star}\right)\right)+\frac{54 \tilde{\eta}^3 \beta}{\alpha^2} \mathcal{C}_{r-1}.
\end{equation}
Now recall that Lemma 14 bounds the client-drift:

\begin{align}
	\begin{split}
		& 9  \tilde{\eta}^2 \frac{M}{K} \mathcal{C}_r \leq\left(1-\frac{\mu \tilde{\eta}}{2}\right) 9 \tilde{\eta}^2 \frac{M}{K} \mathcal{C}_{r-1}+9\left(\frac{\mu \tilde{\eta} M}{2 K}-1\right)\tilde{\eta}^2 \mathcal{C}_{r-1} \\
		& +9\tilde{\eta}^2\left(4 \beta\left(\mathbb{E}\left[f\left(z^{r-1}\right)\right]-f\left(x^{\star}\right)\right)+2 \beta^2 \mathcal{E}\right).
	\end{split}  
\end{align}

Adding all three inequalities together,
we have 
\begin{align}
	\begin{split}
		& \left\|z^{r+1}-x^{\star}\right\|^2+9 \tilde{\eta}^2 \frac{M}{K} \mathcal{C}_r \\
		& \leq\left(1-\frac{\tilde{\eta} \mu}{2}\right)\left\|z^r-x^{\star}\right\|^2+\left(1-\frac{\mu \tilde{\eta}}{2}\right) 9 \tilde{\eta}^2 \frac{M}{K} \mathcal{C}_{r-1}-\left(2 \tilde{\eta}-36 \tilde{\eta}^2 \beta-54 \tilde{\eta}^3 \beta^2\right)\left(f\left(z^r\right)-f\left(x^{\star}\right)\right) \\
		& +\left[-\tilde{\eta} \beta+4\tilde{\eta}^2 \beta^2\right] \mathcal{E}_r+\left( \frac{9 \mu \tilde{\eta} M}{2 S}-9+8+54 \tilde{\eta}\right) \tilde{\eta}^2 \mathcal{C}_{r-1}.
	\end{split}
\end{align}

Finally, with $\tilde{\eta} \leq \frac{1}{81 \beta}$ and $\tilde{\eta} \leq \frac{K}{15 \mu M}$ the lemma follows from noting that
\begin{align}
	& -54 \beta^2\tilde{\eta}^2-36 \beta \tilde{\eta}+2 \geq 0,\\
	&-\tilde{\eta} \beta+4\tilde{\eta}^2 \beta^2\leq 0,\\
	& \frac{9 \mu \tilde{\eta} N}{2 S}-9+8+54 \tilde{\eta}\leq 0.
\end{align}

The final rate for the case of strongly convex follows simply by unrolling the recursive bound and using Lemma 7,
\begin{align}
	\left\|z^{r+1}-x^{\star}\right\|^2+9 \tilde{\eta}^2 \frac{M}{K} \mathcal{C}_r \leq\left(1-\frac{\tilde{\eta} \mu}{2}\right)\left\|z^r-x^{\star}\right\|^2+\left(1-\frac{\mu \tilde{\eta}}{2}\right) 9  \tilde{\eta}^2 \frac{M}{K} \mathcal{C}_{r-1},
\end{align}

\begin{equation}
	\mathbb{E}\left[f\left(\overline{\boldsymbol{z}}^R\right)\right]-f\left(\boldsymbol{x}^{\star}\right) \leq \tilde{\mathcal{O}}\left(\frac{M \mu}{K} \tilde{D}^2 \exp \left(-\min \left\{\frac{K}{30 M}, \frac{\mu}{162 \beta}\right\} R\right)\right) .
\end{equation}
\subsection{2: The convergence rate of general convex and smooth case:}

\begin{align}
	\left\|z^{r+1}-x^{\star}\right\|^2+9 \tilde{\eta}^2 \frac{M}{K} \mathcal{C}_r \leq\left(1-\frac{\tilde{\eta} \mu}{2}\right)\left\|z^r-x^{\star}\right\|^2+\left(1-\frac{\mu \tilde{\eta}}{2}\right) 9  \tilde{\eta}^2 \frac{M}{K} \mathcal{C}_{r-1}.
\end{align}
For general convex  case, we have $\mu=0$, then the following inequality holds:

\begin{align}
	\left\|z^{r+1}-x^{\star}\right\|^2+9 \tilde{\eta}^2 \frac{M}{K} \mathcal{C}_r \leq\left\|z^r-x^{\star}\right\|^2+\left(1-\frac{\mu \tilde{\eta}}{2}\right) 9 \tilde{\eta}^2 \frac{M}{K} \mathcal{C}_{r-1}.
\end{align}

For the general convex setting, averaging over $r$ in Lemma 8,

\begin{align}
	\mathbb{E}\left[f\left(\bar{z}^{R}\right)\right]-f\left(x^{\star}\right) \leq \mathcal{O}\left(\sqrt{\frac{M}{K}} \frac{\beta \tilde{D}^2}{R}\right).
\end{align}

\subsection{3. The convergence rate of non-convex and smooth case:}

Recall that in round $r$, we update the control variate 
\begin{align}
	\boldsymbol{c}_i^r= \begin{cases}\nabla f_{i}\left(\boldsymbol{x}^r\right) & \text { if } i \in \mathcal{S}^r \\ \boldsymbol{c}_i^{r-1} & \text { otherwise }\end{cases}.
\end{align}

We introduce the following notation to keep track of the lag in the update of the control variate: define a sequence of parameters $\left\{\boldsymbol{\alpha}_{i}^{r, t}\right\}$ such that for any $i \in[M]$ and $t \in[T]$ we have $\boldsymbol{\alpha}_{i}^{0, t}:=\boldsymbol{x}^0$ and for $r \geq 1$,

\begin{align}	\boldsymbol{\alpha}_{i}^{r,t}:= \begin{cases}\boldsymbol{y}_{i}^{r, t} & \text { if } i \in \mathcal{S}^r \\ \boldsymbol{\alpha}_{i}^{r-1, t} & \text { otherwise } .\end{cases}
\end{align}

By the update rule for control variates (19) and the definition of $\left\{\boldsymbol{\alpha}_{i}^{r, t}\right\}$ above, the following property always holds:

\begin{align}
	\boldsymbol{c}_{k, j}^r=\nabla f_{k, j}\left(\boldsymbol{x}^r\right).
\end{align}

We can then define the following $\Xi_r$ to be the error in control variate for round $r$:

\begin{align}
	\Xi_r:=\frac{1}{T M} \sum_{t=1}^T \sum_{i=1}^M \mathbb{E}\left\|\boldsymbol{\alpha}_{i}^{r,t}-\boldsymbol{z}^r\right\|^2.
\end{align}

Also recall the closely related definition of client drift caused by local updates:

\begin{align}
	\mathcal{E}_r:=\frac{1}{T M} \sum_{t=1}^T \sum_{i=1}^M \mathbb{E}\left[\left\|\boldsymbol{y}_{i}^{r,t}-\boldsymbol{z}^{r}\right\|^2\right].
\end{align}

From the smoothness of the function, we can obtain
\begin{align}
	\begin{split}
		& \mathbb{E} f\left(z^{r+1}\right) \leq \mathbb{E} f\left(z^r\right)+\mathbb{E}\left\langle\nabla f\left(z^r\right), z^{r+1}-z^r\right\rangle+\frac{\beta}{2} \mathbb{E}\left\|z^{r+1}-z^r\right\|^2 \\
		& \leq \mathbb{E} f\left(z^r\right)+\alpha \eta \underbrace{\mathbb{E}\left\langle\nabla f\left(z^r\right),-\mathbf{G}^r\right\rangle}_{F_1}+\frac{\beta}{2} \eta^2 \alpha^2 \underbrace{\mathbb{E}\left\|\mathbf{G}^r\right\|^2}_{F_2}
	\end{split}.
\end{align}

We will first calculate the upper bound limit for $F_2$,Let us analyze how the control variates effect the variance of the aggregate server update.

\begin{align}
	\begin{split}
		& F_2=\mathbb{E}\left\|\mathbf{G}^r\right\|^2 \\
		& =\sum_{j=1}^N \mathbb{E}\left\|\frac{1}{K} \sum_{k=1}^K \sum_{t=1}^T \nabla_{(j)} f_{k, j}\left(y_{k, j}^{r, t} ; \zeta\right)+c_{(j)}^r-c_{k, j,(j)}^r+\nabla_{(j)} f_{k, j}\left(z^r\right)-\nabla_{(j)} f_{k, j}\left(z^r ; \zeta\right)\right\|^2 \\
		& \leq \frac{1}{M^2} \sum_{j=1}^N \sum_{i=1}^M \sum_{t=1}^T \mathbb{E}\left\|\nabla_{(j)} f_i\left(y_i^{r, t} ; \zeta\right)-\nabla_{(j)} f_i\left(z^r ; \zeta\right)+c_{(j)}^r-c_{i,(j)}^r+\nabla_{(j)} f_i\left(z^r\right)\right\|^2\\
		& \leq \frac{1}{M^2} \sum_{j=1}^N \sum_{i=1}^M \sum_{t=1}^T \mathbb{E}\left\|\nabla f_i\left(y_i^{r, t} ; \zeta\right)-\nabla f_i\left(z^r ; \zeta\right)+c^r-c_i^r+\nabla f_i\left(z^r\right)\right\|^2 \\
		& \leq \frac{1}{M^2} \sum_{i=1}^M \sum_{t=1}^T \mathbb{E}\left\|\nabla f_i\left(y_i^{r, t} ; \zeta\right)-\nabla f_i\left(z^r ; \zeta\right)+c^r-c_i^r+\nabla f_i\left(z^r\right)+\nabla f\left(z^r\right)-\nabla f_i\left(z^r\right)\right\|^2\\
		& \leq\left(\frac{T^2 \beta^2}{M T}\right) \sum_{i=1}^M \sum_{t=1}^T \mathbb{E}\left\|y_i^{r, t}-z^r\right\|^2+\frac{4 T^2}{M^2} \sum_{i=1}^M \mathbb{E}\left\|c_i^r-\nabla f_i\left(z^r\right)\right\|^2  +4 T^2 \mathbb{E}\left\|c^r-\nabla f\left(z^r\right)\right\|^2\\
		&+\left(\frac{4 T^2}{M^2}\right) \sum_{i=1}^M \mathbb{E}\left\|\nabla f\left(z^r\right)\right\|^2  \\
		& \leq\frac{4 T}{M} \sum_{i=1}^M \sum_{t=1}^T \mathbb{E}\left\|y_i^{r, t}-z^r\right\|^2+\frac{4 T^2}{M^2} \sum_{i=1}^M \mathbb{E}\left\|c_i^r-\nabla f_i\left(z^r\right)\right\|^2 \\
		& +4 T^2\mathbb{E} \left\| \frac{1}{M} \sum_{i=1}^M\left[c_i^r-\nabla f_i\left(z^r\right)\right] \right\|^2+\frac{4 T^2}{M^2} \sum_{i=1}^M \mathbb{E}\| \nabla f\left(z^r\right) \|^2.  \\
		& \leq 4 T^2 \beta^2 \mathcal{E}_r+8 \beta^2 T^2 \Xi_{r-1}+4 T^2 \mathbb{E}\left\|\nabla f\left(z^r\right)\right\|^2.
	\end{split}
\end{align}

\begin{lemma}
	Suppose 
	%our step-sizes satisfy $\eta_l \leq \frac{1}{24 \beta K \eta_g}$ and 
	$f_i$ satisfies Assumptions 4-5. We can bound the drift $\mathcal{E}_r \leq\frac{1}{M T} \sum_{i \in M} \mathbb{E}\left\|y_i^{r, t}-z^r\right\|^2$ as
	\begin{align}
		\mathcal{E}_r \leq 24 T^2 \eta^2 \beta^2 \mathbb{E}\left\|z^r-\boldsymbol{\alpha}^r\right\|^2+12 T^2 \eta^2 \mathbb{E}\left\|\nabla f\left(z^r\right)\right\|^2
	\end{align}
\end{lemma}
$Proof.$ 
First, we observe that if $T=1, \mathcal{E}_r=0$ since $\boldsymbol{y}_{i}^{r,0}=\boldsymbol{x}^r$ for all $i \in[M]$ and that $\Xi_{r-1}$ and the right hand side are both positive. Thus the lemma is trivially true if $T=1$ and we will henceforth assume $T \geq 2$. Starting from the update rule (18) for $i \in[N]$ and $t \in[T]$
\begin{align}
	\begin{split}
		& \frac{1}{M} \sum_{i \in M} \mathbb{E}\left\|y_i^{r, t}-z^r\right\|^2 \\
		& =\frac{1}{M} \sum_{i \in M} \mathbb{E}\left\|y_i^{r, t-1}+\eta \nabla f_i\left(y_i^{r, t} ; \zeta\right)-\eta \nabla f_i\left(z^r ; \zeta\right)+\eta c^r-\eta c_i^r+\eta \nabla f_i\left(x^r\right)-z^r\right\|^2 \\
		& \leq(1+a) \frac{1}{M} \sum_{i \in M} \mathbb{E}\left\|y_i^{r, t-1}-z^r\right\|^2+\left(1+\frac{1}{a}\right) \eta^2 \frac{1}{M} \sum_{i \in M} \mathbb{E}\left\|\nabla f_i\left(y_i^{r, t-1} ; \zeta\right)-\nabla f_i\left(z^r ; \zeta\right)+c^r-c_i^r+\nabla f_i\left(z^r\right)\right\|^2 \\
		& \leq\left(1+\frac{1}{T-1}+4 T \beta^2 \eta^2\right) \frac{1}{M} \sum_{i \in M} \mathbb{E}\left\|y_i^{r, t-1}-z^r\right\|^2+4 T \eta^2 \frac{1}{M} \sum_{k \in M} \mathbb{E}\left\|c^r-\nabla f\left(z^r\right)\right\|^2 \\
		& +4 T \eta^2 \frac{1}{M} \sum_{i \in M} \mathbb{E}\left\|\nabla f_i\left(z^r\right)-c_i^r\right\|^2+4 T \eta^2 \mathbb{E}\left\|\nabla f\left(z^r\right)\right\|^2  \\
		& \leq\left(1+\frac{1}{T-1}+4 T \beta^2 \eta^2\right) \frac{1}{M} \sum_{i \in M} \mathbb{E}\left\|y_i^{r, t-1}-z^r\right\|^2+4 T \eta^2 \frac{1}{M} \sum_{i \in M} \mathbb{E}\left\|c^r-\nabla f\left(z^r\right)\right\|^2 \\
		& +4 T \eta^2 \frac{1}{M} \sum_{i \in M} \mathbb{E}\left\|\nabla f_i\left(z^r\right)-c_i^r\right\|^2+4 T \eta^2 \mathbb{E}\left\|\nabla f\left(z^r\right)\right\|^2  \\
		& \leq\left(1+\frac{1}{T-1}+4 T \beta^2 \eta^2\right) \frac{1}{M} \sum_{i \in M} \mathbb{E}\left\|y_i^{r, t-1}-z^r\right\|^2+8 T \eta^2 \beta^2 \mathbb{E}\left\|z^r-\boldsymbol{\alpha}^r\right\|^2+4 T \eta^2 \mathbb{E}\left\|\nabla f\left(z^r\right)\right\|^2 \\
		& \leq 24 T^2 \eta^2 \beta^2 \mathbb{E}\left\|z^r-\boldsymbol{\alpha}^r\right\|^2+12 T^2 \eta^2 \mathbb{E}\left\|\nabla f\left(z^r\right)\right\|^2.
	\end{split}
\end{align}
Averaging the above over $i$, the definition of $c$ and $\Xi_{r-1}$, we have
\begin{align}
	& \frac{1}{M T} \sum_{i \in M} \mathbb{E}\left\|y_i^{r, t}-z^r\right\|^2 \leq 24 T^2 \eta^2 \beta^2 \mathbb{E}\left\|z^r-\boldsymbol{\alpha}^r\right\|^2+12 T^2 \eta^2 \mathbb{E}\left\|\nabla f\left(z^r\right)\right\| .
\end{align}

\begin{lemma}
	For updates of FedBCGD+ and Assumptions 3 and 4, the following holds true for any $\tilde{\eta} \leq \frac{1}{24 \beta}\left(\frac{S}{N}\right)^a$ for $a\in\left[\frac{1}{2}, 1\right]$ where $\tilde{\eta}:=\alpha T\eta$ :
	\begin{align}
		& \Xi_r \leq\left(1-\frac{17 K}{36 M}\right) \Xi_{r-1}+\frac{1}{48 \beta^2}\left(\frac{K}{M}\right)^{2 a-1}\left\|\nabla f\left(z^r\right)\right\|^2+\frac{97}{48}\left(\frac{K}{M}\right)^{2 a-1} \mathcal{E}_r.
	\end{align}
\end{lemma}

$Proof.$
The proof proceeds similar to that of Lemma 13 except that we cannot rely on convexity. Recall that after round $r$, the definition of $\boldsymbol{\alpha}_{i}^{r,t}$  implies that
\begin{align}
	\mathbb{E}\left[\boldsymbol{\alpha}^r\right]=\left(1-\frac{K}{M}\right) \boldsymbol{\alpha}^{r-1}+\frac{K}{M} z^{r-1},
\end{align}
\begin{align}
	\begin{split}
		& \Xi_r=\mathbb{E}\left\|\alpha^r-z^r\right\|^2=\left(1-\frac{K}{M}\right) \cdot \mathbb{E}\left\|\alpha^{r-1}-z^r\right\|^2+\frac{K}{M} \cdot \mathbb{E}\left\|z^{r-1}-z^r\right\|^2 \\
		& \leq\left(1-\frac{K}{M}\right) \mathbb{E}\left(\left\|\alpha^{r-1}-z^{r-1}\right\|^2+\left\|z^r-z^{r-1}\right\|^2+2\left\langle z^r-z^{r-1}, z^{r-1}-\alpha^{r-1}\right\rangle\right)+\frac{K}{M} \cdot \mathbb{E}\left\|z^{r-1}-z^r\right\|^2 \\
		& \leq\left(1-\frac{K}{M}\right) \mathbb{E}\left(\left\|\alpha^{r-1}-z^{r-1}\right\|^2+\left\|z^r-z^{r-1}\right\|^2+\frac{1}{b}\left(2 \tilde{\eta}^2 \beta^2 \mathcal{E}_r+2 \tilde{\eta}^2 \mathbb{E}\left\|\nabla f\left(z^{r-1}\right)\right\|^2\right)+b\left\|\alpha^{r-1}-z^{r-1}\right\|^2\right) \\
		& +\frac{K}{M} \cdot \mathbb{E}\left\|z^{r-1}-z^r\right\|^2  \\
		& \leq\left(1-\frac{K}{M}\right)(1+b) \mathbb{E}\left\|\alpha^{r-1}-z^{r-1}\right\|^2+\left\|z^r-z^{r-1}\right\|^2+\left(1-\frac{K}{M}\right) \frac{1}{b}\left(2 \tilde{\eta}^2 \beta^2 \mathcal{E}_r+2 \tilde{\eta}^2 \mathbb{E}\left\|\nabla f\left(z^r\right)\right\|^2\right) \\
		& \leq\left[\left(1-\frac{K}{M}\right)(1+b)+8 \tilde{\eta}^2 \beta^2\right] \Xi_{r-1}+\left(4 \tilde{\eta}^2 \beta^2+2\left(1-\frac{K}{M}\right) \frac{1}{b} \tilde{\eta}^2 \beta^2\right) \mathcal{E}_r\\
		&+\left(4+2\left(1-\frac{K}{M}\right) \frac{1}{b}\right) \tilde{\eta}^2 \mathbb{E}\left\|\nabla f\left(z^r\right)\right\|^2 .
	\end{split} 
\end{align}

The last inequality applied Lemma 15. Verify that with choice of $b=\frac{K}{2(M-K)}$, we have $\left(1-\frac{K}{M}\right)(1+b) \leq\left(1-\frac{K}{2M}\right)$ and $\frac{1}{b} \leq \frac{2 M}{K}$. Plugging these values along with the bound on the step-size $8 \beta^2 \tilde{\eta}^2 \leq \frac{1}{36}\left(\frac{K}{M}\right)^{2 a} \leq \frac{K}{36M}$,$\tilde{\eta} \leq \frac{1}{24 \beta}\left(\frac{K}{M}\right)^a \text { for } a \in\left[\frac{1}{2}, 1\right]$ completes the lemma.
\begin{align}
	& \Xi_r \leq\left(1-\frac{17 K}{36 M}\right) \Xi_{r-1}+\frac{1}{48 \beta^2}\left(\frac{K}{M}\right)^{2 a-1}\left\|\nabla f\left(z^r\right)\right\|^2+\frac{97}{48}\left(\frac{K}{M}\right)^{2 a-1} \mathcal{E}_r.
\end{align}
\begin{lemma}
	Suppose the updates of FedBCGD+ satisfy Assumptions 2-4. For any effective step-size $\tilde{\eta}$ satisfying $\tilde{\eta} \leq \frac{1}{24 \beta}\left(\frac{K}{M}\right)^{\frac{2}{3}}$
	\begin{align}
		\left(\mathbb{E}\left[f\left(z^r\right)\right]+12 \beta^3 \tilde{\eta}^2 \frac{M}{K} \Xi_r\right) \leq\left(\mathbb{E}\left[f\left(z^{r-1}\right)\right]+12 \beta^3 \tilde{\eta}^2 \frac{M}{K} \Xi_{r-1}\right)-\frac{\tilde{\eta}}{14} \mathbb{E}\left\|\nabla f\left(z^{r-1}\right)\right\|^2
	\end{align}
	
\end{lemma}
$Proof.$
Applying the upper bounds of $F_1$ and $F_2$,
\begin{align}
	\begin{split}
		& \mathbb{E} f\left(z^{r+1}\right) \leq \mathbb{E} f\left(z^r\right)+\mathbb{E}\left\langle\nabla f\left(z^r\right), z^{r+1}-z^r\right\rangle+\frac{\beta}{2} \mathbb{E}\left\|z^{r+1}-z^r\right\|^2 \\
		& \leq \mathbb{E} f\left(z^r\right)+\alpha \eta \underbrace{\mathbb{E}\left\langle\nabla f\left(z^r\right),-\mathbf{G}^r\right\rangle}_{F_1}+\frac{\beta}{2} \eta^2 \alpha^2 \underbrace{\mathbb{E}\left\|\mathbf{G}^r\right\|^2}_{F_2} \\
		& \leq \mathbb{E} f\left(z^r\right)-\frac{\tilde{\eta}}{2}\left\|\nabla f\left(z^r\right)\right\|^2+\frac{\tilde{\eta} \beta^2}{2} \mathcal{E}_r+\frac{\beta}{2} \eta^2 \alpha^2\left[4 T^2 \beta^2 \mathcal{E}_r+8 \beta^2 T^2 \Xi_{r-1}+4 T^2 \mathbb{E}\left\|\nabla f\left(z^r\right)\right\|^2\right] \\
		& \leq \mathbb{E} f\left(z^r\right)-\frac{\tilde{\eta}}{2}\left\|\nabla f\left(z^r\right)\right\|^2+\left(\frac{\tilde{\eta} \beta^2}{2}+2 \beta^3 \tilde{\eta}^2\right) \mathcal{E}_r+4 \beta^3 \tilde{\eta}^2 \Xi_{r-1}+2 \beta \tilde{\eta}^2 \mathbb{E}\left\|\nabla f\left(z^r\right)\right\|^2 \\
		& \leq \mathbb{E} f\left(z^r\right)-\left(\frac{\tilde{\eta}}{2}-2 \beta \tilde{\eta}^2\right)\left\|\nabla f\left(z^r\right)\right\|^2+\left(\frac{\tilde{\eta} \beta^2}{2}+2 \beta^3 \tilde{\eta}^2\right) \mathcal{E}_r+4 \beta^3 \tilde{\eta}^2 \Xi_{r-1} .
	\end{split}
\end{align}
Also recall that Lemmas 16 and 17 state that \begin{align}
	& 12 \beta^3 \tilde{\eta}^2 \frac{M}{K} \Xi_r \leq 12 \beta^3 \tilde{\eta}^2 \frac{M}{K}\left(\left(1-\frac{17 K}{36 M}\right) \Xi_{r-1}+\frac{1}{48 \beta^2}\left(\frac{K}{M}\right)^{2 a-1}\left\|\nabla f\left(z^r\right)\right\|^2+\frac{97}{48}\left(\frac{K}{M}\right)^{2 a-1} \mathcal{E}_r\right) \\
	& \frac{5}{3} \beta^2 \tilde{\eta} \mathcal{E}_r \leq \frac{5}{3 \alpha^2} \beta^3 \tilde{\eta}^2 \Xi_{r-1}+\frac{\tilde{\eta}}{24 \alpha^2} \mathbb{E}\left\|\nabla f\left(z^r\right)\right\|^2.
\end{align}
Adding these bounds on $\Xi_r$ and $\mathcal{E}_r$ to that of $\mathbb{E}[f(z^{r+1})]$ gives
\begin{align}
	& \left(\mathbb{E}\left[f\left(z^{r+1}\right)\right]+12 \beta^3 \tilde{\eta}^2 \frac{M}{K} \Xi_r\right) 
	\leq\left(\mathbb{E}\left[f\left(z^r\right)\right]+12 \beta^3 \tilde{\eta}^2 \frac{M}{K} \Xi_{r-1}\right)+\left(4+\frac{5}{3 \alpha^2}-\frac{17}{3}\right) \beta^3 \tilde{\eta}^2 \Xi_{r-1}\\
	&-\left(\frac{\tilde{\eta}}{2}-2 \beta \tilde{\eta}^2-\frac{1}{4} \beta \tilde{\eta}^2\left(\frac{N}{S}\right)^{2-2 a}-\frac{\tilde{\eta}}{24 \alpha^2}\right)\left\|\nabla f\left(z^r\right)\right\|^2 
	+\left(\frac{\tilde{\eta}}{2}-\frac{5 \tilde{\eta}}{3}+2 \beta \tilde{\eta}^2+\frac{97}{4} \beta \tilde{\eta}^2\left(\frac{M}{K}\right)^{2-2a}\right) \beta^2 \mathcal{E}_r  .
\end{align}
\begin{align}
	\left(\mathbb{E}\left[f\left(z^r\right)\right]+12 \beta^3 \tilde{\eta}^2 \frac{M}{K} \Xi_r\right) \leq\left(\mathbb{E}\left[f\left(z^{r-1}\right)\right]+12 \beta^3 \tilde{\eta}^2 \frac{M}{K} \Xi_{r-1}\right)-\frac{\tilde{\eta}}{14} \mathbb{E}\left\|\nabla f\left(z^{r-1}\right)\right\|^2.
\end{align}
By our choice of $a=\frac{2}{3}$ and plugging in the bound on step-size $\beta \tilde{\eta}\left(\frac{N}{S}\right)^{2-2a} \leq \frac{1}{24}$ proves the lemma.
The non-convex rate of convergence now follows by unrolling the recursion in Lemma 18 and selecting an appropriate step-size $\tilde{\eta}$ as in Lemma 8. Finally, note that if we initialize $\boldsymbol{c}_i^0=\nabla f_i\left(\boldsymbol{x}^0\right)$ then we have $\Xi_0=0$.
We can get
\begin{align}
	\mathbb{E}\left[\left\|\nabla f\left(\bar{z}^{R}\right)\right\|^2\right] \leq \mathcal{O}\left(\frac{\beta F}{R}\left(\frac{M}{K}\right)^{\frac{2}{3}}\right).
\end{align}

\section{Appendix F: More Experimental  Details} In this section, we give some  experimental results: 

\subsection{Methods}
We also demonstrate the robustness of FedBCGD and FedBCGD+  in different settings. For comparison, we use FedAvg \cite{mcmahan2017communication}, SCAFFOLD \cite{karimireddy2020scaffold}, FedAvgM \cite{hsu2019measuring}, FedDC \cite{gao2022feddc} , FedAdam \cite{reddi2020adaptive} FL baselines. The following is a detailed introduction to the experimental setup, model and dataset, and comparison methods.

\subsection{Dataset processing}
We evaluate FL on world datasets of image classification tasks including  CIFAR-10 dataset, CIFAR-100 dataset, Tiny ImageNet dataset, mnist dataset  in our study.Both CIFAR10 and CIFAR100 datasets contain 60000 sheets of 3 × 32 × 32 images. For CIFAR10, there are 10 categories, while there are 100 categories on CIFAR100. For CIFAR10 and CIFAR100, the sample size in the training set is 50000, and the sample size in the test set is 10000. In the experiment, we set up 100 clients with 500 images per client.

Tiny ImageNet Challenge is the default course project for Stanford CS231N. Tiny Imagenet has 200 classes. Each class has 500 training images, 50 validation images, and 50 test images. In the experiment, we set up 100 clients with 1000 images per client. We adjusted the size to 256 × 256 and crop to 224 × 224 to preprocess each image

\subsection{Model}

To test the robustness of our algorithms, we use standard classifiers (including LeNet-5 \cite{lecun2015lenet}, VGG-11, VGG-19 \cite{simonyan2014very}, and ResNet-18 \cite{he2016deep}), Vision Transformer (ViT-Base) \cite{dosovitskiy2020image}, Logistic regression Model \cite{menard2002applied}.
%, and logistic regression \cite{menard2002applied}
We divided the parameters of the model into 5 blocks or more blocks and provide the detailed parameter block division of the model in the Appendix.

\subsection{Hyper-parameter setting}

We provide hyperparameter settings for different datasets. For all real-world datasets in the convolutional network, including CIFAR10 and CIFAR100, set the sampling rate to 10\% for 100 clients. We set the batch size to 50, the number of local epochs for one round of communication to 5, and the initial learning rate is searched in $\{0.01,0.03,0.05,0.1,0.2,0.3\}$. The learning rate decay for each round is 0.998, and the weight decay is 0.001. We searched for FedBCGD and FedAvgM $\alpha$ in $\{0.4, 0.5, 0.6, 0.7, 0.8, 0.9\}$, FedDC settings $\alpha$ = 0.01, FedAdam setting $\alpha$ = 0.9.

For the VIT model, experiments were conducted on Tiny ImageNet and CIFAR100 datasets, and a pre trained model was adopted, with a sampling rate of 10\% for 100 clients. We set the batch size for local training to 16, the number of local epochs for one round of communication to 1, and the initial learning rate to search in $\{0.01,0.03,0.05,0.1,0.2,0.3\}$. The learning rate decay for each round is 0.998, and the weight decay is 0.001. We searched for FedBCGD and FedAvgM $\alpha$ in $\{0.4, 0.5, 0.6, 0.7, 0.8, 0.9\}$, FedDC settings $\alpha$ = 0.01, FedAdam setting $\alpha$ = 0.9.

For the logical classification model, we set the batch size to 50, the number of local epochs in one round of communication to 1 on EMNIST. The initial learning rate is searched in $\{0.01,0.03,0.05,0.1,0.2,0.3\}$, with a learning rate decay of 0.998 and a weight decay of 0.001 for each round. We searched for FedBCGD and FedAvgM $\alpha$ in $\{0.4, 0.5, 0.6, 0.7, 0.8, 0.9\}$, FedDC settings $\alpha$ = 0.01, FedAdam setting $\alpha$ = 0.9.

\subsection{Results on Logistic Regression}
%In the above experiment, the FedBCGD+ algorithm is not the best algorithm, which makes  a gap between complex deep learning experiments and theory.
We use a logistic regression model to verify the consistency between FedBCGD+'s practice and theory results.
%The following two experiments are used to verify that the experimental results of FedBCGDM+ can match its theoretical results.
We conducted the classification tests on the EMNIST dataset by using strongly convex and non-convex loss function models.
To test the performance of our algorithms, we use classical logistic regression problems, whose function has the following form:
\vspace{-2mm}
\begin{align}
	f(x)=\frac{1}{N} \sum_{i=1}^N \log \left(1+\exp \left(-b_i a_i^{\top} x\right)\right)+\frac{\lambda}{2}\|x\|^2,
\end{align}
where $a_i \in \mathbb{R}^d$ and $b_i \in\{-1,+1\}$ are the data samples, and $N$ is their total number. We set the regularization parameter $\lambda=10^{-4} L$, where $L$ is the smoothness constant.

From the results of logistic regression in Figure \hyperref[fig:7]{8} (a), we observe that our FedBCGD and FedBCGD+ algorithms demonstrate faster convergence speed. Particularly, under the strong convexity condition with high client data heterogeneity, our FedBCGD+ algorithm exhibits even faster convergence compared to our FedBCGD, which aligns with our theoretical analysis.

\begin{figure}[H]
	\centering
	\subcaptionbox{ Strongly Convex}{\includegraphics[width=0.45\textwidth]{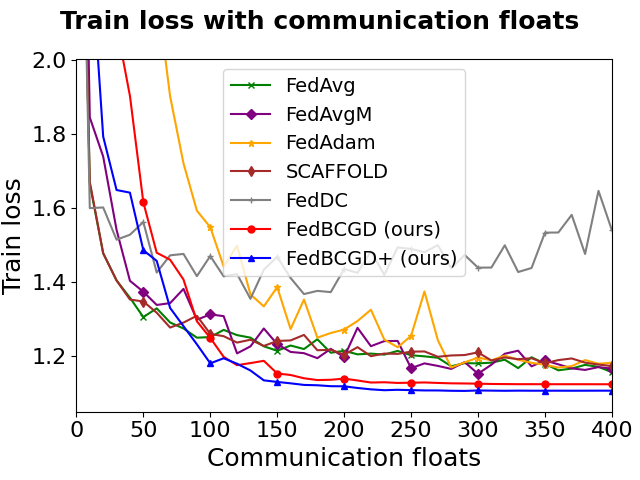}}
	\subcaptionbox{ Non-convex}{\includegraphics[width=0.45\textwidth]{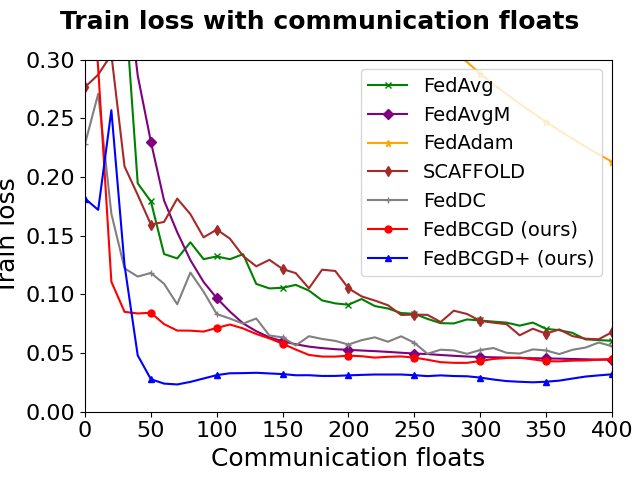}}
	\vspace{-2mm}
	\caption{(a) Logistic regression with $E\!=\!1$ and $\rho\!=\!0.1$. (b) The problem with non-convex loss, where $E\!=\!1$ and $\rho\!=\!0.1$. The number of blocks is set to $N=5$.}
	\label{fig:7}
\end{figure}

\textbf{ERM with Non-Convex Loss:}
We also apply our algorithms to solve the regularized Empirical Risk Minimization (ERM) problem with non-convex sigmoid loss:
\vspace{-2mm}
\begin{align}
	\min _{x \in \mathbb{R}^d} \frac{1}{n} \sum_{i=1}^n f_i(x)+\frac{\lambda}{2}\|x\|^2,
\end{align}
where $f_i(x)=1 /\left[1+\exp \left(b_i a_i^{\top} x\right)\right]$.  Here, we consider binary classification on EMNIST. Note that we only consider classifying the first class in EMNIST.

From the results of the ERM problem in Figure \hyperref[fig:7]{8} (b), we observe that our algorithms exhibit much faster convergence speeds than other algorithms. Moreover, in the case of high client data heterogeneity, FedBCGD+ demonstrates faster convergence than FedBCGD, which is consistent with our theoretical results.

\subsection{Parameter Block Division}
In this section we will show the parameter block division.
\begin{table}[H] 
	\vspace{-2mm}
	\centering
	\setlength{\tabcolsep}{6.19pt}
	\begin{tabular}{llccc}
		\hline &Parameter Block & 
		Network Layers &Number of parameters   &   \\
		
		\hline
		& Block 1 & conv3-64  & 4800 & \\
		& Block 2 & conv3-64  & 102400 & \\
		& Block 3 &  FC-1600 &614400  &  \\
		& Block 4 &  FC-384  &73728  &  \\
		& Block 5 &  FC-192 (share) & 1920 &  \\
		\hline
	\end{tabular}
	\caption{The parameters block division of the LeNet-5 network.}
\end{table}

\begin{table}[H] 
	\vspace{-2mm}
	\centering
	\setlength{\tabcolsep}{6.19pt}
	\begin{tabular}{llccc}
		\hline &Parameter Block & 
		Network Layers &Number of parameters   &   \\
		
		\hline
		& Block 1 & conv3-64  & 1728 & \\
		& Block 1 & conv3-128  & 73728 & \\
		\hline
		& Block 2 &  conv3-256 & 294912 &  \\
		& Block 2 &  conv3-256 &589824  &  \\
		\hline
		& Block 3 &  conv3-512  &1179648  &  \\
		& Block 3 &  conv3-512  &2359296  &  \\
		\hline
		& Block 4 &  conv3-512 &2359296  &  \\
		& Block 4 &  conv3-512 &2359296  &  \\
		\hline
		& Block 5 &  FC-2048 & 2359296 &  \\
		& Block 5 &  FC-2048 &2359296  &  \\
		\hline
		& Block share &  FC-100  & 102400 &  \\
		\hline
	\end{tabular}
	\caption{The parameter block division of the VGG-11 network.}
\end{table}

\begin{table}[H] 
	\vspace{-2mm}
	\centering
	\setlength{\tabcolsep}{6.19pt}
	\begin{tabular}{llccc}
		\hline &Parameter Block & 
		Network Layers &Number of parameters   &   \\
		\hline
		& Block 1 & conv3-64  & 1728 & \\
		& Block 1 & conv3-64  &36864  & \\
		& Block 1 & conv3-64  &36864 & \\
		& Block 1 & conv3-64   &36864 & \\
		& Block 1 & conv3-64   &36864  & \\
		\hline
		& Block 2 &  conv3-128 &73728  &  \\
		& Block 2 &  conv3-128 & 147456 &  \\
		& Block 2 &  conv3-128 & 147456 &  \\
		& Block 2 &  conv3-128 &147456  &  \\
		\hline
		& Block 3 &  conv3-256  &294912  &  \\
		& Block 3 &  conv3-256  &589824  &  \\
		& Block 3 &  conv3-256  &589824  &  \\
		& Block 3 &  conv3-256  &589824  &  \\
		\hline
		& Block 4 &  conv3-512 &1179648  &  \\
		& Block 4 &  conv3-512 &2359296  &  \\
		& Block 5 &  conv3-512 &2359296  &  \\
		& Block 5 &  conv3-512 &2359296  &  \\
		\hline
		& Block share &  FC-512 (share) &5120  &  \\
		\hline
	\end{tabular}
	\caption{The parameters block division of the ResNet-18 network.}
\end{table}

\begin{table}[H] 
	\vspace{-2mm}
	\centering
	\setlength{\tabcolsep}{6.19pt}
	\begin{tabular}{llccc}
		\hline &Parameter Block & 
		Network Layers &Number of parameters   &   \\
		
		\hline
		& Block 1 & conv3-64  &1728  & \\
		& Block 1 & conv3-64  &36864  & \\
		& Block 1 & conv3-128  &73728  & \\
		& Block 1 & conv3-128  &147456  & \\
		\hline
		& Block 1 &  conv3-256 &294912  &  \\
		& Block 1 &  conv3-256 &589824  &  \\
		& Block 1 &  conv3-256 &589824  &  \\
		& Block 1 &  conv3-256 &589824  &  \\
		\hline
		& Block 2 &  conv3-512  &1179648  &  \\
		& Block 2 &  conv3-512  &2359296 &  \\
		& Block 3 &  conv3-512  &2359296  &  \\
		& Block 3 &  conv3-512  &2359296  &  \\
		\hline
		& Block 4 &  conv3-512 &2359296  &  \\
		& Block 4 &  conv3-512 &2359296  &  \\
		& Block 5 &  conv3-512 &2359296  &  \\
		& Block 5 &  conv3-512 &2359296  &  \\
		\hline
		& Block 5 &  FC-2048 &1048576  &  \\
		& Block 5 &  FC-2048 &131072  &  \\
		\hline
		& Block share &  FC-100  &25600  &  \\
		\hline
	\end{tabular}
	\caption{The parameters block division of the VGG-19 network.}
\end{table}

\begin{table}[H] 
	\vspace{-2mm}
	\centering
	\setlength{\tabcolsep}{6.19pt}
	\begin{tabular}{llccc}
		\hline &Parameter Block & 
		Network Layers &Number of parameters   &   \\
		\hline
		& Block 1 & ViT-Block 1  &14299520  & \\
		\hline
		& Block 2 &  ViT-Block 2 &14299520  &  \\
		\hline
		& Block 3 &  ViT-Block 3  &14299520  &  \\
		\hline
		& Block 4 &  ViT-Block 4 &14299520  &  \\
		\hline
		& Block 5 &  ViT-Block 5 &14299520  &  \\
		\hline
		& Block share &  FC-100  &153600  &  \\
		\hline
	\end{tabular}
	\caption{The parameters block division of the VGG-19 network.}
\end{table}

\section{Appendix G: FedBCGD and FedBCGD+ Algorithms} The proposed  FedBCGD+ and FedBCGD algorithms as shown in Algorithms 2 and 3, respectively.

\begin{algorithm}[H]
	\caption{FedBCGD+}
	\begin{algorithmic}[1] %[1] enables line numbers
		\STATE $\textbf{Initialize } \boldsymbol{x}_{i}^{0,0}=\boldsymbol{x}^{i n i t}$, $\forall i \in [M]$.
		\STATE \textbf{Divide} the model parameters $\boldsymbol{x}$ into $N$ blocks.
		\FOR{$r=0,...,R$}
		\STATE \textbf{Client:} 
		\STATE $\textbf{Sample} \text{ clients } \mathcal{S} \subseteq\{1, \ldots, M\}$,$|\mathcal{S}|=N K$;
		\STATE \textbf{Divide} \text{the sampled clients into} $N$ \text{blocks};
		\STATE $\textbf{Communicate }(\boldsymbol{x}, \boldsymbol{c}) \text { to all clients } i \in \mathcal{S}$;
		\FOR{$j=1, \ldots, N$ client blocks in parallel}	
		\FOR{$k=1, \ldots, K$ clients in parallel}	
		\STATE \textbf{Compute} full batch gradient $\nabla f_{k, j}(\boldsymbol{x^r})$;
		\FOR{$t=1, \ldots, T$ local update}	
		\STATE \textbf{Compute} mini-batch gradient $\nabla f_{k, j}\left(\boldsymbol{x}_{k, j}^{r,t} ; \zeta\right)$ and $\nabla f_{k, j}(\boldsymbol{x^r} ; \zeta)$;
		\STATE $\quad \boldsymbol{x}_{k, j}^{r,t+1}=\boldsymbol{x}_{k, j}^{r,t}-\eta \nabla f_{k, j}\left(\boldsymbol{x}_{k, j}^{r,t} ; \zeta\right)+\eta \boldsymbol{c}-\eta \boldsymbol{c}_{k, j}+\eta \nabla f_{k, j}(\boldsymbol{x^r})-\eta \nabla f_{k, j}(\boldsymbol{x^r} ; \zeta)$;				
		\ENDFOR
		\STATE $\boldsymbol{c}_{k,j}^{+} \leftarrow$  $ \nabla f_{k, j} (\boldsymbol{x^r})$;
		\STATE Send $\boldsymbol{x}_{k, j,(j)}^{r, T},\boldsymbol{x}_{k, j,s}^{r, T}$ and $\Delta \boldsymbol{c}_{(j)}=\boldsymbol{c}_{k, j,(j)}^{+}-\boldsymbol{c}_{k, j,(j)},\Delta \boldsymbol{c}_{s}=\boldsymbol{c}_{k, j,s}^{+}-\boldsymbol{c}_{k, j,s}$ to server;
		\STATE $\boldsymbol{c}_i \leftarrow \boldsymbol{c}_i^{+}$;
		\ENDFOR
		\ENDFOR
		\STATE \textbf{Server:}
		\FOR{$j=1, \ldots, N$ Blocks in parallel}	
		\STATE Block $j$ computes,
		\STATE$\boldsymbol{x}_{(j)}^r=\frac{1}{K} \sum_{k=1}^K \boldsymbol{x}_{k, j,(j)}^{r, T}$;
		\STATE$v_{(j)}^r=\lambda v_{(j)}^{r-1}+\boldsymbol{x}_{(j)}^{r}-\boldsymbol{x}_{(j)}^{r-1}$;
		\STATE$\boldsymbol{x}_{(j)}^r=\boldsymbol{x}_{(j)}^{r}+v_{(j)}^{r},$
		\STATE$\boldsymbol{c}_{(j)}=\boldsymbol{c}_{(j)}+\frac{1}{M} \sum_{k=1}^{K} \Delta \boldsymbol{c}_{k, j,(j)}$;
		\ENDFOR
		\STATE$\boldsymbol{x}_{s}^{r}=\frac{1}{NK}\sum_{j=1}^N\sum_{k=1}^K\boldsymbol{x}_{k, j,s}^{r, T}$;
		\STATE	$v_{s}^r=\lambda v_{s}^{r}+\boldsymbol{x}_{s}^{r-1}-\boldsymbol{x}_{s}^{r-1}$;
		\STATE$\boldsymbol{x}_{s}^r=\boldsymbol{x}_{s}^{r}+v_{s}^{r}$;
		\STATE$\boldsymbol{c}_{s}=\boldsymbol{c}_{s}+\frac{1}{MN}\sum_{j=1}^{N} \sum_{k=1}^{K} \Delta \boldsymbol{c}_{k, j,s}$;
		\STATE $\boldsymbol{x}^r=\big[\boldsymbol{x}_{(1)}^{r \top}, \ldots, \boldsymbol{x}_{(N)}^{r \top},\boldsymbol{x}_{s}^{r \top}\big]^{\top}$;
		\STATE $\boldsymbol{v}^r=\big[\boldsymbol{v}_{(1)}^{r \top}, \ldots, \boldsymbol{v}_{(N)}^{r \top},\boldsymbol{v}_{s}^{r \top}\big]^{\top}$;
		\STATE $\boldsymbol{c}=\left[\boldsymbol{c}_{(1)}^{\top}, \ldots, \boldsymbol{c}_{(N)}^{\top},\boldsymbol{c}_{s}^{\top}\right]^{\top}$;
		\ENDFOR
	\end{algorithmic}
	%\label{algorithm:2}
\end{algorithm}

\end{document}